%% file: template.tex
\documentclass[preprints,article,accept,moreauthors,pdftex]{mdpi}

\usepackage{graphicx}
\usepackage{subcaption}

\usepackage{color}
\usepackage{xfrac}
\usepackage{tikz}
\usetikzlibrary{shapes.geometric}

\newlist{paragraphs}{itemize*}{1}
\setlist[paragraphs]{
	label=(\textbf{\thesection}),
	itemjoin=\newline\hspace*{\parindent}
}

\newcommand{\newtext}[1]{\textcolor{black}{{#1}}}
\newcommand{\newtextt}[1]{\textcolor{black}{{#1}}}
\newcommand{\newtexttt}[1]{\textcolor{black}{{#1}}}
\newcommand{\newtextttt}[1]{\textcolor{black}{{#1}}}

\definecolor{almond}{rgb}{0.75, 0.75, 0.75}
\definecolor{dgreen}{rgb}{0.133,0.545,0.133}
\definecolor{amber}{rgb}{1.0, 0.75, 0.0}
\definecolor{antiquewhite}{rgb}{0.98, 0.92, 0.84}
\definecolor{burgundy}{rgb}{0.52, 0., 0.125}

\firstpage{1} 
\pubvolume{xx}
\issuenum{1}
\articlenumber{5}
\pubyear{2019}
\copyrightyear{2019}
\history{Received: date; Accepted: date; Published: date}
\Title{Super-Resolution-based Snake Model --- An Unsupervised Method for Large-Scale Building Extraction using Airborne LiDAR Data and Optical Image}
\Author{Thanh Huy Nguyen $^{1,2\dagger,}$*, %\orcidA{}, 
	Sylvie~Daniel $^{2}$, Didier~Guériot $^{1}$, %\orcidC{}, 
	Christophe~Sintès $^{1}$ %\orcidD{}
	and Jean-Marc~Le~Caillec $^{1}$}%\orcidE{}}

\AuthorNames{Thanh Huy Nguyen, Sylvie~Daniel, Didier~Guériot, Christophe~Sintès and Jean-Marc~Le~Caillec}

\address{%
$^{1}$ \quad IMT Atlantique, Lab-STICC, UMR CNRS 6285, F-29238 Brest, France; 
thanh.nguyen@imt-atlantique.fr (T.H.N.); didier.gueriot@imt-atlantique.fr (D.G.); christophe.sintes@imt-atlantique.fr (C.S.); jm.lecaillec@imt-atlantique.fr (J.-M.L.C.)\\
$^{2}$ \quad Department of Geomatics, Université Laval, Québec City, QC G1V 0A6, Canada; sylvie.daniel@scg.ulaval.ca (S.D.)}

\corres{Correspondence: thanh.nguyen@imt-atlantique.fr; Tel.: +33 7 89 67 93 12%(optional; include country code; if there are multiple corresponding authors, add author initials) +xx-xxxx-xxx-xxxx (F.L.)
}

\abstract{
Automatic extraction of buildings in urban and residential scenes has become a subject of growing interest in the domain of photogrammetry and remote sensing, 
particularly %with the emergence of LiDAR systems 
since mid-1990s. 
Active contour model, colloquially known as snake model, has been studied to extract buildings from aerial and satellite imagery. 
However, this task is still very challenging due to the complexity of building size, shape, and its surrounding environment.
This complexity leads to a major obstacle for carrying out a reliable large-scale building extraction, since the involved prior information and assumptions on building such as shape, size, and color cannot be generalized over large areas.
\newtextt{This paper presents an efficient snake model to overcome such challenge, called Super-Resolution-based Snake Model (SRSM).
The SRSM operates on high-resolution LiDAR-based elevation images---called $ z $-images---generated by a super-resolution process applied to LiDAR data.
The involved balloon force model is also improved to shrink or inflate adaptively, instead of inflating continuously.
This method is applicable for a large scale such as city scale and even larger, while having a high level of automation and not requiring any prior knowledge nor training data from the urban scenes (hence \textit{unsupervised}). 
It achieves high overall accuracy when tested on various datasets.
For instance, the proposed SRSM yields an average area-based Quality of 86.57\% and object-based Quality of 81.60\% on the ISPRS Vaihingen benchmark datasets.
Comparing to other methods using this benchmark dataset, this level of accuracy is highly desirable even for a supervised method.
Similarly desirable outcomes are obtained when carrying out the proposed SRSM on the whole City of Quebec (total area of 656 km$ ^2 $), yielding an area-based Quality of 62.37\% and an object-based Quality of 63.21\%.}
}

\begin{document}
%%%%%%%%%%%%%%%%%%%%%%%%%%%%%%%%%%%%%%%%%%

%%%%%%%%%%%%%%%%%%%%%%%%%%%%%%%%%%%%%%%%%%
%\setcounter{section}{-1} %% Remove this when starting to work on the template.
%\section{How to Use this Template}
%The template details the sections that can be used in a manuscript. Note that the order and names of article sections may differ from the requirements of the journal (e.g., the positioning of the Materials and Methods section). Please check the instructions for authors page of the journal to verify the correct order and names. For any questions, please contact the editorial office of the journal or support@mdpi.com. For LaTeX related questions please contact latex@mdpi.com.
%The order of the section titles is: Introduction, Materials and Methods, Results, Discussion, Conclusions for these journals: aerospace,algorithms,antibodies,antioxidants,atmosphere,axioms,biomedicines,carbon,crystals,designs,diagnostics,environments,fermentation,fluids,forests,fractalfract,informatics,information,inventions,jfmk,jrfm,lubricants,neonatalscreening,neuroglia,particles,pharmaceutics,polymers,processes,technologies,viruses,vision

%\newpage
%\tableofcontents

\section{Introduction}
\unskip
\subsection{Motivation}
%The introduction should briefly place the study in a broad context and highlight why it is important. It should define the purpose of the work and its significance. The current state of the research field should be reviewed carefully and key publications cited. Please highlight controversial and diverging hypotheses when necessary. Finally, briefly mention the main aim of the work and highlight the principal conclusions. As far as possible, please keep the introduction comprehensible to scientists outside your particular field of research. Citing a journal paper \cite{ref-journal}. And now citing a book reference \cite{ref-book}. Please use the command \citep{ref-journal} for the following MDPI journals, which use author-date citation: Administrative Sciences, Arts, Econometrics, Economies, Genealogy, Humanities, IJFS, JRFM, Languages, Laws, Religions, Risks, Social Sciences.

Automatic and accurate extraction of building footprints from urban scenes using remote sensing data %, e.g. high to very high resolution aerial and satellite imagery, or  Light Detection and Ranging (LiDAR) data, 
has become a subject of growing interest for a wide range of applications, such as urban planning \cite{fredericque2008populating}, city digital twin construction \cite{daniel2013geosmartcity}, census studies \cite{xie2015population}, disaster and crisis management, namely earthquake and flood \cite{al2010geo,alamdar2016towards}.
%However, the nature of urban environments can be very complex where buildings can be found with various sizes and shapes within urban areas of different density and vegetation coverage. 
%However, the manual delineation of building footprints from remotely sensed data is expensive, labor-intensive and can be prone to human bias \cite{cordts2016cityscapes}.
%Therefore, automatic methods are needed in order to extract building footprints effectively. % from large urban areas composed of numerous and complex structures.
%
%%Automatic building extraction has become a subject of growing interest in the domain of photogrammetry, computer vision and remote sensing. 
%This research topic has been referred to by many different terms, such as building detection, building footprint delineation, or building instance segmentation and so on. %Also, since building is certainly one of the most important objects of an urban scene, the building extraction is also covered by the subject of 2-D semantic labeling or semantic segmentation, or also known as dense pixel-wise classification.
%%\modif{Despite these different terms, it should be noted that this subject is separated with the task of {three-dimensional (3-D) building reconstruction} or 3-D building modeling. These two tasks are usually performed together with the accuracy of the building reconstruction is subject to a reliable building detection \needref.}
%
%
%\input{literature.tex}
%
%The previous paragraph has described the challenges due to the employed data source.
\newtexttt{
%Production and updating of a \modif{portrait} of flood risk anticipation in the Quebec province (Canada) has been asserted to be of a paramount importance \cite{blin2005cartographie}.
%It also aims at populating a database on the buildings in areas that are high risk to flooding.
This research work addresses an effective solution for extracting buildings from urban and residential environments in a large scale. %, i.e. city scale.
Such task plays an important role in the context of flood risk anticipation, which is asserted with a particular importance in the Quebec province, Canada \cite{blin2005cartographie}. 
Such context requires accurate and regularly updated building location and boundary, which enable the extraction of further essential structural and occupational characteristics of buildings (e.g. first floor, basement openings).
In addition, the scalability of this solution---i.e. the ability to maintain its effectiveness when expanding from a local area to a large area \cite{el2005advanced}---is crucially important considering the scale of the study \newtextttt{(i.e. at the scale of the Quebec province)}.
}

\newtexttt{The nature of urban and residential environments can be very complex, where buildings can be found with various sizes, colors and shapes, within urban areas of different density and vegetation coverage. 
Such complexity is problematic for developing a large-scale building extraction solution. %, \modif{which is desirable...}.
Indeed, a number of studies have been reported over the years with relatively significant results by assuming building shapes \cite{kim1999development,karantzalos2008recognition,ngo2016shape}, enforcing geometrical constraints \cite{gruen2001news}, or limiting on specific urban areas.
However, such assumptions and constraints limit the scalability of the building extraction method, in particular over large areas %, where buildings can be found with various size, shape and color.
%Therefore, more efforts are needed to develop an effective, generic and scalable building extraction solution for large urban areas 
composed of numerous and complex structures.
Based on these premises, it is necessary that %such solution is fast, accurate, and scalable.
%To achieve such needs, 
such solution is (\textit{i}) \newtextttt{versatile---applicable on different urban scenes} without relying on predefined assumptions, constraints, or prior knowledge about the involved scenes and buildings; \newtextttt{(\textit{ii}) highly accurate;} %, \modif{and achieve a high accuracy}; 
(\textit{iii}) and easily scalable over large areas with a relative computational simplicity.
To the best of our knowledge, such a solution has not yet been found. %must not require any assumptions or prior knowledge on the buildings or the urban areas.
%This research work is devoted to an effective building extraction solution over large areas.
}

%%%%%%%%%%%%%%%%%%%%%%%%%%%%%%%%%%%%%%%%%%
\subsection{Literature review}
\newtexttt{
	A large number of building extraction methods have been reported over the last few decades, particularly with the emergence of Light Detection and Ranging (LiDAR) systems since mid-1990s \cite{tomljenovic2015building}. 
	%%However, this task is still very challenging due to the complexity of building size and shape, as well as its surrounding environment. 
	%%Even with the recent developments of Deep Learning approaches for this problem, the task is still considered to be far from being solved \cite{wang2017torontocity}.
	%%\subsection{Literature review and challenges}
	%Based on the type of data utilized, the existing building extraction methods can be classified into three categories: \textit{(i)} using aerial or satellite imagery, \textit{(ii)} using 3-D data, e.g. directly acquired by LiDAR or constructed from multi-view or stereo images, and \textit{(iii)} using jointly imagery and 3-D data.
	However, this task remains very challenging due to various difficulties.
	For instance, many works \cite{huertas1988detecting,lee2003class,turker2015building,huang2019automatic} have been carried out using aerial and satellite imagery. 
	They face many problems due to occlusions, poor contrasts, shadows, and disadvantageous image perspectives \cite{ekhtari2009automatic}.	
	Since height changes allow distinguishing urban objects more effectively than the spectral and textural changes from optical images, numerous works \cite{khoshelham2013segment,zhang2013svm} proposed to exploit 3-D information from LiDAR to extract buildings.
	%In the second category, many works carried out the building extraction by using a digital surface model (DSM), derived directly from LiDAR point cloud \cite{bredif2013extracting} or generated from stereo images \cite{tian2013building}.
	%%Brédif \textit{et al.} \cite{bredif2013extracting} proposed an approach to extract rectangular buildings directly from the DSM, using a Marked Point Process of rectangles and then to refine them into polygonal boundaries.
	%Other research works involve segmentation and classification techniques on LiDAR data, using attributes such as building size, shape, height and Principal Component Analysis features \cite{khoshelham2013segment,zhang2013svm}. 
	However, these methods usually face problems of misclassification of vegetation as buildings \cite{zhang2017advances}. %, where the appearance of trees and buildings can be very similar in complex urban scenes. % \cite{zhang2017advances}.
	In addition, the accuracy of extracted boundaries can be compromised due to the LiDAR point cloud sparsity \cite{chen2012building}. 
	Therefore, many researchers have developed a consensus strategy to use multi-source data in order to increase the building detection rate.
	%
	%In order to take advantage the complementary properties of both data sources, and compensate for each other's weakness, 
	Hence, a number of studies \cite{sohn2007data, awrangjeb2013automatic} focusing on the integration of LiDAR and optical imagery data have been reported.
	%such as \cite{Schenk2002, Habib2005} for better describing building surfaces. 
	%Sohn and Dowman \cite{sohn2007data} focused on exploiting the synergy of IKONOS multispectral imagery combined with a hierarchical segmentation of a LiDAR digital elevation model to extract buildings. 
	%Awrangjeb \textit{et al.} \cite{awrangjeb2010automatic} proposed a building detection based on building masks obtained from LiDAR and multispectral imagery. 
	They succeed at improving the building extraction accuracy, compared to the use of individual data source \cite{zhang2010multi}. 
	%However, they require an accurate registration between the employed datasets.
	However, such approach of integrating multi-source data can be problematic due to data misalignment \cite{gilani2016automatic}. 
}

\newtexttt{
	The International Society for Photogrammetry and Remote Sensing (ISPRS) Working Group II/4 \textquotedblleft3D Scene Reconstruction and Analysis\textquotedblright~provided a taxonomy for methods submitted to the urban object detection benchmark test \cite{rottensteiner2014results}, % made available to the scientific community by the ISPRS Working Group II/4, 
	%As such, building extraction methods are categorized 
	based on their processing strategy.
	%	For instance, Niemeyer \textit{et al.} \cite{niemeyer2012conditional} and Chai \cite{chai2016probabilistic} are two of the most accurate methods in this benchmark test.
	\newtexttt{Some of the methods are categorized as supervised methods requiring training data from LiDAR point cloud or optical image, such as Niemeyer \textit{et al.} \cite{niemeyer2012conditional} and Chai \cite{chai2016probabilistic}. They provided two of the highest accuracy methods submitted to the ISPRS Vaihingen benchmark.}
	Many other methods are categorized as model-based methods, as they rely on an explicit model \newtexttt{or a set of predefined rules} on the appearance of the buildings in the data.
	For instance, Bayer \textit{et al.} \cite{bayerbrief} proposed a segmentation-based method involving multiple thresholds applied on the Digital Surface Model (DSM) and Normalized Difference Vegetation Index (NDVI) to separate buildings and trees. % with an area-based \textit{Quality} of 89.8\%. 
	Similarly, Grigillo \textit{et al.} %(acronym: LJU1 and LJU2) 
	\cite{grigillo2012urban} proposed two versions of a \newtexttt{model-based} method based on rule-set classifiers on image pixel colors and NDVI.
	However, the selection of such thresholds and rules is strongly scene-dependent. %, i.e. set specifically for a particular scene and datasets.
}

%A very interesting building extraction approach involves the study of active contour model, or colloquially called snake model.
Active contour model \cite{kass1988snakes}, or colloquially known as snake model, is an object boundary extraction technique widely used in computer vision and image processing \cite[Ch. 5]{szeliski2010computer}.
This technique has also been intensively studied to extract buildings from urban and residential areas.
\newtextt{In contrast to other approaches mentioned above,} it provides a building extraction solution without prior knowledge about the image and the building shapes. % \cite{kabolizade2010improved}.
\newtextt{Moreover, this technique provides a computational simplicity, and an advantageous flexibility allowing external constraint forces introduced by the user.}
\newtexttt{These characteristics show that snake model is suitable to be developed into a large-scale solution that fits our purposes.}

\subsection{Snake model-based related works}\label{ssec:review_snake_model}
%\modif{With such features, snake model can be promising to be developed into a large-scale unsupervised building extraction solution.} 
%\modif{Snake model-based methods can be categorized either in the first category (i.e. using only optical imagery) or the third one (i.e. using jointly optical imagery and LiDAR data).}
%\modif{What are the benefits of snake models????}
%\modif{Compared with traditional models, the proposed method can converge to the true building contours more quickly and more stably even in complex environments.}
%\subsection{Related works involving snake model}\label{ssec:related}
%Originally introduced by Kass \textit{et al.} \cite{kass1988snakes}, snake model has been intensively studied for building extraction using aerial and satellite imagery. % from urban and residential area. % 
Guo and Yasuoka \cite{guo2002snake} used snake model with balloon force to extract buildings using high-resolution satellite images and height data. 
Peng \textit{et al.} \cite{peng2005improved} focused on improving the stability of snake convergence on aerial images.
Kabolizade \textit{et al.} \cite{kabolizade2010improved} proposed a snake model using imagery data coupled with a DSM generated from LiDAR data.  
This model involves %initializing the snake with edges extracted from the DSM, as well as adding
the minimization of variances of height and gray-level between snake points. 
Consequently, it requires height information for every pixel of the image, in other words, the DSM must be of the same size and resolution as the optical image. 
Such requirement is problematic since LiDAR datasets usually have subsampled spatial resolution compared to the aerial imagery, yet a simple interpolation of height data could be unreliable. 
In contrast, Ahmadi \textit{et al.} \cite{ahmadi2010automatic} proposed a geometrical snake model to detect building boundaries from aerial images, without height information or manual initial points. But this model requires \textit{a priori} gray levels of buildings and ground, and uses them as training data to attract the snakes toward desired buildings. 
Consequently, it yields a high number of mis-detected buildings when they consist of untrained color.
Additionally, it does not work well with the building roofs having varying gray levels.	
Fazan and Dal Poz \cite{fazan2013rectilinear} proposed a method involving exhaustive searches for rectilinear building corners in the optical images, based on the basic snake model optimized by dynamic programing. Yet this method depends heavily on initial points to have decent results. 
Snake models have also been demonstrated as an efficient tool to refine the public Geographic Information System (GIS) building footprints \cite{griffiths2019improving}. 
The improved footprints are then feed into Convolutional Neural Networks (CNNs) as labeled data for the building segmentation.
%To summarize, the main common issues of snake models in building boundary extraction are:
%\textit{(i)} sensitivity to noise and image details, 
%\textit{(ii)} dependence on initial points or training data,
%%	\item Problem of parametrization for each building $ \rightarrow $ overfitting
%\textit{(iii)} weak convergence to building corners, %However, if we reduce the tension weighting parameter, the snake may not converge.
%%	\item In some case, e.g. building that has homogeneous color and very distinctive with respective to its surrounding, meanshift segmentation may even yield better extraction result than snake models
%\textit{(iv)} snake's convergence sensitivity to its number of points and weighting parameters.

%In order to address these limitations of existing snake models, 
Our previous work \cite{nguyen2019unsupervised2} presented an unsupervised and automatic snake model to extract buildings from optical imagery. % and airborne LiDAR data, without manual initial points or training data. 
It is carried out based on a snake model operating on optical image, initialized and enhanced by integrating with LiDAR data.
This snake model involves a novel external energy term computed based on the shape similarity between the snake and the projected LiDAR building boundary. 
Such an energy term  encourages the snake to maintain a  shape similar to the building boundary extracted from LiDAR data, while moving under the attractions of salient features provided by optical image.
%This approach makes the small details and noises from the optical image be not as perturbing as in other snake models.
In contrast to the snake models mentioned above, this method succeeds at extracting buildings in various difficult cases, e.g building roof with similar color to its background, gable-roof houses or varying-color roof buildings. 
%It yields high building extraction accuracy without needing any human intervention nor training data. 
%While being an unsupervised building extraction method, it yields a better accuracy than many other existing methods, and a comparable accuracy against  supervised methods.
Without any human intervention or training data, it is able to achieve higher accuracy than existing snake models and many existing building extraction methods such as \cite{gilani2016automatic,awrangjeb2014automatic,yang2013automated} on multiple test areas (see  \cite{nguyen2019unsupervised2} for the full assessment).
Nevertheless, similarly to other existing snake models, it still concedes a number of challenges, 
namely its sensitivity against image noise and undesired details, and the hyperparameter tuning for snake model in a large scale.

While there is not currently any effective solution regarding the former problem (i.e. snake sensitivity) when using optical imagery,
the latter problem (i.e. hyperparameter tuning) has been partially addressed by Marcos \textit{et al.} \cite{marcos2018learning} with a deep learning-based approach.
It involves using a CNN to learn the characteristics of the snake model elements, i.e. parameters and energy terms, from training optical images and associated ground truth polygons. %, which allow to yield the most accurate building boundaries. 
%The elements learned from CNN are the weights of the internal energy terms of the snake (i.e. $ \alpha $ and $ \beta $), the image-based energy term (i.e $ E_\mathrm{img} $), and the balloon force term (i.e. $ F_\mathrm{balloon} $).
%Additionally, the authors also asserted that scalar $ \alpha $ and $ \beta $ for all buildings can lead to problems of over-smoothing at building corner regions, but under-smoothing at other regions.
%Thus, they proposed a local penalized active contours to avoid such problem, by assigning a different $ \beta $ penalizations to each pixels, depending of which part of the object lies underneath, whereas $ \alpha $ remains scalar for all pixels. 
%Similar approach was applied to the balloon force term $ F_\mathrm{balloon} $.
%This pixel-wise approach of $ \beta $, as well as the external terms $ E_\mathrm{img} $ and $ F_\mathrm{balloon} $---
The CNN-inferred parameters and energy terms enabled this snake model to achieve higher accuracy compared to other deep learning-based building extraction methods. 
However, the main drawback of this method is that it involves every image patch---each one containing a building---to have the same size, i.e. 512 $ \times $ 512 pixels, for both the training dataset and the test dataset. 
It means that all the concerned buildings (training and testing) must have similar size in order for the CNN to learn and predict the parameters and energy terms.
In other words, in order to resolve the snake parametrization problem, this approach proposed by Marcos \textit{et al.} requires the building size consensus.
This requirement affects directly the method reproducibility on buildings of different sizes.
\newtexttt{Consequently, such CNN-based snake parametrization approach is not scalable for large areas consisting of buildings of various sizes.}

\subsection{Contribution}
%	Each one of the  snake models mentioned previously has different problems. 
%Three problems remain the most challenging when applying 
The objective of this research work is to develop a large-scale automatic and accurate building extraction based on snake model, fulfilling the following requirements.
%which can be decomposed into three parts.
Firstly, such an effective snake model would require an automatic and reliable initialization.
Secondly, the snake model should not be sensitive to noise and details in the image.
Thirdly, %it is how to parametrize the snake model on 
the snake model parameters should be relevant when applied to a large extended area with buildings of various shapes, sizes, and colors.
%In this paper, we elaborate these problems and propose an approach to overcome them.
\newtextt{While \newtextttt{the first requirement is addressed by using the boundaries preliminarily extracted from LiDAR point cloud}, the second and the third remain very challenging.
In this regard, the contributions of this work are threefold:
\begin{itemize}[leftmargin=*,labelsep=5.8mm]
	\itemsep1cm
%	\item The preliminary boundary extraction from the LiDAR point cloud is carried out automatically;
%	\item \newtextt{We elaborate on the sparsity problem related to LiDAR data and the snake parametrization problem;}
	\item We propose an effective solution to compute the external energy for the snake model---\newtextttt{which is initialized by the LiDAR-based boundaries.}
	Such solution enables the snake model to be insensitive to image noise and details, as well as easing the snake model parametrization. In addition, this  snake model involves an improved balloon force that behaves adaptively by either shrinking or inflating the snake (as opposed to the classic balloon force that always inflates it).
	\item In order to build a \newtexttt{reliable foundation for this novel snake model}, a super-resolution process is proposed to reliably improve the LiDAR point cloud sparsity. 
	Such sparsity issue has been problematic to \newtexttt{building extraction methods using LiDAR data, including snake models}.
	\item Lastly, we present a comprehensive performance assessment of the proposed SRSM on two different geographical contexts, namely Europe (with the Vaihingen benchmark dataset) and North America (with the Quebec City dataset). \newtexttt{Such contexts involve various differences in terms of compactness, density and regularity of urban areas \cite{huang2007global}, demonstrating the scalability and \newtextttt{versatility} of the proposed method.}
%	\modif{Significant results are obtained as the resulting accuracy of the proposed method on this assessment is shown to be highly desirable, compared with other existing methods.}
\end{itemize}
Together, these elements constitute a large-scale automatic and unsupervised building extraction method, which achieves high thematic and geometrical accuracy when tested on various urban scenes.}

\subsection{Paper organization}
This paper is structured as follows: this Section has been devoted to an introduction to the building extraction research topic, our motivation and a literature review of the related works. The contributions of this research work have also been summarized. 
%Section \ref{sec:snake_model}  brings an introduction of active contour model for building extraction, as well as comprehensive analyses of its related problems. 
Section \ref{sec:method} presents the proposed method. % consisting of multiple processes.
Then, multiple %qualitative and quantitative 
assessments on the performance of the SRSM involving various study areas and datasets are carried out in Section \ref{sec:results}. % and \ref{sec:discussion}. 
Next, Section \ref{sec:discussion} brings the discussions on the relevance of the proposed SR, then on the SRSM results, and lastly on the impact of the snake model parametrization.
Finally, Section \ref{sec:conclusions} provides conclusions and perspectives of this work.

%\input{old_sec2.tex}

%%%%%%%%%%%%%%%%%%%%%%%%%%%%%%%%%%%%%%%%%%
\section{Proposed Method}\label{sec:method}
%\unskip
%\subsection{Flowchart}

\begin{figure}[h]
	\centering
	\begin{tikzpicture}[every text node part/.style={align=center},every node/.style={scale=0.92}]%,font=\sffamily]
	
	\node[draw,fill=almond,text=black,minimum height=0.5cm,text width=2.25cm] (SR) at (8.5,-1.9) {\footnotesize Super-resolution};
	%	\node[text=black,minimum height=0.35cm,text width=1cm] (SR ref) at (9.85,-1.9) {\scriptsize \textit{(\ref{ssec:sr})}};
	\node[trapezium,draw,trapezium stretches=false,trapezium left angle=110, trapezium right angle=70,minimum height=0.35cm,text width=1.5cm] (zimg) at (8.5,-5) {\footnotesize $ z $-image};
	
	\node[trapezium,draw,trapezium stretches=false,trapezium left angle=110, trapezium right angle=70,minimum height=0.35cm,text width=2.5cm] (Lidar) at (5.5,-1) {\footnotesize LiDAR point cloud};
	\node[trapezium,draw,trapezium stretches=false,trapezium left angle=110, trapezium right angle=70,minimum height=0.35cm,text width=3.25cm]  (OptImg) at (1.5,-1.9) {\footnotesize \newtextttt{Registered optical image}};
	
	\node[draw,fill=almond,text=black,minimum height=0.35cm,text width=3cm] (lidarBE) at (5.5,-1.9) {\footnotesize Preliminary extraction};
	%	\node[text=black,minimum height=0.35cm,text width=1cm] (SR ref) at (3.65,-1.9) {\scriptsize \textit{(\cite{nguyen2019unsupervised2})}};
	%	\node[draw,fill=almond,text=black,minimum height=0.5cm,text width=2.25cm] (CR) at (3,-2.25) {\footnotesize Registration};
	%	\node[text=black,minimum height=0.35cm,text width=1cm] (Regis ref) at (3,-1.7) {\scriptsize \textit{(\ref{ssec:regis})}};
	%	\node[trapezium,draw,trapezium stretches=false,trapezium left angle=110, trapezium right angle=70,minimum height=0.35cm,text width=2cm]	(TM) at (6,-3) {\footnotesize Camera pose {\small $ \theta $}};
	
	\node[trapezium,draw,trapezium stretches=false,trapezium left angle=110, trapezium right angle=70,minimum height=0.7cm,text width=3.25cm] (Bi) at (5.5,-2.95) {\footnotesize Candidate 3-D building boundary points $ \mathbf{B}_i $};
	\node[trapezium,draw,trapezium stretches=false,trapezium left angle=110, trapezium right angle=70,minimum height=0.35cm,text width=2.5cm] (Mi) at (2,-5) {\footnotesize Building masks $ \mathbf{M}_i $};
	\node[draw,fill=almond,text=black,minimum height=0.35cm,text width=3cm]
	(TM*Bi) at (5.5,-4) {\footnotesize 3-D projection $ \mathcal{P}_\theta(\mathbf{B}_i) $};
	\node[trapezium,draw,trapezium stretches=false,trapezium left angle=110, trapezium right angle=70,minimum height=0.35cm,text width=1cm]
	(bi0) at (5.5,-5) {\footnotesize $ \mathbf{b}_i^0 $};
	\node[draw,fill=almond,text=black,minimum height=0.5cm,text width=2cm] (snake) at (5.5,-6) {\footnotesize Snake model};
	%	\node[text=black,minimum height=0.35cm,text width=2cm] (SR ref) at (7.35,-6) {\scriptsize \textit{(\ref{ssec:z_snake_model}) + (\ref{ssec:snake_param})}};
	\node[trapezium,draw,trapezium stretches=false,trapezium left angle=110, trapezium right angle=70,minimum height=0.7cm,text width=3.3cm] (bi) at (5.5,-7.1) {\footnotesize $ \mathbf{b}_i=\{x_i,y_i\} $: sets of building boundary points};
%	\node[draw,fill=almond,text=black,minimum height=0.5cm,text width=2.5cm] (poly) at (5.5,-8.1) {\footnotesize Polygonization};
%%	\node[text=black,minimum height=0.35cm,text width=1cm] (SR ref) at (7.15,-8.1) {\scriptsize \textit{(\ref{ssec:poly})}};
%	\node[trapezium,draw,trapezium stretches=false,trapezium left angle=110, trapezium right angle=70,minimum height=0.7cm,text width=2.5cm] (Si) at (5.5,-9.1) {\footnotesize $ \mathbf{s}_i $: sets of building polygon};
	%	
	%	\draw[->,draw=black] (Lidar) -- (CR);
	\draw[->,draw=black] (OptImg) -- (lidarBE);
	\draw[->,draw=black] (Lidar) -- (SR);
	\draw[->,draw=black] (Lidar) -- (lidarBE);
	%	\draw[->,draw=black] (OptImg) --node[sloped,above,midway,text width=2.5cm]{\scriptsize NDVI}  (lidarBE);
	\draw[->,draw=black] (lidarBE) -- (Bi);
	\draw[->,draw=black] (Mi.east) --node[sloped,above,midway,text width=2cm]{\scriptsize balloon force}  (snake);
	%	\draw[->,draw=black] (CR) -- (TM);
	%	\draw[->,draw=black] (TM) -- (SR);
	\draw[->,draw=black] (Bi) -- (TM*Bi);
	%	\draw[->,draw=black] (TM) -- (TM*Bi);
	\draw[->,draw=black] (TM*Bi) -- (bi0);
	\draw[->,draw=black] (TM*Bi) -- (Mi);
	\draw[->,draw=black] (SR) -- (zimg);
	\draw[->,draw=black] (bi0) --node[sloped,above,midway,text width=2cm]{\scriptsize init.} (snake);
	\draw[->,draw=black] (zimg) --node[sloped,above,midway,text width=0.5cm]{\footnotesize $ E_{\mathrm{img}} $}  (snake);
	\draw[->,draw=black] (snake) -- (bi);
%	\draw[->,draw=black] (bi) -- (poly);
%	\draw[->,draw=black] (poly) -- (Si);
	\end{tikzpicture}
	\caption{Flowchart of the proposed building extraction method based on the SRSM.}%The section number in parentheses next to a procedure block denotes their respective descriptive part.}% in this paper.}% \modif{TOP-DOWN PRESENTATION}} 
	\label{fig:flowchart_all}
\end{figure}
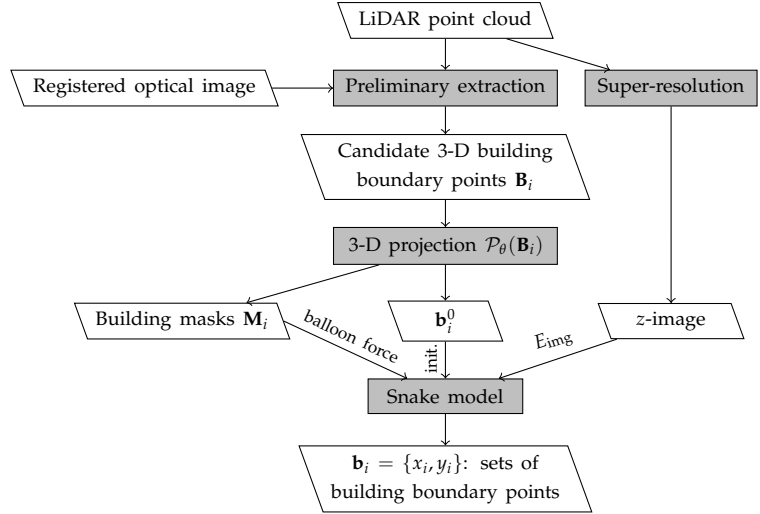

This paper presents a novel unsupervised building extraction method, built around the super-resolution-based snake model (SRSM).
Figure \ref{fig:flowchart_all} depicts the flowchart of the proposed method. % large-scale building extraction method based on an improved snake model.
%Such data provide a more accurate cue for extracting buildings compared to optical imagery, and by extension, allow to avoid many optical image aberrations mentioned above.
%However, a building extraction method using solely LiDAR data can suffer from other problems such as point cloud sparsity, and occlusion and misclassification with nearby trees and vegetation.
%In order to determine the initial points for the snake model
\newtextt{First, the SRSM is automatically initialized by the preliminary candidate building boundaries extracted from the LiDAR point cloud. 
This extraction process is carried out as presented in \cite{nguyen2019unsupervised2}.}
\newtexttt{Since LiDAR-based building extraction can be difficult due to nearby vegetation, this process also involves a vegetation removal based on the Normalized Difference Vegetation Index (NDVI) derived from an optical image \cite{rottensteiner2007building}.
As the two data sources are used jointly, a registration is necessary in order to avoid misalignment problems.
This registration can be carried out \textit{a priori} (i.e. data acquisition using the same platform) or \textit{a posteriori} \cite{nguyen2019robust,nguyen2019coarsetofine}.}
%\modif{The building regions extracted from LiDAR point cloud are used under two forms, namely 3-D building boundary points denoted by $ \mathbf{B}_i $ and 3-D building masks denoted by $ \mathbf{M}_i $.}
The 3-D building boundary points extracted from the LiDAR point cloud are denoted by $ \mathbf{B}_i $, where $ i $ represents the building index. 
\newtexttt{The registration results in a set of camera pose parameters $ \theta $, which is then used for the projection of the 3-D building boundary points $ \mathbf{B}_i $ onto the image space, denoted by $ \mathcal{P}_\theta(\mathbf{B}_i) $.}
Then, they are used as initial points (denoted by $ \mathbf{b}_i^0 $) for the snake model, as well as to generate the building masks (denoted by  $ \mathbf{M}_i $) used in the balloon force.
%The proposed method uses additional information from optical imagery in order to remove vegetation aberrations. % from the building extraction process from LiDAR point cloud.
%\newtextt{Since a LiDAR-based building extraction can be difficult due to nearby vegetation, this process also involves a vegetation removal based on the Normalized Difference Vegetation Index (NDVI) derived from the optical image \cite{rottensteiner2007building}.
%As the two data sources are used jointly, a registration is necessary in order to avoid misalignment problems.
%Thus, a registration can be carried out a priori (i.e. data acquisition using the same platform) or a posteriori \cite{nguyen2019robust,nguyen2019coarsetofine}.}
%The registration results in a set of camera pose parameters $ \theta $, which is then used for the projection the 3-D building boundary points $ \mathbf{B}_i $ onto the image space, denoted by $ \mathcal{P}_\theta(\mathbf{B}_i) $.
The SRSM operates on high-resolution LiDAR-based  $ z $-images generated by a super-resolution process.
%This process is presented in sub-section \ref{ssec:sr}, 
%This process is the intended solution to the point cloud sparsity problem mentioned above.
%and the snake model is described in sub-section \ref{ssec:z_snake_model}.
It also involves an improved balloon force model based on the building masks $ \mathbf{M}_i $.
The resulting building boundary is denoted by $ \mathbf{b}_i $.
%\modif{The related snake parametrization will be discussed in sub-section \ref{ssec:snake_param}.
%Lastly, a polygonization step is carried out to regularize the resulting boundaries into polygonal segments, described in sub-section \ref{ssec:poly}.}
%\newtextt{Lastly, the resulting boundaries are regularized into proper polygonal segments---allowing further usages, such as integration into a GIS database \cite{maggiori2017polygonization}---using the ERSI ArcMap built-in boundary regularization algorithm \cite{gribov2017searching}.}

%\vspace{2cm}
%\modif{All these spaces should be removed when we remove the Table of content.}

%\input{preliminary_BE.tex}

%\input{registration_and_SR.tex}

%\subsection{Proposed snake model}\label{ssec:z_snake_model}
%\subsection{Super-resolution-based snake model}\label{ssec:z_snake_model}
%This sub-section describes the snake model based on \modif{the $ z $-image resulted from the presented SR process.}
%\subsubsection{Traditional Snake Model}\label{ssec:snake_model}
\subsection{Mathematical formulation}\label{ssec:snake_model}
An active contour, or a snake is a dynamic curve $ \mathbf{x}(s)=(x(s),y(s)) $, where $ s \in [0,1]$ is the normalized arc length, defined within an image domain that is deformable under the influence of internal and external forces. The behaviors of the snake are governed by an energy function defined as follows,

\begin{subequations}\label{eq:snake1}
	\begin{equation}
	E_{\mathrm{snake}} = \int_{0}^{1}(E_{\mathrm{int}}(\mathbf{x}(s)) + E_{\mathrm{ext}}(\mathbf{x}(s))) ds%\\[10pt]
	\end{equation}
	\begin{equation}
	\textnormal{with~~~}~E_{\mathrm{int}}(\mathbf{x}(s)) = \dfrac{1}{2} \left(\alpha \left|\dfrac{\partial\mathbf{x}}{\partial s}\right|^2 + \beta\left|\dfrac{\partial^2\mathbf{x}}{\partial s^2}\right|^2\right) %\\[10pt]
	\end{equation}
	\begin{equation}
	\textnormal{and~~~}~E_{\mathrm{ext}}(\mathbf{x}(s)) = E_{\mathrm{img}}(\mathbf{x}(s)) + E_{\mathrm{con}}(\mathbf{x}(s))
	\end{equation}
\end{subequations}
where  $ E_{\mathrm{int}} $ and $ E_{\mathrm{ext}} $, respectively, represent the internal and external energy terms. 
The internal energy term relates to the amount of stretch and curvature of the snake, respectively controlled by weighting parameters $ \alpha $ and $ \beta $. % are weighting parameters that control, respectively, the tension (the amount of stretch) and the rigidity (the amount of curvature) of the active contour.  
%Whereas $ \mathbf{x}'(s) $ and $ \mathbf{x}''(s) $
%denote the first and second derivatives of $ \mathbf{x} $ with respect to $ s $. 
Small value of $ \alpha $ and $ \beta $, respectively, encourage short and smooth contours, and vice versa. 
The external energy $ E_{\mathrm{ext}} $ is composed of the forces due to the image itself $ E_{\mathrm{img}} $ and other constraint forces $ E_{\mathrm{con}} $. 
%For example, Cohen \textit{et al.} proposed an inflation force introduced by balloon model \cite{cohen1991active}.
%This model mimics the inflation of a balloon by continuously pushing the snakes' vertices outwards. 
%Thus it prevents the snake to collapse to a single point.
%The notation $ E_{\mathrm{img}}(I, (\mathbf{x}(s))) $ indicates the value in $ E_{\mathrm{img}}(I) $ indexed by the position $ \mathbf{x}(s)=(x(s),y(s)) $.
%For numerical implementation \cite{kass1988snakes}, the snake can be considered  as a polygon of $ N $ nodes, $ \mathbf{x}_i=(x_i,y_i), i=1..N $.
The external image-based energy $ E_{\mathrm{img}} $ involving salient features of the image i.e. lines, edges and terminations (i.e. line segment end-points, corners) is formulated as follows,

\begin{equation}\label{eq:E_img}
E_{\mathrm{img}}=w_{line}E_{line}+w_{edge}E_{edge}+w_{term}E_{term}
\end{equation}
where $ w_{line}, w_{edge}, w_{term} $ are the weights of the respective salient features. Mathematical formulation of these energy terms \cite{kass1988snakes} are provided in Appendix \ref{app:img_ext}.

A snake that minimizes $ E_{\mathrm{snake}}$ described by Equation \eqref{eq:snake1} must satisfy the following Euler equation,

\begin{equation}\label{eq:euler}
\alpha\times\dfrac{\partial\mathbf{x}^2}{\partial s^2} + \beta\times\dfrac{\partial\mathbf{x}^4}{\partial s^4} + \nabla E_{\mathrm{ext}}=0
\end{equation}
In order to solve Equation \eqref{eq:euler}, the snake is made dynamic by regarding $ \mathbf{x} $ as a function of time $ t $ as well as of the arc length $ s $. Then, the partial derivative of $ \mathbf{x} $ with respect to $ t $ is then set equal to the left hand side of Equation \eqref{eq:euler}, as follows,

\begin{equation}\label{eq:snake_iter}
\dfrac{\partial\mathbf{x}}{\partial t}=-\alpha\times\dfrac{\partial\mathbf{x}^2}{\partial s^2} - \beta\times\dfrac{\partial\mathbf{x}^4}{\partial s^4} - \nabla E_{\mathrm{ext}}%(\mathbf{x})
\end{equation}
As $ \mathbf{x}(s,t) $ stabilizes, the partial derivative term $ {\partial\mathbf{x}}/{\partial t} $ vanishes and a solution for Equation \eqref{eq:euler} is obtained. 
%This dynamic equation can also be viewed as a gradient descent algorithm designed to solve \eqref{eq:snake1}.
A numerical approach for Equation \eqref{eq:snake_iter} can be carried out by discretizing the equation and solving the discrete problem iteratively \cite{kass1988snakes}.

%\subsubsection{Snake robustness against image details and noises} % ---  constraint forces}
%\label{sssec:constraint}

External constraint forces are added to the snake energy function in order to guide it toward or away from a particular feature, as well as addressing snake problems such as initialization, convergence, robustness against noise.
In this regard, Xu and Prince \cite{xu1997gradient} proposed Gradient Vector Flow (GVF) to improve the traditional snake model by allowing more flexible initialization and encouraging its convergence to boundary concavities, as well as improving its robustness. 
%\modif{Detailed descriptions on GVF can be found in Appendix \ref{app:gvf}.}
GVF field is defined as the vector field $ \mathbf{v}(x,y) = (u(x,y), v(x,y)) $ that minimizes the energy functional

\begin{equation}\label{gvf1}
E_{\mathrm{GVF}} = \int\int\mu_{\mathrm{GVF}} (u_x^2+u_y^2+v_x^2+v_y^2) + |\nabla f|^2|\mathbf{v}-\nabla f|^2dxdy
\end{equation}
with $ \mu_{\mathrm{GVF}}  $ is a controllable smoothing term, and $ f $ represents external forces from Equation \eqref{eq:euler}, i.e. $ f(x,y)=-E_\mathrm{ext} $.
Using  \cite{courant2008methods} the GVF field $ \mathbf{v} $ can be found by solving

\begin{subequations}\label{gvf2}
	\begin{equation}
	%\begin{array}{l}
	\mu_{\mathrm{GVF}} \nabla^2u-(u-f_x)(f_x^2+f_y^2)=0%\\[5pt]
	\end{equation}
	\begin{equation}
	\mu_{\mathrm{GVF}} \nabla^2v-(v-f_y)(f_x^2+f_y^2)=0
	%\end{array}
	\end{equation}
\end{subequations}
where $ \nabla^2 $ is the Laplacian operator. The Euler equations \eqref{gvf2} can also be solved by regarding $ u $ and $ v $ as functions of time,

\begin{subequations}
	%\begin{array}{l}
	\begin{equation}
	\dfrac{\partial u}{\partial t} =\mu_{\mathrm{GVF}} \nabla^2u(x,y,t)- [u(x,y,t)-f_x(x,y)]\cdot[f_x(x,y)^2+f_y(x,y)^2]
	\end{equation}
	\begin{equation}
	\dfrac{\partial v}{\partial t} =\mu_{\mathrm{GVF}} \nabla^2v(x,y,t)-[v(x,y,t)-f_y(x,y)]\cdot[f_x(x,y)^2+f_y(x,y)^2]
	\end{equation}
	%\end{array}
\end{subequations}
Once computed $ \mathbf{v}(x,y) $ replaces the potential force $ -\nabla E_{\mathrm{ext}} $ in the dynamic Equation \eqref{eq:euler}, yielding

\begin{equation}
\dfrac{\partial\mathbf{x}}{\partial t}  = -\alpha\times\dfrac{\partial\mathbf{x}^2}{\partial s^2} - \beta\times\dfrac{\partial\mathbf{x}^4}{\partial s^4} + \mathbf{v}
\end{equation}
This equation is solved similarly as the traditional snake model, i.e. by discretization and iterative solution. The parametric curve solving the above dynamic equation is thus called a GVF snake.

Cohen \textit{et al.} proposed an inflation term as an external force, known as balloon model \cite{cohen1991active}, as follows,
%The balloon model introduces an inflation term into the forces acting on the snake

\begin{equation}\label{eq:balloon_force}
F_{\mathrm{balloon}}=\kappa \times \vec {n}(s)
\end{equation}
where $ \kappa $ is the magnitude of the force and $ {\vec  n}(s) $ stands for the normal unitary vector of the curve at $ \mathbf{x}(s) $. 
This model mimics the inflation of a balloon by continuously pushing the snake points outward. 
Thus it prevents the snake from shrinking into a single point.

%
%%\subsection{Several problems of the optical image-based snake model}
%\modif{In \cite{nguyen2019unsupervised2}, a novel external energy term was also proposed. It is based on the shape similarity between the snake and the projected LiDAR building boundary. 
%The proposed energy term aims to encourage the snake to maintain a similar shape as the building boundary extracted from LiDAR data, while moving under the attractions of image salient features and GVF fields.
%This makes the small details and noises from the optical image be not as perturbing as in other snake models.
%As a result, this model is able to achieve higher accuracy than existing snake models, as well as many existing building extraction methods such as \cite{gilani2016automatic,awrangjeb2014automatic,yang2013automated} on multiple test areas (see  \cite{nguyen2019unsupervised2} for the full assessment).}

%\newpage
\subsection{Proposed $ z $-image-based energy term}\label{sssec:img-term}
Despite the recent developments, % aforementioned in \ref{ssec:review_snake_model}, 
the existing snake models \newtextt{still struggle to yield a satisfactory reproducibility in complex environments.
Such low reproducibility stems from a number of reasons. 
For instance, these environments can be composed of complex structures such as multi-planar roof buildings which can also be shadowed or occluded by trees. 
For a multi-planar roof building, different roof planes can have different shades, causing ridge lines (i.e. the intersection lines between the different planes) exhibiting high-gradient values in the image-based energy term.}
In addition, the performance of the snake model on optical images can be affected by image small details, namely roof objects (like chimneys, attic windows), cars, trees, etc.  
There also are  possible null-valued pixels on the orthoimage. %, e.g. the case of ISPRS Vaihingen datasets. 
Consequently, if a building involves these unwanted elements, then the snake model would be drawn toward them. 
Hence, the resulting performance on delineating such building would decrease significantly. 
%\modif{\textbf{Need an illustrating figure...}}
%
%\begin{figure}[H]
%	\tiny
%	\centering
%	\begin{subfigure}{0.24\linewidth}
%		\centering\includegraphics[trim=6.2cm 2.8cm 4.8cm 1.8cm,clip,height=3.25cm]{fig/opt_img_w_gt.png}\caption{}\label{sfig:opt_img_w_gt}
%	\end{subfigure}
%	\hspace{0.01cm}
%	\begin{subfigure}{0.24\linewidth}
%		\centering\frame{\includegraphics[trim=7.5cm 3.5cm 6.25cm 2.4cm,clip,height=3.25cm]{fig/grad_opt_img.png}}\caption{}\label{sfig:grad_opt}
%	\end{subfigure}
%	\hspace{0.01cm}
%	\begin{subfigure}{0.24\linewidth}
%		\centering\frame{\includegraphics[trim=7.5cm 3.5cm 6.25cm 2.4cm,clip,height=3.25cm]{fig/grad_z_img.png}}\caption{}\label{sfig:grad_z}
%	\end{subfigure}
%	\hspace{0.01cm}
%	\begin{subfigure}{0.24\linewidth}
%		%		\centering\includegraphics[trim=4.5cm 2.8cm 5.5cm 2cm,clip,height=3.25cm]{fig/seg_res_lidar2}\caption{}\label{sfig:}
%	\end{subfigure}
%	\caption{Comparing between the gradients of an optical image and a $ z $-image. (\textbf{a}) Optical image presenting a building with the ground truth boundaries in yellow; (\textbf{b}) Gradient magnitude of the optical image; (\textbf{c}) Gradient magnitude of the associated $ z $-image generated from LiDAR data.}
%	\label{fig:opt_img_vs_z_img}
%\end{figure}
%
Fortunately, these problems relate directly to the use of the optical image.
Therefore, we propose to operate the snake model on the $ z $-image derived from LiDAR data. 
%This $ z $-image is resulted from a process of super-resolution of LiDAR elevation data, thus making it have the same resolution as the optical image.
%As a result, we avoid the mentioned undesired features and detail due to the optical image.
This approach allows the snake model to focus only on the most salient features in a $ z $-image, i.e. height changes involving off-terrain objects such as building and trees.

\input{registration_and_SR_new.tex}

\subsubsection{The $ z $-image based energy term}
\begin{figure}[h]
	\tiny
	\centering
	\begin{subfigure}{0.2\linewidth}
		\centering\includegraphics[trim=3.5cm 1.5cm 5cm 1cm,clip,height=5.5cm]{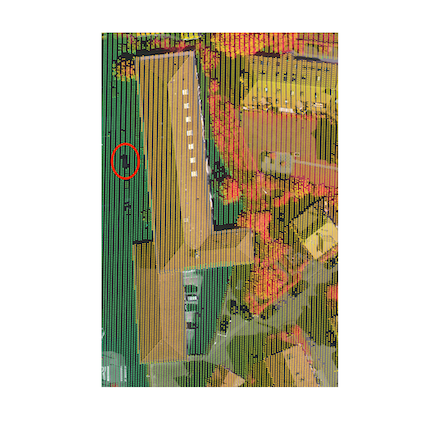}\caption{LiDAR+image}\label{sfig:3.4_0}
	\end{subfigure}
	\hspace{0.01cm}
	\begin{subfigure}{0.17\linewidth}
		\centering\includegraphics[trim=3.5cm 3cm 4cm 1cm,clip,height=5.5cm]{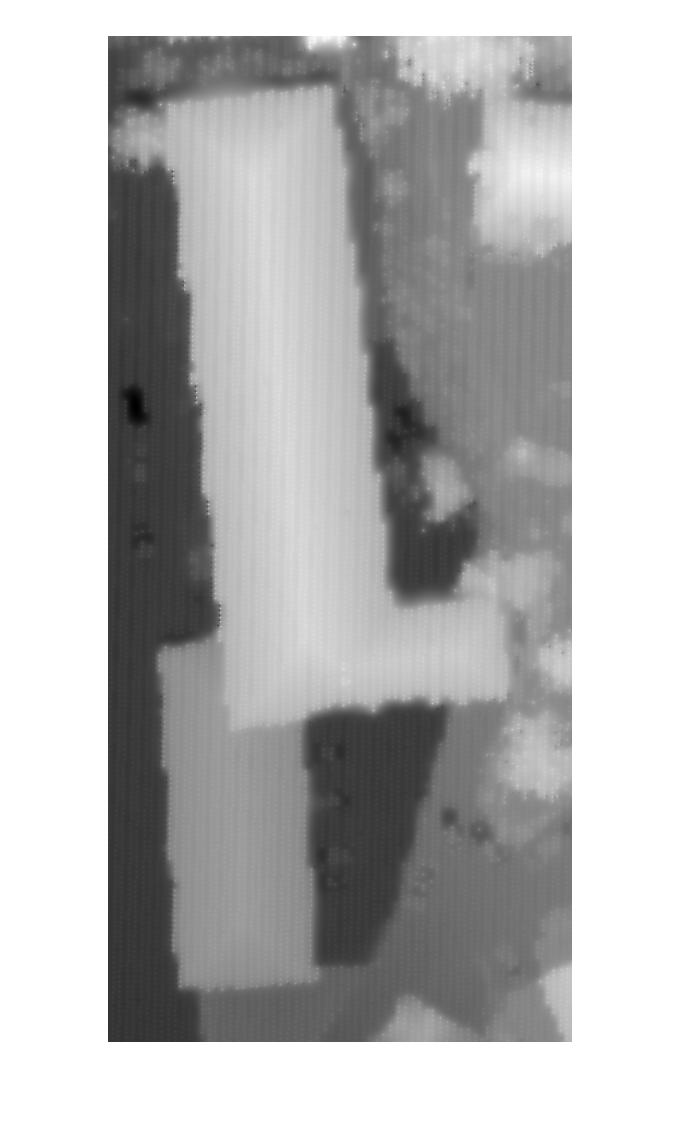}\caption{$ z $-image $ \phi $}\label{sfig:3.4_1}
	\end{subfigure}
	\hspace{0.01cm}
	\begin{subfigure}{0.17\linewidth}
		\centering\includegraphics[trim=14cm 1cm 10cm 0cm,clip,height=5.5cm]{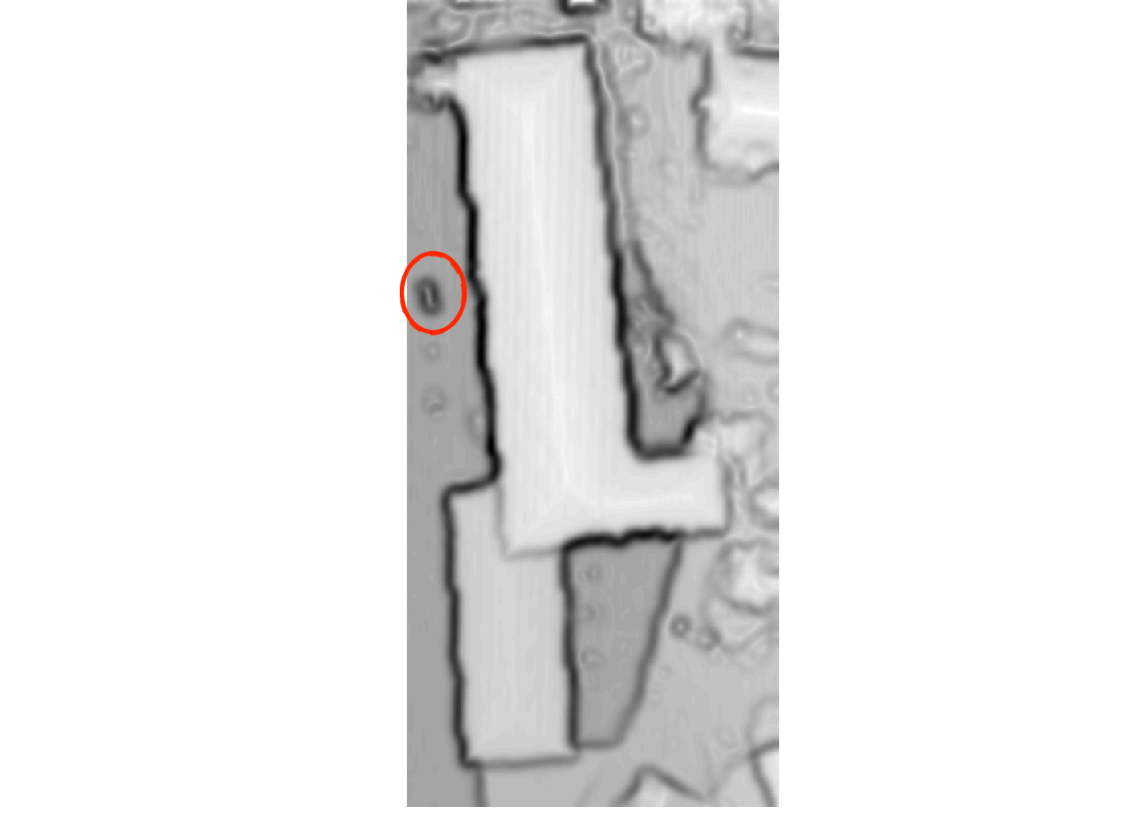}\caption{$ E_\mathrm{img}(\phi) $}\label{sfig:3.4_2}
	\end{subfigure}
	\hspace{0.01cm}
	\begin{subfigure}{0.17\linewidth}
		\centering\includegraphics[trim=4cm 3cm 4cm 2cm,clip,height=5.5cm]{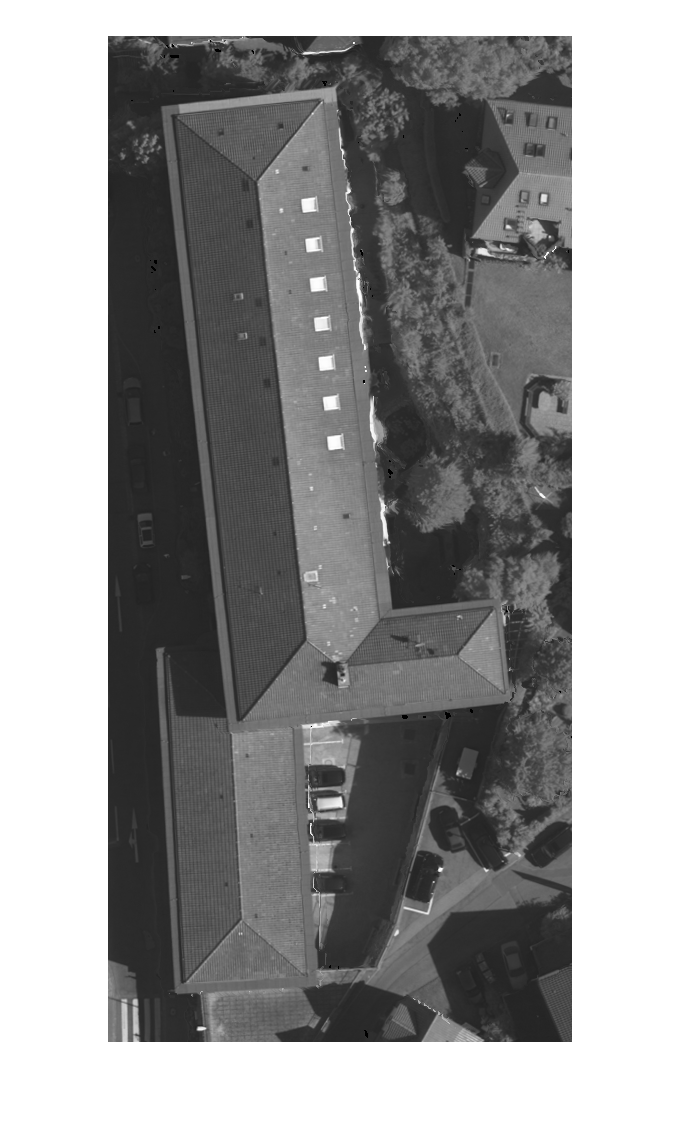}\caption{Optical image $ I $}\label{sfig:3.4_3}
	\end{subfigure}
	\hspace{0.01cm}
	\begin{subfigure}{0.17\linewidth}
		\centering\includegraphics[trim=14cm 1cm 10cm 0cm,clip,height=5.5cm]{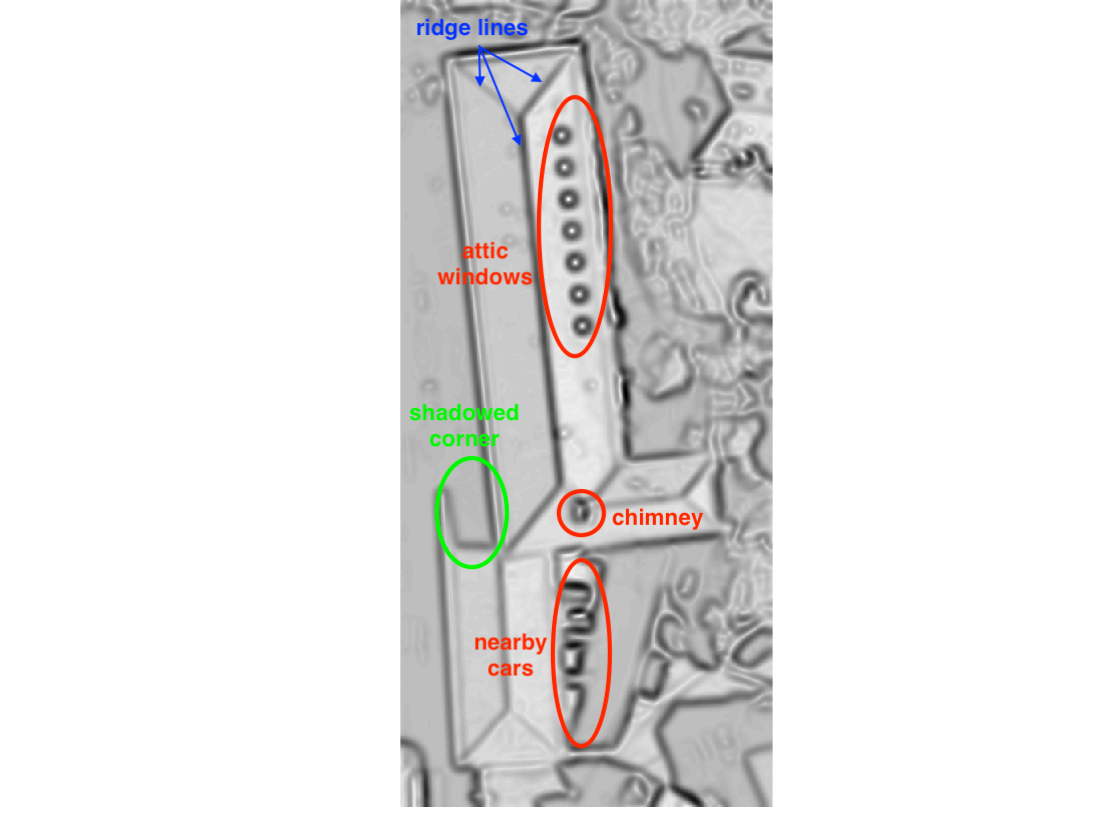}\caption{$ E_\mathrm{img}(I) $}\label{sfig:3.4_4}
	\end{subfigure}
	%	\hspace{0.01cm}
	%	\begin{subfigure}{0.24\linewidth}
	%		\includegraphics[trim=5cm 2.8cm 6.75cm 2cm,clip,height=3.25cm]{fig/labeled2}\caption{}\label{sfig:3.4_3}
	%	\end{subfigure}
	%	\hspace{0.01cm}
	%	\begin{subfigure}{0.24\linewidth}
	%		\includegraphics[trim=4.5cm 2.8cm 5.5cm 2cm,clip,height=3.25cm]{fig/seg_res_lidar2}\caption{}\label{sfig:3.4_4}
	%	\end{subfigure}
	\caption{Comparison between the energy terms computed from the $ z $-image and from the optical image. (\textbf{a}) LiDAR 3-D point cloud overlaying on the optical image for visual comparison; (\textbf{b}) $ z $-image; (\textbf{c}) $ E_\mathrm{img} $ computed from the $ z $-image; (\textbf{d}) optical image; (\textbf{e}) $ E_\mathrm{img} $ computed from the optical image.}% In  (\textbf{c}) and  (\textbf{e}) where the grayscales reflect the values of $ E_\mathrm{img} $, a dark pixel represents a low value, whereas a bright one represents high value, as such the snake will have the tendency to move towards dark pixels.}
	\label{fig:z_snake}
\end{figure}

Figure \ref{fig:z_snake} depicts a comparison between the use of the $ z $-image $ \phi $ and the optical image (denoted by $ I $) of a multi-planar roof building with several roof objects and nearby cars.
Figure \ref{sfig:3.4_0} reveals the LiDAR point cloud overlaying on the optical image of the exemplified building.
Figure \ref{sfig:3.4_1} and \ref{sfig:3.4_2} show, respectively, the z-image and the energy term computed from the $ z $-image.
Then, the optical image and the associated energy term are depicted, respectively, in Figure \ref{sfig:3.4_3} and \ref{sfig:3.4_4}.
%The image-based energy term computed from the $ z $-image and the optical image are shown, respectively, by Figure \ref{sfig:3.4_2} and \ref{sfig:3.4_4}.}
In Figure \ref{sfig:3.4_2} and \ref{sfig:3.4_4}, the grayscale reflects the value of the energy term $ E_\mathrm{img} $.
The dark pixels represent the low-energy pixels, whereas the bright pixels are the high-energy ones.
By design, a snake is attracted to the dark pixels, and will iteratively move toward them.
Comparing the two energy terms (Figure \ref{sfig:3.4_2} and \ref{sfig:3.4_4}), it can be noted that the sources of attraction for the snake models, i.e. the dark pixels in the energy term, provided by the $ z $-image are more relevant than the ones from the optical image.
Indeed, the dark pixels from the energy term computed from the z-image $ E_\mathrm{img}(\phi) $ (Figure \ref{sfig:3.4_2}) are found mainly at the edges of the building, with a few exceptions caused by the nearby trees. 
There exists also one particular aberration which is circled in red on Figure \ref{sfig:3.4_2}. 
It is caused by an absence of LiDAR points in this small region, as highlighted in the red circle on Figure \ref{sfig:3.4_0}.

On the other hand, the dark pixels from the optical image-based energy term $ E_\mathrm{img}(I) $ (Figure \ref{sfig:3.4_4}) stem from many undesirable artifacts, namely cars, attic windows and chimneys found in this building roof.
They are highlighted by the red circles on Figure \ref{sfig:3.4_4}. 
\newtexttt{By comparing Figure \ref{sfig:3.4_2} and \ref{sfig:3.4_4}, it can be noted that some of these artifacts do not exist, or are much less visible on the $ z $-image---due to their low elevation variations compared to the building-to-ground ones.} %---and thus less problematic to the snake model operating on $ z $-image.} 
Also, the effect of shadow casting over a corner of the building---circled in green on Figure \ref{sfig:3.4_4}---is also problematic to a building boundary extraction.
Such effect and problem do not exist in the $ z $-image.
We can also remark that this multi-planar roof building exhibit many ridge lines which also produce low values in the energy term. 
In reality, they are not false non-building details like trees or cars, but when focusing on the extraction of the building boundaries, they can be considered undesirable.
These premises show that it is more relevant to carry out the snake model on the $ z $-image than on the optical image.

\subsection{Improved balloon force}
%As proposed by Marcos \textit{et al.} \cite{marcos2018learning}, the balloon force can be learned from the training data, instead of  relying only on the local normal vector of the curve.
%These masks are provided by the process of preliminarily extracting the buildings from LiDAR data, and not the ground truth building boundaries.
%These masks are then used to compute the balloon force.
%As one can see previously in \ref{sssec:constraint}, 
\begin{figure}[h]
	\tiny
	\centering
	\begin{subfigure}[b]{0.4\linewidth}
		\centering\includegraphics[trim=0 0.25cm 0 0,clip,height=3.5cm]{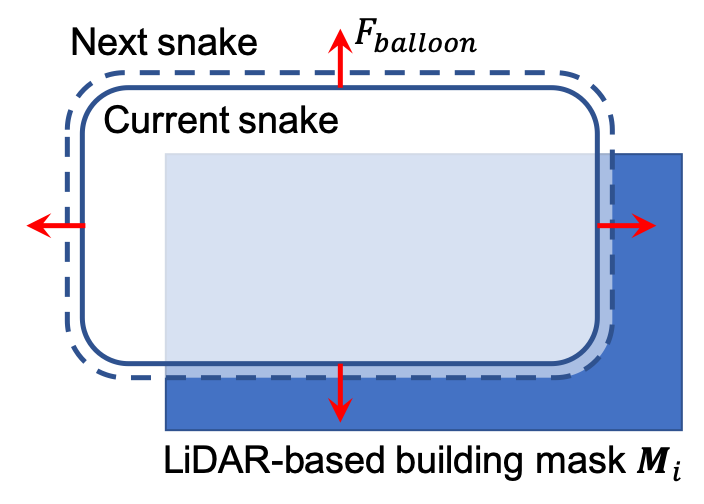}\caption{}\label{sfig:old_balloon}
	\end{subfigure}
	\hspace{0.01cm}
	\begin{subfigure}[b]{0.4\linewidth}
		\centering\includegraphics[trim=0 0.25cm 0 0,clip,height=3.5cm]{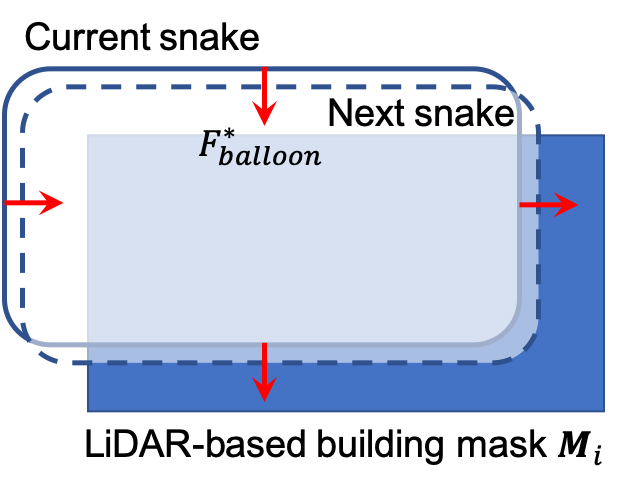}\caption{}\label{sfig:new_balloon}
	\end{subfigure}
	\caption{Illustration of the balloon force on a rectangle building. (\textbf{a}) Original balloon force continuously inflating; (\textbf{b}) Improved balloon force behaviors, adjusted based on the snake local curvature and its relation to the LiDAR-based building mask (blue rectangle). The red arrows represent the balloon force applied to the snake points, moving from the current iteration (solid line) to the next one (dashed line).}% at its next iteration under the influence of the present balloon force.}
	\label{fig:balloon}
\end{figure}

As aforementioned, the classical balloon force  is conceived to constantly push the snake outward based on its local curvature (Equation \eqref{eq:balloon_force}).
Such behavior becomes less relevant when addressing buildings with complex shape.
Therefore, we propose to adapt the balloon force to push outward at some particular region and shrink inward at some others.
Such adaptation is explained in the following.
%We also propose an adaptation to the balloon force from indicators provided by LiDAR data. 
Using the 3-D building boundaries preliminarily extracted from LiDAR point cloud, a set of building masks $ \mathbf{M}_i $  can be created. 
%They are denoted by $ \mathbf{M}_i $ with $ i $ stands for the building index.
Such masks are generated by projecting the 3-D building boundaries onto the image space, and then determining the enclosed region inside the projected boundaries.
Then, the adapted behavior is carried out through a signed magnitude matrix computed using the mask $ \mathbf{M}_i $, as defined in Equation \eqref{eq:new_K}.

\begin{equation}\label{eq:new_K}
	K_i(x,y) = \left\lbrace
	\begin{array}{ll}
	\kappa, & \textnormal{if } (x,y)\in \mathbf{M}_i \\
	-\kappa, & \textnormal{if } (x,y)\notin \mathbf{M}_i
	\end{array}
	\right.
\end{equation}
with $ (x,y) $ is the coordinates of a point on the snake, and \newtextt{$ \kappa $ is the force magnitude weight}.
As a result, the improved balloon force for a building $ i $ is given as in Equation \eqref{eq:balloon_force_new},

\begin{equation}\label{eq:balloon_force_new}
F^*_{\mathrm{balloon},~i}= K_i(x,y) \times \vec {n}(x,y) 
\end{equation}
where $ {\vec  n} $ stands for the normal  vector of the curve at $ (x,y) $. 
%This model mimics the inflation of a balloon by continuously pushing the snake points outward. 
%Thus it prevents the snake to shrink into a single point.

%\modif{As illustrated by Figure \ref{fig:balloon}, the improved balloon force should influence the snake to approach the LiDAR-based building masks.}

Figure \ref{fig:balloon} depicts a schematic representation illustrating the proposed adaptation. 
It demonstrates how the snake behaves differently  (i.e. either inflating or shrinking) based on its relation with the given LiDAR-based building mask.
With the LiDAR-based building masks $ \mathbf{M}_i $ represented by the blue rectangle, the balloon force behavior (represented by the red arrows) is improve to shrink or inflate at different snake points.

\begin{figure}[h]
	\centering
	\begin{subfigure}{0.49\linewidth}
		\centering\includegraphics[trim=0 9cm 24.5cm 0,clip,width=\linewidth]{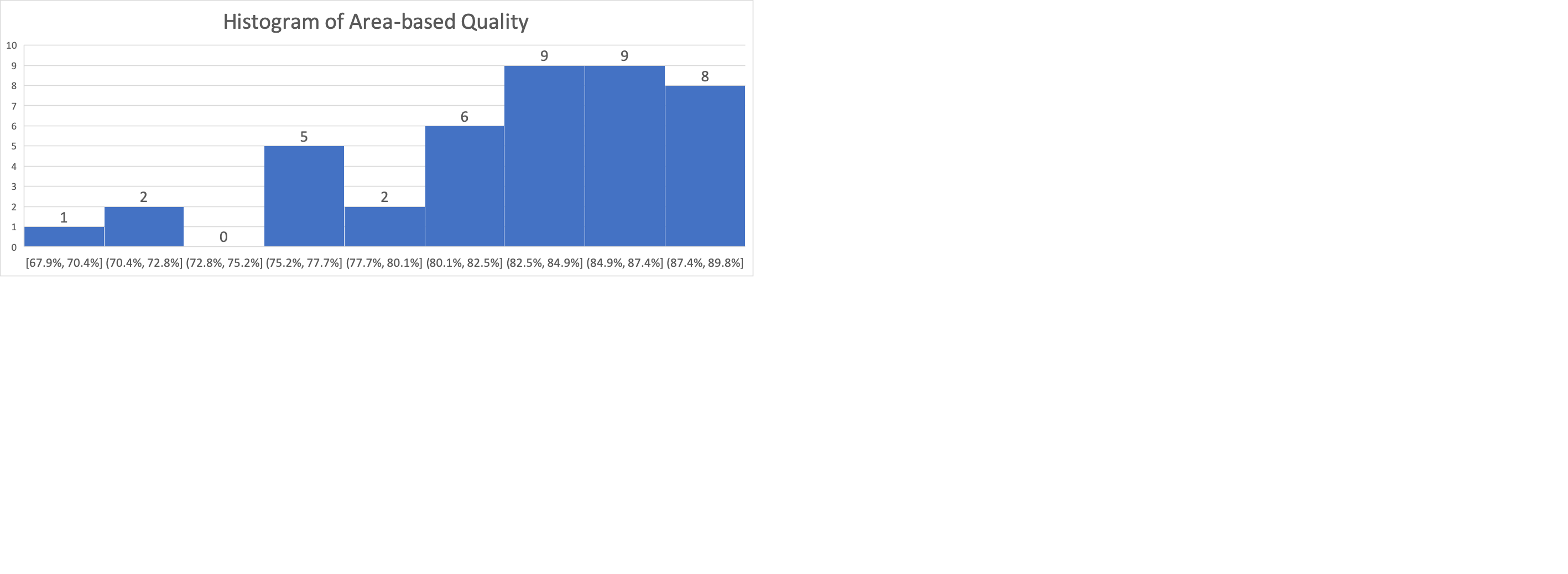}\caption{Area-based \textit{Quality}}\label{sfig:IoU_all_methods}
	\end{subfigure}
	\hspace{0.1cm}
	\begin{subfigure}{0.49\linewidth}
		\centering\includegraphics[trim=0 9cm 24.5cm 0,clip,width=\linewidth]{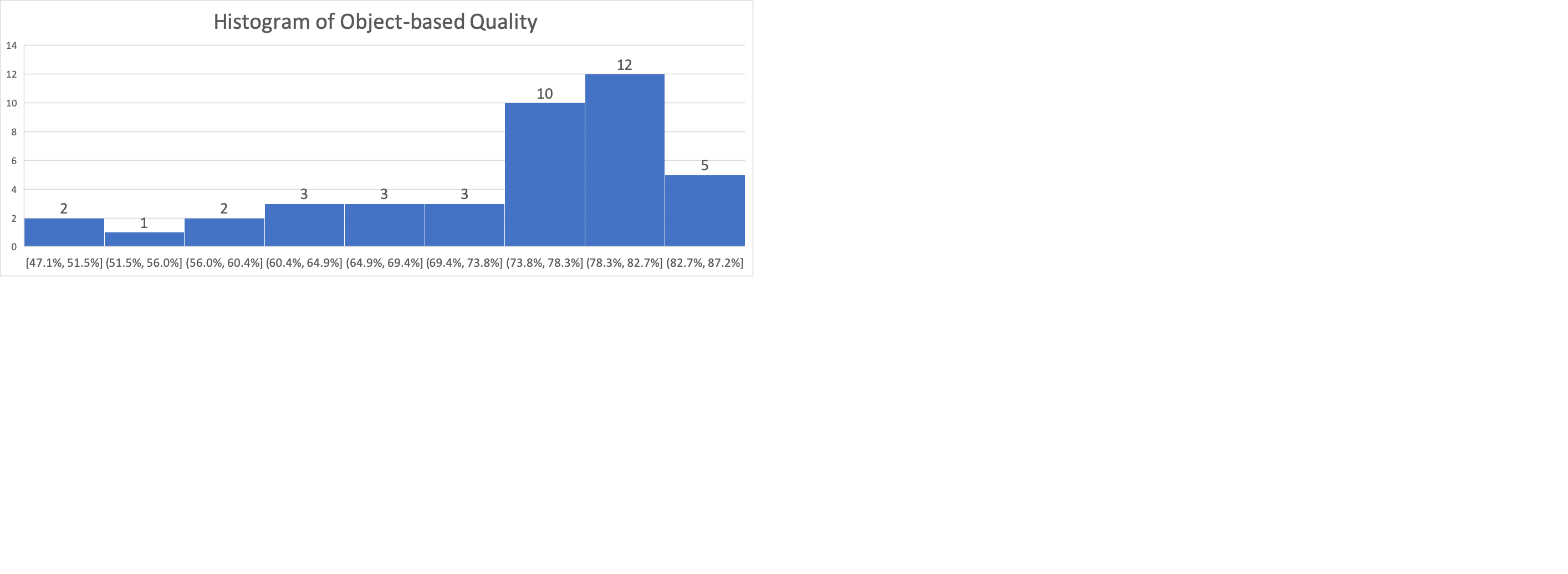}\caption{Object-based \textit{Quality}}\label{sfig:Q_obj_all_methods}
	\end{subfigure}
	
	\vspace{0.25cm}
	\begin{subfigure}{0.49\linewidth}
		\centering\includegraphics[trim=0 9cm 26.5cm 0,clip,width=\linewidth]{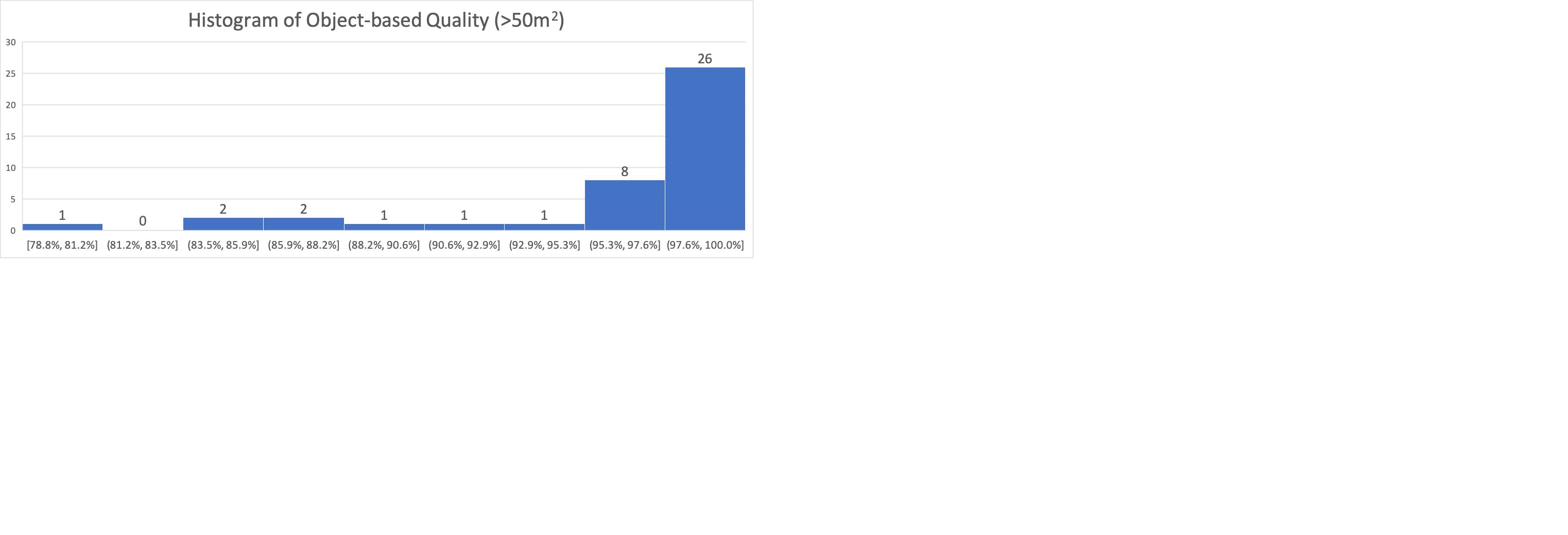}\caption{Object-based \textit{Quality} (larger than 50 m$ ^2 $)}\label{sfig:Q_obj50_all_methods}
	\end{subfigure}
	\hspace{0.1cm}
	\begin{subfigure}{0.49\linewidth}
		\centering\includegraphics[trim=0 9cm 24.5cm 0,clip,width=\linewidth]{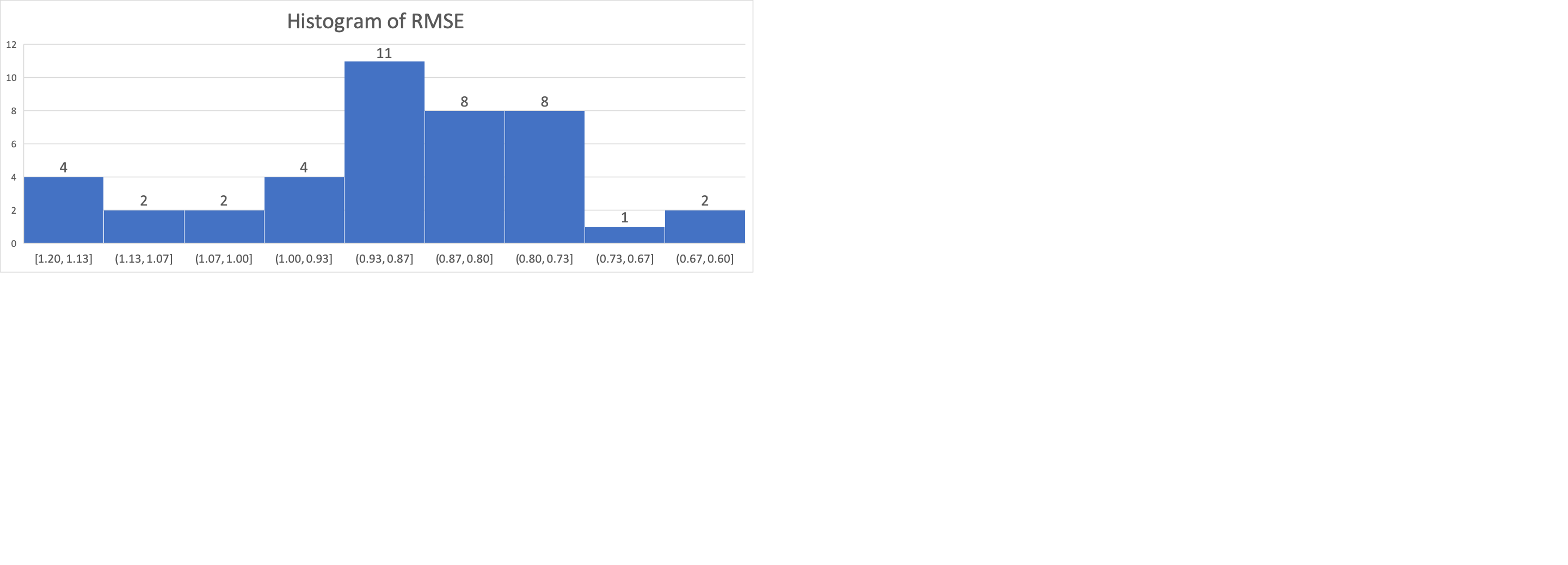}\caption{RMSE}\label{sfig:RMSE_all_methods}
	\end{subfigure}
	\caption{Submitted methods to the ISPRS Vaihingen benchmark dataset, by counting the number of methods divided by the resulting accuracy. (\textbf{a}) Area-based \textit{Quality}; (\textbf{b}) Object-based \textit{Quality}; (\textbf{c}) Object-based \textit{Quality} (larger than 50 m$ ^2 $); (\textbf{d}) RMSE.}
	\label{fig:histograms}
\end{figure}

\input{chap4_result.tex}
The resulting accuracy of the SRSM is then compared with other works submitted to the ISPRS Vaihingen benchmark  portal\footnote{\url{http://www2.isprs.org/commissions/comm3/wg4/results.html}}.
As of January 26th 2020, there are currently 42 submitted methods.
Four histograms are shown by Figure \ref{fig:histograms}, summarizing the resulting accuracy of these methods averaged on the three areas.
Each histogram shows the distribution of methods according to the area-based \textit{Quality} (Figure \ref{sfig:IoU_all_methods}), object-based \textit{Quality} (Figure \ref{sfig:Q_obj_all_methods}), object-based \textit{Quality} for objects larger than 50 m$ ^2 $ (Figure \ref{sfig:Q_obj50_all_methods}), and RMSE (Figure \ref{sfig:RMSE_all_methods}).
All four histogram are presented with their bins sorted in an increasing quality order, i.e. any particular bin involves a higher quality (i.e. higher \textit{Quality} percentage and lower RMSE) than the bins on its left.
Through these histograms, one can  remark the developed consensus on results of the state-of-the-art methods.
For instance, from Figure \ref{sfig:IoU_all_methods}, it is shown that the majority of the methods (i.e. approximately 62\% or 26/42 methods) yield an area-based \textit{Quality} ranging from 82.5\% to 89.8\%. 
On the other hand, 64\% of the methods yield an object-based \textit{Quality} ranging from 73.8\% to 87.2\% (Figure \ref{sfig:Q_obj_all_methods}).
For object-based \textit{Quality} for buildings larger than 50 m$ ^2 $ (Figure \ref{sfig:Q_obj50_all_methods}), a result greater than 97.6\% is desirable, considering that 62\% of the methods are capable of yielding such outcome.

%\vspace{0.5cm}
%\noindent\modif{Note: Cet espace ne sera pas là lors de l'enlèvement de la Table des matières.}

As a fully unsupervised and automatic building extraction method, our method yields very high accuracy.
Indeed, considering the resulting average area-based and object-based \textit{Quality}, respectively 86.57\% and 81.60\%, our method is placed among the top 20\% of all benchmark methods, i.e. the 10$ ^{th} $ or 9$ ^{th} $ among 42 methods.
These results are highly desirable compared to other existing methods.
%Figure \ref{fig:top10} also shows the 10 highest area-based and object-based \textit{Quality} methods.
It is also worth-noting that many among the top-accuracy methods are supervised or model-based methods \cite{rottensteiner2014results}.
Indeed, the supervised methods proposed by Niemeyer \textit{et al.} %(acronym: HANC3) 
\cite{niemeyer2012conditional} and Chai %(acronym: ZJU) 
\cite{chai2016probabilistic}, result in area-based \textit{Quality} of, respectively, 87.8\% and 89.7\%.
The model-based methods proposed by Bayer \textit{et al.} \cite{bayerbrief} yields an area-based \textit{Quality} of 89.8\%, and the two versions of a method by Grigillo \textit{et al.} %(acronym: LJU1 and LJU2) 
\cite{grigillo2012urban} yield an area-based \textit{Quality} of 89.4\% and 89.7\%.
%However, they are strongly scene-dependent, in which many thresholds are set specifically for the Vaihingen scenes and datasets.
%The acronym of other methods can be found in the benchmark results \cite{rottensteiner2014results}.}
In addition, considering the object-based \textit{Quality} for buildings with an area larger than 50 m$ ^2 $, our method is placed 12$ ^{th} $ among 42 methods.
However, considering the RMSE (Figure \ref{sfig:RMSE_all_methods}), our method yields a result (averaging 1.09 m) among the highest RMSE, in other words, the least desirable.
Future works will concentrate on improving such accuracy.

%\begin{figure}[H]
%	\centering
%	\begin{subfigure}{0.48\linewidth}
%		\centering\includegraphics[trim=0 0 0 0,clip,width=\linewidth]{fig/top10_IoU}\caption{Area-based \textit{Quality} methods}\label{sfig:top10_IoU}
%	\end{subfigure}
%	\hspace{0.1cm}
%	\begin{subfigure}{0.48\linewidth}
%		\centering\includegraphics[trim=0 0 0 0,clip,width=\linewidth]{fig/top10_Q_obj}\caption{Object-based \textit{Quality} methods}\label{sfig:top10_Q_obj}
%	\end{subfigure}
%	\caption{The 10 highest \textit{Quality} methods according to the (\textbf{a}) area-based \textit{Quality}; and (\textbf{b}) object-based \textit{Quality}.}
%	\label{fig:top10}
%\end{figure}

%\modif{Comparer la caractéristique/la complexité différente entre les régions (Vaihingen dataset) qui impactent les métriques résultantes finales.}
%
%\modif{Commenter/analyser sur les différents types d'erreur présentantes aux résultats.}

%\modif{Commenter/analyser sur l'amélioration du résultat de cet nouvel article par rapport à l'article PIA.}

%\modif{Comparer la fonctionnement de la méthode proposée sur les différentes villes du Québec; et entre la méthode proposée avec les résultats du Microsoft avec des analyses des points forts et points faibles.}

%\modif{Perspectives sur des  travaux futurs, e.g. concentrer sur comment enlever les végétations plus efficacement, notamment les végétations ombrées (shadowed trees); concentrer sur une parametrage automatique efficace des modele de contours actifs; concentrer sur une propre polygonisation, etc.}
The proposed SRSM also faces several problems when performed on the ISPRS Vaihingen benchmark dataset, such as the  problem of nearby shadowed vegetation shown in Figure \ref{sfig:shadowed_vege}.
Grigillo \textit{et al.} \cite{grigillo2012urban} proposed to solve such problem with the rule-set classifiers on image pixel colors and NDVI. 
However, this approach involves multiple manually selected thresholds which requires a high level of supervision.
%This method yields a very high \textit{Quality}, i.e. 89.7\%.
%However, this method is strongly scene-dependent, in which many thresholds are set specifically for the Vaihingen scenes and datasets.
%Consequently, it requires a manual resetting of these thresholds on a new urban scene.
There also exists other classification approaches (in order to better classify shadowed trees from buildings) involving graph-cut-based method \cite{ok2013automated}.
However, such method may require a high amount of a priori information or user inputs in order to yield accurate results \cite{szeliski2010computer}.
Therefore, by opting for such mentioned approaches, the level of supervision of the building extraction method should be reconsidered.

\subsection{Performance on Quebec City}
\begin{figure}[!h]
	\centering
	\begin{subfigure}[b]{0.495\linewidth}
		\centering\includegraphics[trim=2cm 1cm 2cm 0,clip,height=7cm]{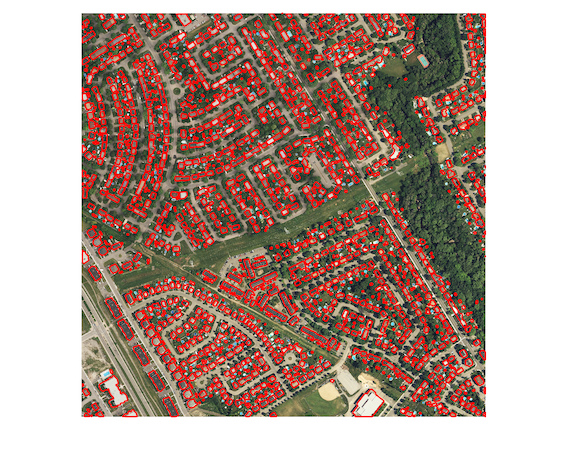}\caption{}\label{sfig:4190}
	\end{subfigure}\hfill
	\begin{subfigure}[b]{0.495\linewidth}
		\centering\includegraphics[trim=3cm 3cm 3cm 0,clip,height=7cm]{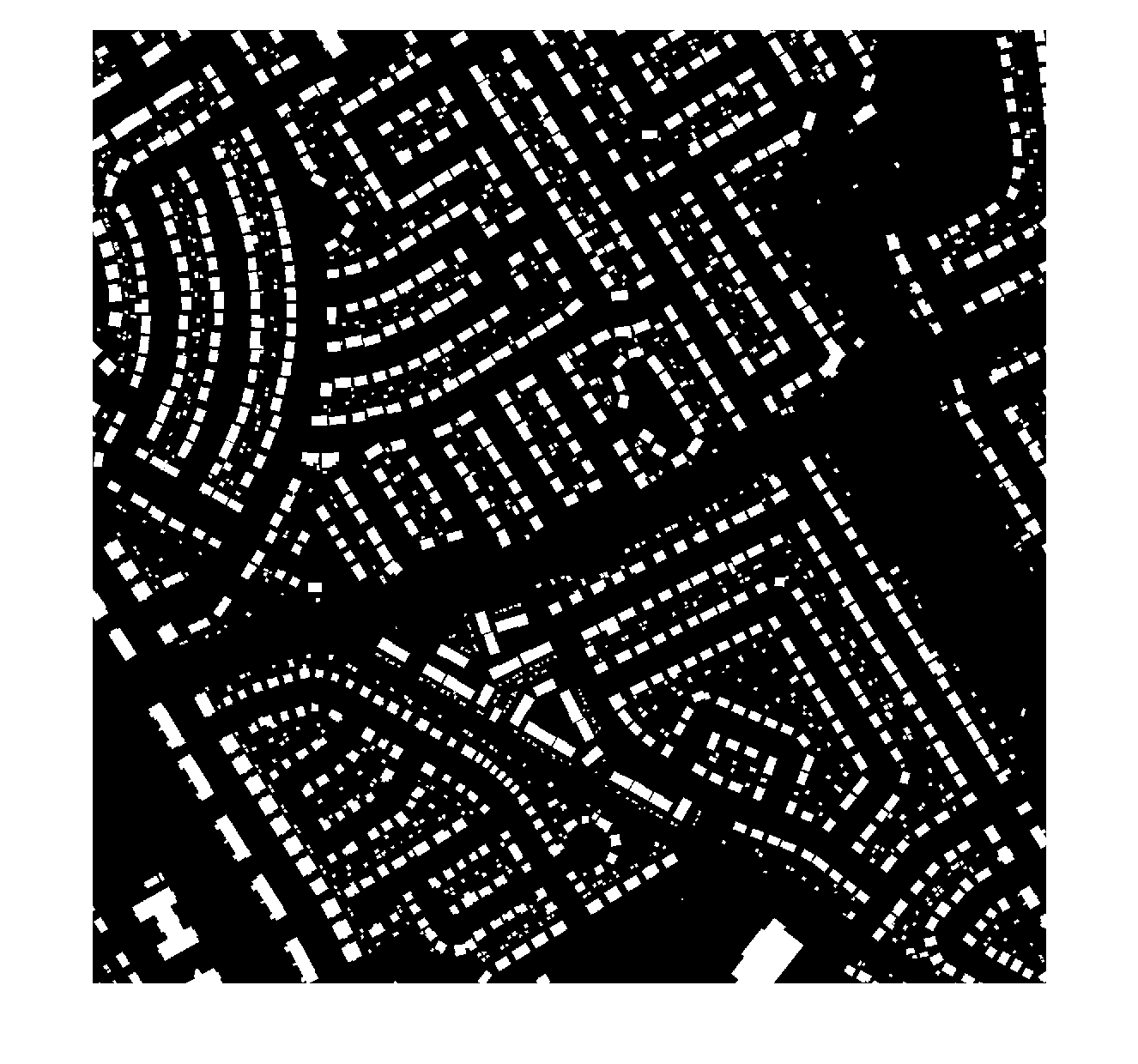}\caption{}\label{sfig:4190_gt}
	\end{subfigure}

	\vspace{0.5cm}
	\begin{subfigure}[b]{0.495\linewidth}
		\centering\includegraphics[trim=2cm 1cm 2cm 0,clip,height=7cm]{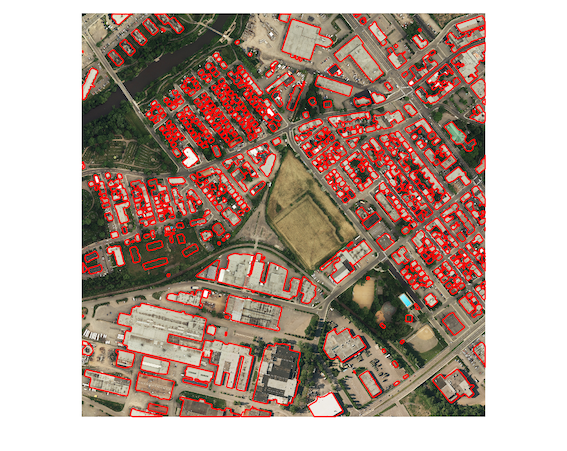}\caption{}\label{sfig:4785}
	\end{subfigure}\hfill
	\begin{subfigure}[b]{0.495\linewidth}
		\centering\includegraphics[trim=3cm 3cm 3cm 0,clip,height=7cm]{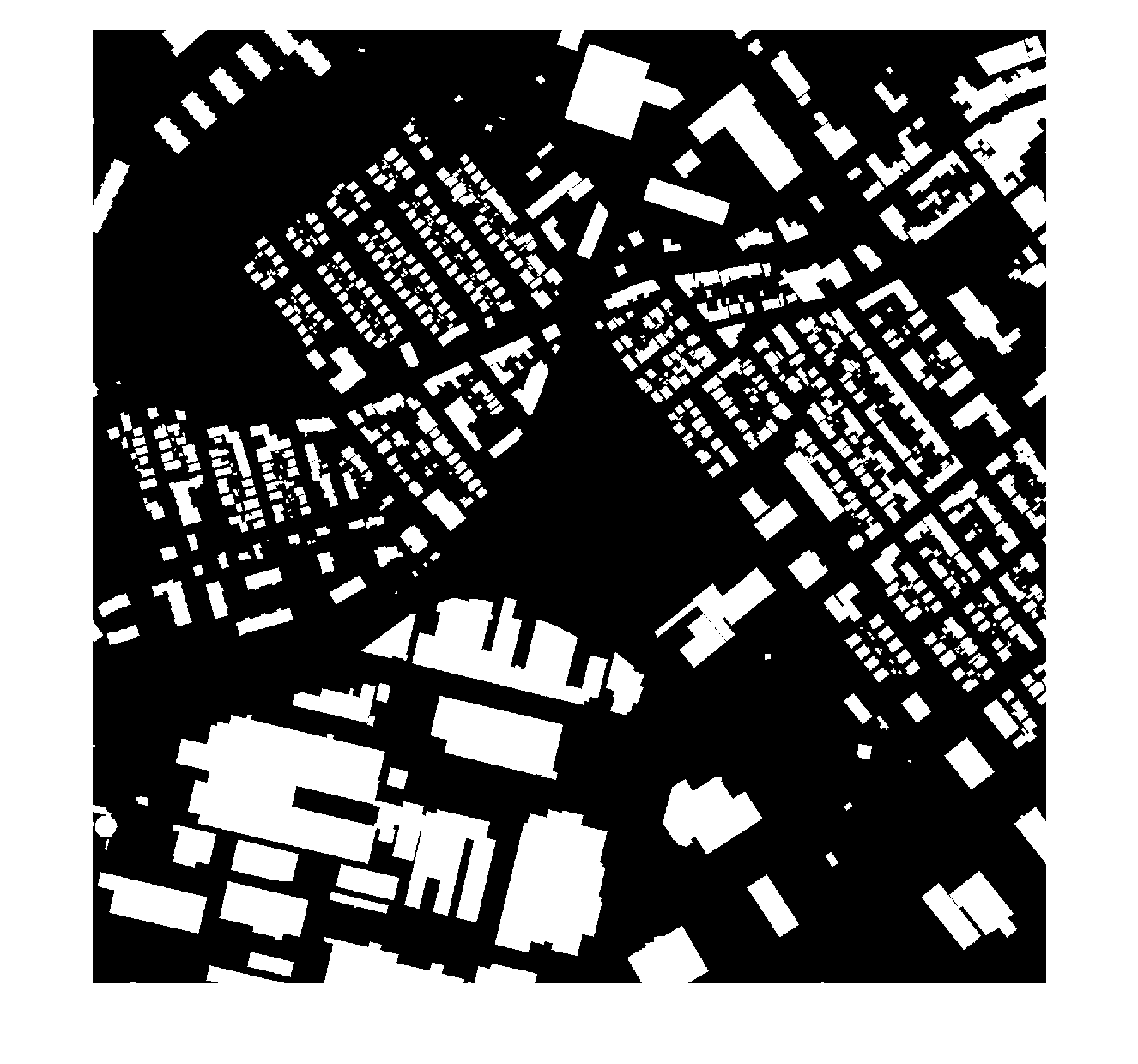}\caption{}\label{sfig:4785_gt}
	\end{subfigure}
	\caption{SRSM results (in red outlines) on typical urban and residential areas in Quebec City, and the corresponding ground truth. Each example covers a 1 km $ \times  $1 km area.}
	\label{fig:qc_snake}
\end{figure}

%\modif{To-be-available in a later version. Tests on Quebec City will be started in the next week. We are currently waiting for the Repentigny (both LiDAR and imagery datasets) and the Montreal (orthoimage dataset).}
\newtextt{In order to test the performance and applicability of the proposed SRSM on a large scale, we carry it out on the Quebec City dataset.
Many areas on Quebec City are composed of different types of urban, residential and industrial scenes. 
Two of these typical scenes are shown in Figure \ref{fig:qc_snake}.
They are also representative of the North American context.
Based on a visual assessment, the SRSM succeeds at delineating accurately the building boundaries on the two exemplified scenes.
\newtexttt{Typically, the size of the buildings shown in both scenes varies greatly from small to very large buildings.
One can remark that many buildings have similar color as their background (i.e. parking lots, open areas, etc.) are also well delineated.
Other optical image-related problems such as roof objects, nearby cars are also avoided.
This reemphasizes the benefits of using the $ z $-images encoding LiDAR elevation data instead of the optical images.
In addition, similar to the Vaihingen datasets (particularly area 1 and 2), the shape of buildings presented in these two examples---also verified across the whole Quebec City area---can be very complex.
These three factors related to the scene complexity---i.e. varying building size, color, and shape---can be problematic to other methods, whereas the proposed SRSM is able to overcome such complexity.}
}

Table \ref{tab:QC_accuracy_pixel} summarizes the area-based and object-based accuracy yielded by the Microsoft open Canada building footprints, and the proposed SRSM.
\newtexttt{It can be noted that the \textit{Completeness} and \textit{Correctness} yielded by the two methods are quite different.
These differences mainly stem from the fact that the two methods were carried out using different data sources with different characteristics.}
However, based on the resulting \textit{Quality} values reflecting the overall accuracy, it can be noted that the SRSM provides a competitive outcome compared to the Microsoft method.
Indeed, the \textit{Quality} margins between the SRSM and the Microsoft method are well balanced.
\newtexttt{The SRSM yields a 6.65\% higher object-based \textit{Quality}, while in contrast, the Microsoft method provides a 7.40\% higher area-based \textit{Quality}.
%The SRSM reaches a 6.65\% higher Quality when considering the object-based representation, while Microsoft reaches a 7.40\% higher Quality when considering the area-based representation
On the one hand, the difference of the area-based \textit{Quality} stems from the fact that the resulting footprints from SRSM have the tendency to be slightly \textquotedblleft rounded\textquotedblright~around the building corners. 
Whereas the Microsoft footprints were generated (with their own polygonization method) without such problem.
%As a result, this contributes to the higher area-based \textit{Quality} for the Microsoft footprints.
On the other hand, the SRSM with the advantage of the $ z $-images encoding elevation data allows to detect the buildings more precisely, hence yielding the higher  object-based \textit{Quality}.
Nevertheless, it is always worth-noting that such competitive %large-scale building extraction 
accuracy is produced by an unsupervised approach, compared to the heavily supervised approach from Microsoft which was trained on three million labeled images.}
The complete dataset of extracted building boundaries in Quebec City by the SRSM, as well as the high-resolution version of  Figure  \ref{fig:qc_snake}, are made publicly available at {\normalsize \texttt{\url{https://github.com/nthuy190991/SRSM_QuebecCity_building_extraction}}}.

\begin{table}[H]
	\centering
	\caption{Area-based and object-based accuracy of the SRSM on the Quebec City dataset, compared with the Microsoft open Canada building footprints.}
	\label{tab:QC_accuracy_pixel}
	\begin{tabular}{cccccccc}
		\toprule
		& \multicolumn{3}{c}{\textbf{Area-based accuracy}} && \multicolumn{3}{c}{\textbf{Object-based accuracy}} \\
		\cline{2-4}\cline{6-8}\\[-7pt]
		\textbf{Method} & $ \mathbf{Cp} $ & $ \mathbf{Cr} $ & $ \mathbf{Q} $ && $ \mathbf{Cp} $ & $ \mathbf{Cr} $ & $ \mathbf{Q} $ \\\midrule
		Microsoft building footprints & 77.42 \% & 87.61 \% & 69.77 \% && 59.01 \% & 93.16 \% & 56.56 \% \\
		SRSM footprints & 82.32 \% & 72.02 \% & 62.37 \% && 74.25 \% & 80.95 \% & 63.21 \%\\
		\bottomrule
	\end{tabular}
\end{table}

\newtexttt{The outcomes of the SRSM  on the Quebec City dataset are relevant, visually and quantitatively. 
%As an unsupervised method, it yields competitive results compared to the ones provided by a heavily-supervised approach (by deep learning)---the Microsoft open Canada building footprints.
However, there still remain two issues. % of the SRSM.
Firstly, from a practical perspective, the SRSM was carried out separately on tiles (Figure \ref{fig:qc_lidar_coverage}) for the sake of processing time and memory constraint.
Then, the tile-based results were combined in QGIS. 
Such step is crucial for the buildings located in the transitioning areas between two neighboring tiles.
Several of those buildings can be identified near the borders of the tiles shown in Figure \ref{fig:qc_snake}.
%Future works will concentrate on developing an end-to-end building extraction method including these transitioning areas.
Secondly, the SRSM is unable to separate connected or nearby buildings with similar height. 
Given the $ z $-images involves only elevation information, such a separation task can be difficult. 
Therefore, we shall investigate the usefulness of other information for such task.
Overall, these two issues can affect unfavorably the resulting accuracy of the SRSM.
%As a matter of fact, the accuracy level of the SRSM results---even though exhibits the competitive level with the Microsoft footprint results---is much less desirable than the results obtained in the ISPRS Vaihingen dataset. 
%Even though this can be expected with such a large-scale dataset, 
Future efforts will concentrate on addressing these two issues to improve the SRSM results. % on such a large-scale dataset.
}

%Authors should discuss the results and how they can be interpreted in perspective of previous studies and of the working hypotheses. The findings and their implications should be discussed in the broadest context possible. Future research directions may also be highlighted.

%%%%%%%%%%%%%%%%%%%%%%%%%%%%%%%%%%%%%%%%%%%
%\section{Materials and Methods}
%
%Materials and Methods should be described with sufficient details to allow others to replicate and build on published results. Please note that publication of your manuscript implicates that you must make all materials, data, computer code, and protocols associated with the publication available to readers. Please disclose at the submission stage any restrictions on the availability of materials or information. New methods and protocols should be described in detail while well-established methods can be briefly described and appropriately cited.
%
%Research manuscripts reporting large datasets that are deposited in a publicly available database should specify where the data have been deposited and provide the relevant accession numbers. If the accession numbers have not yet been obtained at the time of submission, please state that they will be provided during review. They must be provided prior to publication.
%
%Interventionary studies involving animals or humans, and other studies require ethical approval must list the authority that provided approval and the corresponding ethical approval code. 

%%%%%%%%%%%%%%%%%%%%%%%%%%%%%%%%%%%%%%%%%%
\section{Discussions}\label{sec:discussion}
In this Section, three discussions are drawn to the attention: (\textit{i}) on the relevance of the proposed SR,  (\textit{ii}) on the SRSM results, and (\textit{iii}) on the impact of the snake model parametrization.%. First, we discuss on the relevance of the proposed SR process in the building extraction topic as well as some others. Then, we dis
\subsection{Relevance of the super-resolution}
\newtexttt{As suggested by the name of the proposed method (i.e. SRSM), the SR process plays a critical role.
However, such process is not only relevant for snake models.
Indeed,  the need and potential of such process to enhance the spatial resolution of LiDAR data is high. 
For instance, in the topic of building extraction, several methods \cite{griffiths2019improving,huang2019automatic} proposed to replace the blue channel of RGB images with a normalized DSM (nDSM).
Such composite image---i.e. red, green, and nDSM---are then feed into deep neural networks for extracting buildings.
However, these approaches did not account for the fact that the two input images---the RGB image and the nDSM---usually have different resolution, hence a SR was not proposed.
On the other hand, the SR process could resolve one of the problems of the snake model proposed by Kabolizade \textit{et al.} \cite{kabolizade2010improved} (cf. sub-section \ref{ssec:review_snake_model}).
A super-resolved DSM could improve the height variance-based external energy term proposed in their work. 
However, it is worth-noting that the main drawback of their snake model is still the use of optical image as the target image, i.e. for computing the $ E_{\mathrm{img}} $.
In other topics, the study of SR applied to LiDAR depth measurements is also very active.
Indeed, a reliable SR would benefit many applications, such as calibration for autonomous driving \cite{castorena2018motion}, or land cover classification \cite{luo2015fusion}.
}

\subsection{Discussion on the SRSM resulting footprints}
\newtexttt{The accuracy level of the SRSM results carried out on the Vaihingen dataset and the Quebec City dataset have been shown---through multiple assessments and comparisons---to be desirable.
It has achieved our objectives for a large-scale high-accuracy building extraction method, without any assumptions on the building characteristics, nor any training data.
However, two important aspects concerning the building footprints provided by the proposed method should be discussed.
First, it can be noted from the results in Vaihingen (Figure \ref{fig:isprs_areas}) and Quebec City (Figure \ref{fig:qc_snake}) that, the resulting snakes have the tendency to be slightly \textquotedblleft rounded\textquotedblright~around building corners. 
Such problem can be addressed with an efficient polygonization method. 
However, such step can be quite challenging considering the complexity of building shape on the two study areas.}

\newtexttt{The second worth-mentioning aspect involves the acquisition time difference between the LiDAR data, the optical image data, and the reference ground truth boundaries.
On the one hand, considering a benchmark dataset like the ISPRS Vaihingen dataset, such aspect is minimal since the data were acquired almost concurrently (cf. Table \ref{tab:datasets_ISPRS}). In addition, the Vaihingen ground truth building boundaries were prepared using the same data.
On the other hand, considering the large scale of Quebec City, such temporal aspect is much more complicated.
Firstly, the LiDAR data were acquired one year after the optical images.
Secondly, the \textit{Empreintes des bâtiments} dataset consisting of the ground truth building boundaries was produced using multiple different sources, and updated monthly.
Thirdly, the comparative Microsoft results were carried out using Bing Imagery data. 
Since Bing Imagery is a composite of multiple sources, we are unable to determine the exact dates for individual pieces of data\footnote{\url{https://github.com/microsoft/CanadianBuildingFootprints}}.
Such temporal difference and uncertainty can affect the building extraction accuracy.
This issue requires a dedicated study in order to account for all of the involved factors.
}

\subsection{Impacts of snake parametrization}\label{ssec:snake_param}
A snake model involves a number of parameters, such as $ \alpha, \beta $, $\kappa$ (the balloon force magnitude), $\mu_{\mathrm{GVF}} $ (the GVF smoothing parameter), etc.
In the existing models %from an imagery dataset
\cite{peng2005improved,kabolizade2010improved,ahmadi2010automatic}, these parameters have been  set empirically in order to extract buildings effectively. 
%\modif{However, several of them are more important than the others.}
The snake parametrization becomes extremely difficult over a large extended area. 
However, some parameters are more important than the others.
In this regard, %such problem has been partially addressed by the CNN-based approach proposed by 
Marcos \textit{et al.} \cite{marcos2018learning} partially addressed such problem with a CNN-based approach.
It involves learning
%It involves using a CNN to learn 
the characteristics of the most important elements of the snake model, % from training optical images and associated ground truth polygons. 
%Indeed, the most important elements have been addressed,
namely the snake internal energy term weights ($ \alpha $ and $ \beta $), the image-based energy term ($ E_\mathrm{img} $), and the balloon force ($ F_\mathrm{balloon} $).
Additionally, they asserted that one scalar value of $ \beta $ for all parts of a building can lead to problems of over-smoothing at building corners, and under-smoothing at other regions. 
To avoid such problem, they proposed a local penalization approach, by assigning a different $ \beta $ penalization to each pixel depending on whether the pixels are near the building edges or corners, whereas $ \alpha $ remains scalar for every pixel.
In this discussion, let us analyze the relevance of such parametrization approach, \newtextt{and compare with our fixed parametrization for the SRSM.}
%%	present a \modif{work-around/an adaptation} %for snake parametrization 
%%	built upon the work by Marcos \textit{et al.}}
%Let us consider the characteristics of the snake energy terms and parameters involved by the CNN-based approach.
The characteristics of the CNN-inferred energy terms and parameters differ with respect to the features from the optical image (e.g. building corners, edges, etc.), as summarized by Table \ref{tab:params_characteristics}. 
%It is also worth mentioning that $  E_\mathrm{img} $ can have either positive or negative values, whereas $ F_\mathrm{balloon} \ge 0 $ and $ \beta \ge 0 $.

\begin{table}[h]
	\centering
	\caption{Characteristics of the  CNN-inferred balloon force term $ F_\mathrm{balloon} $, image-based energy term $  E_\mathrm{img} $ and snake curvature weight $ \beta $ among the optical image features (resulted by \cite{marcos2018learning}). $  E_\mathrm{img} $ can have either positive or negative values, whereas $ F_\mathrm{balloon} \ge 0 $ and $ \beta \ge 0 $.}
	\label{tab:params_characteristics}
	\begin{tabular}{cccc}
		\toprule 
		& \multicolumn{3}{c}{\textbf{CNN-inferred energy terms and parameter}} \\
		\cline{2-4}\\[-8pt]
		\textbf{Feature} & $ \mathbf{F_\mathrm{\mathbf{balloon}}} $ & $ \mathbf{E_\mathrm{\mathbf{img}}} $  & $ \beta $  \\ 
		\midrule 
		Corner  & very positive & very negative & almost 0 \\ 
		\midrule 
		Edge  & very positive & negative & very positive  \\ 
		\midrule 
		Inside boundary & positive  & positive & low but positive\\ 
		\midrule 
		Outside boundary & 0  & positive & low but positive\\ 
		\bottomrule 
	\end{tabular} 
\end{table}

%It involves three aspects: the balloon force $ F_\mathrm{balloon} $, the image-based energy term $  E_\mathrm{img} $, and the curvature weight $ \beta $.
Firstly, %let us analyze the behavior of the balloon force inferred by the CNN.
concerning the balloon force, %there are significant differences between the CNN-based approach and the classical approach.
%%it is predicted based on the learning from training data with ground truth.
%%On the other hand, the classical balloon force model is computed only based on the local curvature of the snake. %, and is unaware of the building-of-interest.
%In the CNN-based approach, 
the second column of Table \ref{tab:params_characteristics} shows the characteristics of the balloon force inferred by the CNN.
If a snake is initialized inside a building boundary, the balloon force---being positive---will inflate it outward until it reach the building corners and edges. 
%where the force value is very high. 
Then, the balloon force sharply drops to zero and remains zero right outside the building boundary, which means that the snake is not allowed to inflate anymore.
However, if the snake is provided with initial points outside the building boundary, the balloon force---being null-valued---is unable to shrink inward to approach the building true boundaries.
Such behavior is not optimal.
In contrast, the approach to generate $ F_\mathrm{balloon} $ proposed in this paper based on the LiDAR-based building mask is more relevant.
It allows the snake to be shrank or inflated adaptively, regardless where it is initialized without relying on any learning process.

\begin{figure}[H]
	\centering
	\begin{subfigure}{0.3\linewidth}
		\centering\includegraphics[trim=10cm 9cm 10cm 4cm, clip, height=3cm]{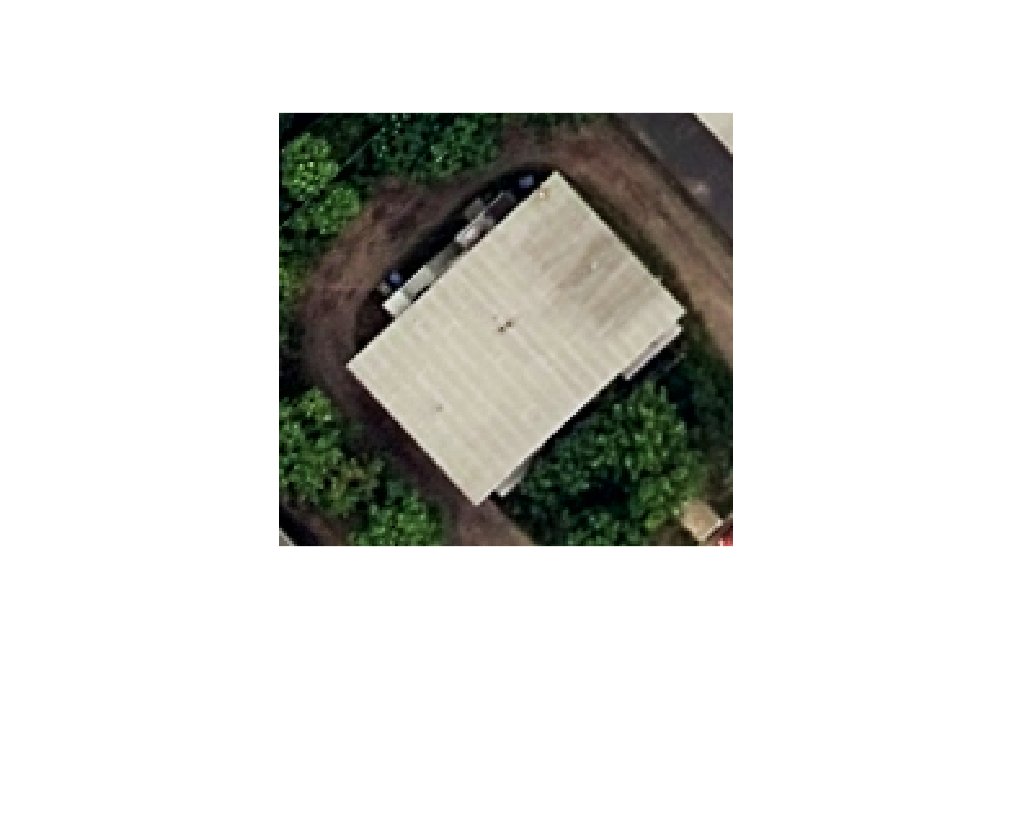}\caption{}\label{sfig:opt_rectangular}
	\end{subfigure}
	\begin{subfigure}{0.3\linewidth}
		\centering\includegraphics[trim=5cm 3cm 2.95cm 2.8cm, clip, height=3cm]{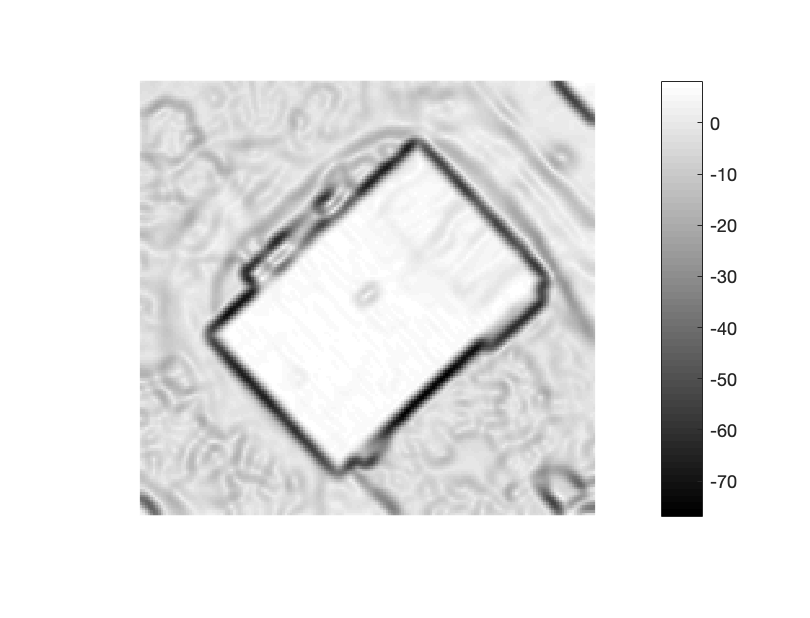}\caption{}\label{sfig:E-img}
	\end{subfigure}
	\caption{Image-based energy term $  E_\mathrm{img} $ of a rectangular building with a color-consistent roof.}
	\label{fig:E_img}
\end{figure}

Secondly, we address the image-based energy term $  E_\mathrm{img} $.
The characteristics of the CNN-inferred image-based energy term $  E_\mathrm{img} $ are revealed in Table \ref{tab:params_characteristics}. 
However, they are similar to those exhibited by the traditional snake model mathematical approach (cf. Equation \eqref{eq:E_img}). 
%Indeed, these characteristics of the corners and edges stem from the energy terms related to the edges and the terminations, i.e. respectively $ E_{edge} $ and $ E_{term} $, in the image-based energy term $  E_\mathrm{img} $.
Such similarity is illustrated by Figure \ref{fig:E_img}.
The energy term $ E_\mathrm{img} $ of a rectangular building (Figure \ref{sfig:opt_rectangular}) is mathematically computed and shown in Figure \ref{sfig:E-img}. 
As illustrated,  the building edges and corners exhibit negative to very negative values, whereas the pixels inside and outside of the building exhibit positive values.
Such characteristics among building features are analogous to the CNN-based approach.
Therefore, in the proposed SRSM, the energy term $  E_\mathrm{img} $ is retained as in the traditional snake model.
The sole change is that the target image %(i.e. from which the energy term $ E_\mathrm{img} $ is computed) 
is the $ z $-image instead of the optical image. 
Such change is relevant as the $ z $-image-based energy term provides more desirable features---i.e. height changes instead of color changes.

\begin{figure}[H]
	\centering
	\begin{subfigure}{0.3\linewidth}
	\centering\includegraphics[trim=6cm 8.5cm 2.95cm 2.75cm, clip, height=3.5cm]{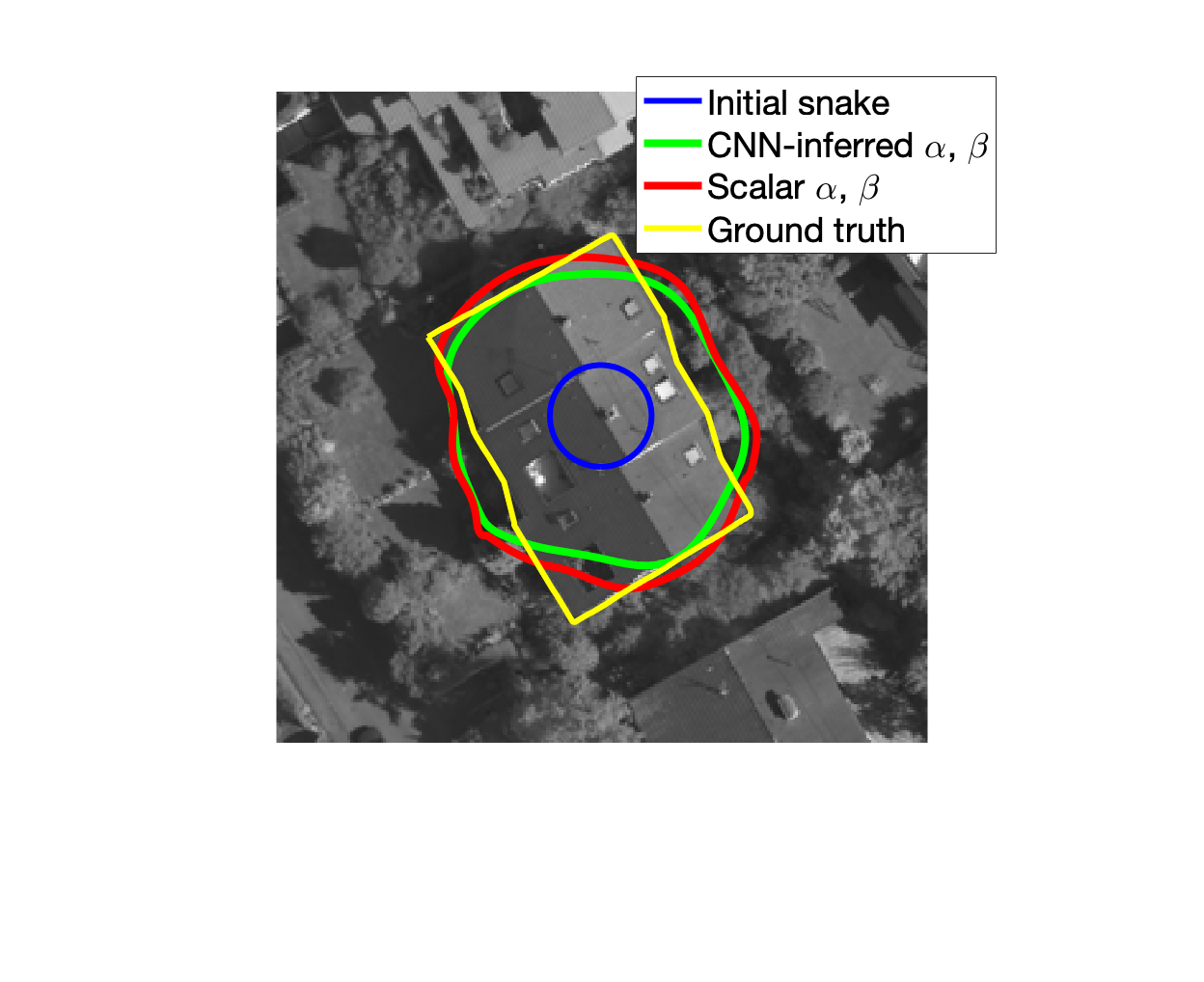}\caption{Snake results}\label{sfig:compare_snakes}
	\end{subfigure}
	\begin{subfigure}{0.3\linewidth}
		\centering\includegraphics[trim=5cm 3cm 3cm 2.8cm, clip, height=3.5cm]{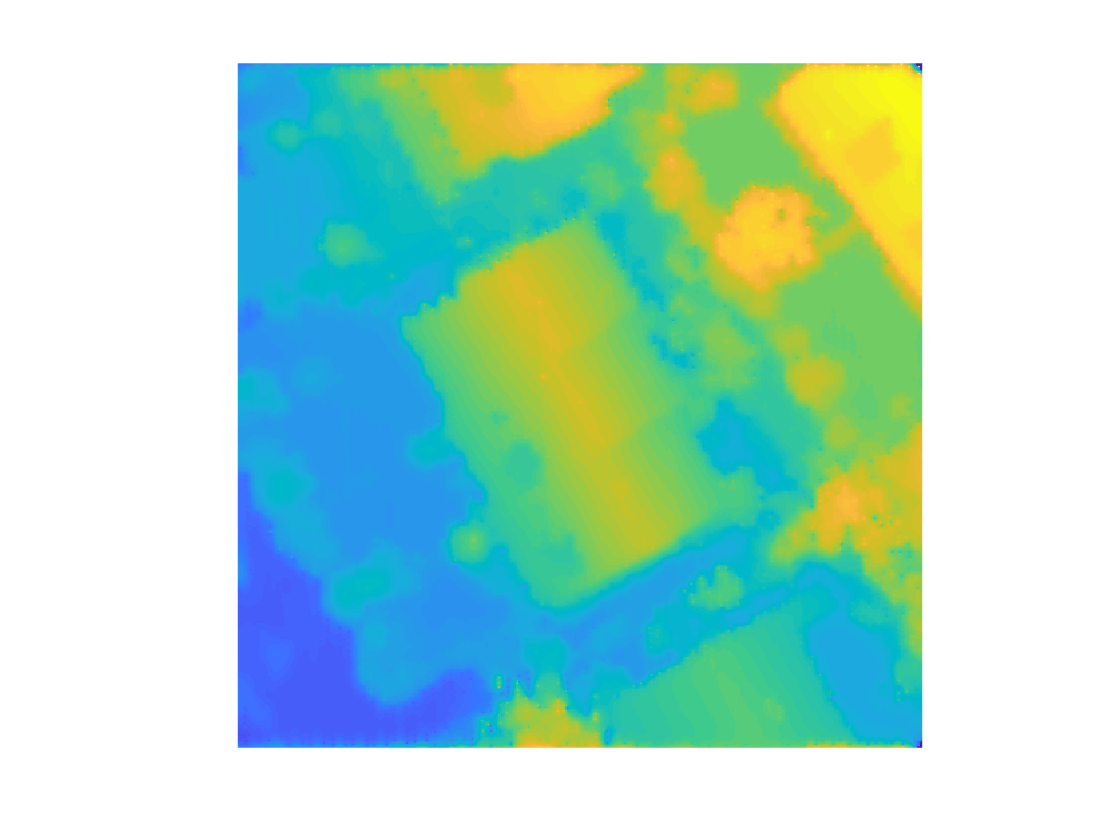}\caption{$ z $-image}\label{sfig:z-image-building49}
	\end{subfigure}
	\begin{subfigure}{0.3\linewidth}
		\centering\includegraphics[trim=5cm 3cm 2.95cm 2.cm, clip, height=3.5cm]{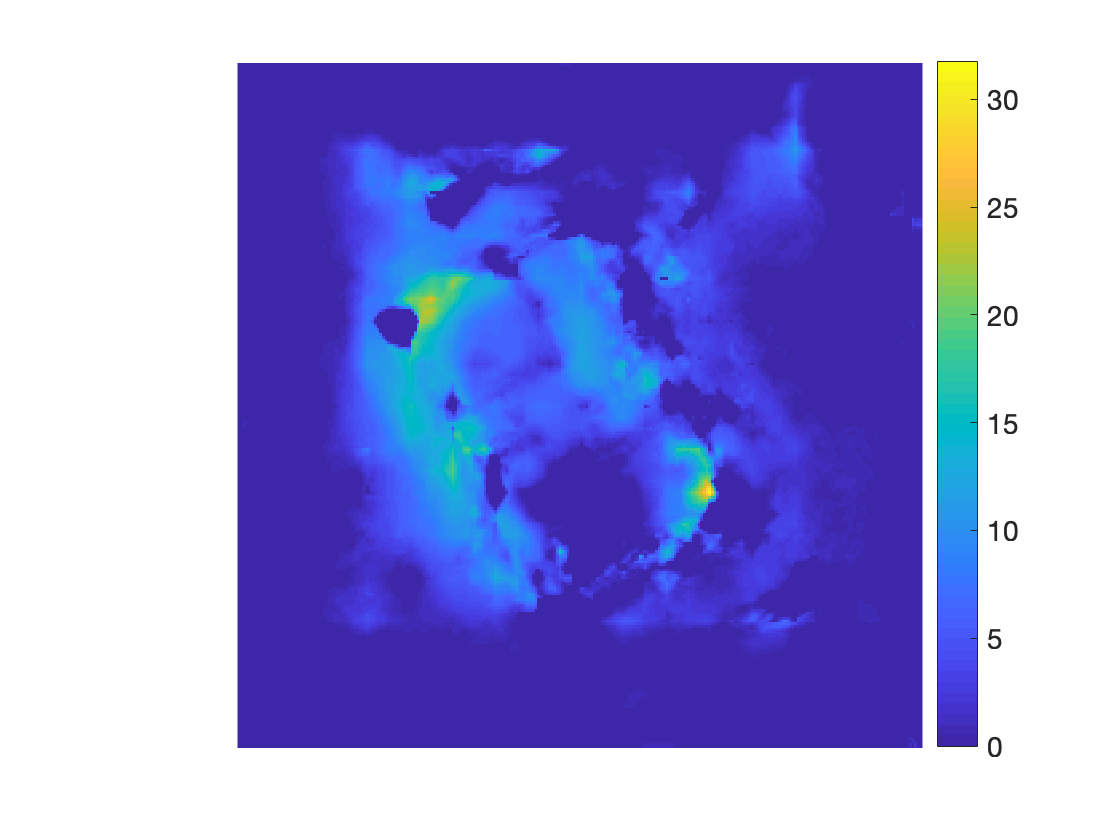}\caption{CNN-inferred $ \beta $}\label{sfig:pw_beta}
	\end{subfigure}
	\caption{Comparison between the use of the CNN-inferred $ \alpha $ and $ \beta $ and the fixed scalar values. (\textbf{a}) Snake results parametrized by CNN-inferred values compared with fixed scalar values; (\textbf{b}) The $ z $-image used in the snake model; (\textbf{c}) The pixel-wise $ \beta $ resulted from CNN \cite{marcos2018learning}.}
	\label{fig:pixelwise_vs_scalar}
\end{figure}

Lastly, concerning the snake curvature weight $ \beta $, we retain the use of a fixed scalar $ \beta $ in our method. % because of two reasons.
The immediate reason is that without a training phase, the generation of a different $ \beta $ value for each pixel is difficult, or even virtually impossible.
In addition, as we changed the target image of the snake model, the needed dynamics for $ \beta $ should also change.
Since the only sources of attraction for the snake model are now the height changes from off-terrain objects, the snake curvature does not need to be different pixel to pixel. % as it is required when using the optical image. 
The snake should be able to correct itself from such sources of attraction. 
%Therefore, the curvature of the snake does not need to be regularized by a different $ \beta $  at each pixel.
%
%In order to consolidate the foundation of such expectation, 
A comparison is conducted to confirm whether using the CNN-inferred pixel-wise $ \beta $ would bring a real benefit compared with a fixed scalar $ \beta $.
As such, the SRSM is experimented where the value of $ \alpha $ and $ \beta $ are, either inferred from CNN as in \cite{marcos2018learning} or set to fixed scalar values.
%Firstly, they are inferred from CNN using the method proposed by \cite{marcos2018learning}.
%Secondly, they are fixed with scalar values.}
\newtextt{Such comparison is carried out on seven buildings in the proximity of the area 1 (ISPRS benchmark dataset) selected by Marcos \textit{et al.}  \cite{marcos2018learning}.
One of these buildings is exemplified in Figure \ref{fig:pixelwise_vs_scalar}. }
The optical image and the initial points for SRSM in blue are revealed in Figure \ref{sfig:compare_snakes}, 
whereas the $ z $-image is shown in Figure \ref{sfig:z-image-building49}.
The SRSM carried out with the CNN-inferred $ \alpha $ and $ \beta $ results in the building boundary in green (Figure \ref{sfig:compare_snakes}). 
The CNN-inferred value of $ \alpha $  is 0.767, whereas the image of $ \beta $ values \newtextttt{(each pixels with a different $ \beta $ value)} is shown by Figure \ref{sfig:pw_beta}. 
%As summarized in Table \ref{tab:params_characteristics}, we can see here that t
%Such matrix exhibits near-zero values of $ \beta $ at the building corners, high values at the building edges, and low values inside and outside the building boundary.
Then, the SRSM carried out using the scalar $ \alpha $ and $ \beta $---both set equal to 0.2---yields the red building boundary (Figure \ref{sfig:compare_snakes}).
%It is also worth-noting that the selected building is from the training dataset, not the test dataset.}}
The two snakes \newtextttt{in red and in green} are shown to be  similar.
%
%In order to provide a quantitative comparison between the two snakes, we also measure the overall area-based \textit{Quality}.
%Such metric is used for evaluating building extraction accuracy, which will be described in sub-section \ref{ssec:metric}.
Quantitatively, %the green boundary (resulted with the CNN-based approach) yields an area-based \textit{Quality} of 73.6\%, whereas the red boundary (with fixed scalar  parametrization approach) yields 76.5\%.
%Table \ref{tab:compare_fixed_beta_vs_DSAC_beta} shows the average area-based \textit{Quality} on all seven buildings.
the area-based \textit{Quality} provided by the CNN-based approach on all seven buildings averages 73.62\%, whereas the fixed scalar parametrization approach yields 72.03\%.
%In other words, the two approaches yield the respective area-based \textit{Quality} with only 1.59 \% of difference.}
By visual and quantitative assessment, it is shown that the CNN-inferred approach as well as the pixel-wise $ \beta $ does not bring a practical benefit to our SRSM.
%Nevertheless, it requires an enormous amount of computational cost due to the learning step.
%On the contrary, such CNN-based approach requires an enormous amount of computational cost due to the learning step. %\cite{marcos2018learning}.
%Therefore, it has been shown that such pixel-wise approach is not optimal, and as a result, the two weights of the snake internal energy are maintained scalar with fixed values in the proposed SRSM.
%The approach using fixed scalar  $ \alpha $ and $ \beta $ is shown to be relevant.

%\begin{table}[H]
%	\centering
%	\caption{Average area-based \textit{Quality} yielded by SRSM with the CNN-inferred $ \alpha $ and $ \beta$ compared to the fixed scalar   $ \alpha $ and $ \beta $.}
%	\label{tab:compare_fixed_beta_vs_DSAC_beta}
%	\begin{tabular}{cc}
%		\toprule 
%		& \textbf{Average area-based \textit{Quality}} \\ \midrule 
%		\textbf{CNN-inferred $ \alpha $ and $ \beta $} & 73.62 \% \\
%		%& 81.50 \% &  77.43 \% &  65.30 \% &  77.21 \% &  63.89 \% &  69.86 \% &  80.14 \%\\
%		\textbf{Fixed scalar $ \alpha $ and $ \beta $} & 72.03 \% \\
%		%& 88.68 \% &  77.99 \% &  59.04 \% &  61.13 \% &  68.69 \% &  69.52 \% &  79.18 \%\\
%		\bottomrule 
%	\end{tabular} 
%\end{table}

%Moreover, the characteristics of building appearances on their respective $ z $-image should be all similar, since almost every building exhibits a strong elevation variation with respect to its surrounding area.
%Thus the generalization of a single scalar $ \beta $, as well as $ \alpha $ and other parameters, over extended area using the  $ z $-image-based snake model is more relevant.
\newtext{In summary, the proposed SRSM succeeds in providing a relevant solution, regarding all three main aspects of the snake parameterization.
	Indeed, since almost every building exhibits a strong elevation variation with respect to its surrounding area, the characteristics of building appearances on their respective $ z $-image should be all similar.
	As a result, the proposed SRSM can be generalized with the same set of influential parameters on buildings of various size and shape, as well as in complex environments.
}

\section{Conclusions}\label{sec:conclusions} 
In this paper, we proposed and evaluated an unsupervised and automatic building extraction method dedicated to a large-scale urban scene.
This method is built around an efficient snake model, named SRSM.
%It involves in using predominantly the LiDAR data, with additional cue from the optical imagery data for vegetation removal.
First, a preliminary extraction of building boundaries from the LiDAR point cloud is carried out. 
%It also involves in using additional cue from the optical imagery data to remove vegetation, which could be misclassified as buildings without this step.
%It aims to provide the snake model with reliable initial points.
These boundaries are used as initial points for the SRSM, as well as in the improved balloon force.
Second, in order to resolve the sparsity problem related to the LiDAR data spatial resolution compared to an optical imagery dataset \cite{chen2012building}, we propose a super-resolution process.
%Since a building extraction method using LiDAR data is usually compromised by the sparsity problem \cite{chen2012building}, we propose a super-resolution process.
Such process is devoted to the projection and propagation of LiDAR data onto the image space, enabling to augment its spatial resolution.
Then, the snake model is carried out based on the resulting $ z $-images. % (i.e. the LiDAR-based high-resolution image). 
Such $ z $-images encoding LiDAR elevation data %provides the LiDAR elevation data with a spatial resolution equivalent to the optical imagery data.
are highly beneficial since the height changes provide more reliable cue for extracting buildings than the spectral and textural changes provided by the optical images.
In addition to such benefit, the useful elevation data are now provided with high spatial resolution.
Third, the balloon force is improved to behave more adaptively compared to the classical balloon force.
%Indeed, the improved balloon force will shrink or inflate the snake adaptively, instead of continuously inflating it.
%\modif{Lastly, a polygonization step is carried out based on the method proposed by \cite{gribov2017searching}.}

By using the $ z $-image, a number of typical problems related to the optical image have also been addressed.
Until now, all of the existing snake models have conceded the sensitivity problem against image noises and details, such as roof objects and nearby cars and trees.
Such scene elements prompt undesired sources of attraction, causing the snake model unable to converge toward the true building  edges.
Operating on the $ z $-image which only exhibits significant height changes, the SRSM is provided with relevant sources of attraction.
In addition, such fundamental replacement---i.e. using the $ z $-image instead of the optical image---also affects the parametrization of the snake model.
Indeed, the  need for a hyperparameter tuning, e.g. by a deep learning approach \cite{marcos2018learning}, becomes less substantial.
Thus, the SRSM is parametrized with fixed scalar values.
By the virtue of the proposed improvements, such static parametrization does not restrain the applicability and scalability of the $ z $-image-based snake model over large extended area.
A comprehensive comparison and discussion of this parametrization with the deep learning approach by \cite{marcos2018learning} has also been carried out in this paper.

Concerning the performance assessment, the SRSM is tested in two different geographical contexts, namely Europe (with the Vaihingen benchmark dataset) and North America (with the Quebec City dataset). 
\newtexttt{The two contexts involve various differences in terms of compactness, density and regularity of urban areas \cite{huang2007global}.}
The proposed SRSM yields very high accuracy on the ISPRS Vaihingen benchmark dataset, namely 86.57\% of area-based \textit{Quality} and 81.60\% of object-based \textit{Quality}.
These values show that the SRSM is highly desirable, especially as a fully unsupervised method, as opposed to many other high-accuracy methods.
%\modif{\textbf{Conclusions on the performance on Quebec City data.}
%One of the drawbacks... %is the fact that we are obligated to carry out the SRSM on tiles, for the sake of computational cost.
%Future efforts will be invested into resolving such drawback.
%}
\newtextt{Concerning the Quebec City dataset with the total area of 656 km$ ^2 $, the SRSM succeeds at providing a relatively high accuracy, namely area-based \textit{Quality} of 62.37\% and object-based \textit{Quality} of 63.21\%.
Such accuracy level on this dataset may seem less desirable than the one on the Vaihingen dataset mentioned above.
However,  it can be well expected on such a large-scale dataset, with various types of complex residential, urban and industrial scenes.
Indeed, compared to the building footprints produced by Microsoft by a deep neural network approach, our unsupervised method succeeds at providing a competitive accuracy level. }
The two geographical contexts also show the very high capacity of the SRSM for extending over very large and complex areas. %, and on different geographical contexts.
\newtexttt{With the proposed SRSM, this study has achieved our objectives for a scalable, versatile and accurate building extraction solution, in the context of the flood risk assessment in the province of Quebec.}
Future works will focus on improving the resulting geometrical accuracy, as well as on several remaining problems such as shadowed vegetation, and mis-detection of small buildings.

%%%%%%%%%%%%%%%%%%%%%%%%%%%%%%%%%%%%%%%%%%
%\section{Patents}
%This section is not mandatory, but may be added if there are patents resulting from the work reported in this manuscript.

%%%%%%%%%%%%%%%%%%%%%%%%%%%%%%%%%%%%%%%%%%
\vspace{6pt} 

%%%%%%%%%%%%%%%%%%%%%%%%%%%%%%%%%%%%%%%%%%
%% optional
%\supplementary{The following are available online at \linksupplementary{s1}, Figure S1: title, Table S1: title, Video S1: title.}


%%%%%%%%%%%%%%%%%%%%%%%%%%%%%%%%%%%%%%%%%%
\authorcontributions{
%For research articles with several authors, a short paragraph specifying their individual contributions must be provided. The following statements should be used ``conceptualization, X.X. and Y.Y.; methodology, X.X.; software, X.X.; validation, X.X., Y.Y. and Z.Z.; formal analysis, X.X.; investigation, X.X.; resources, X.X.; data curation, X.X.; writing--original draft preparation, X.X.; writing--review and editing, X.X.; visualization, X.X.; supervision, X.X.; project administration, X.X.; funding acquisition, Y.Y.'', please turn to the  \href{http://img.mdpi.org/data/contributor-role-instruction.pdf}{CRediT taxonomy} for the term explanation. Authorship must be limited to those who have contributed substantially to the work reported.
Conceptualization, T.H.N.; 
Funding acquisition, S.D., D.G., C.S. and J-M.L.C.;
Investigation, T.H.N., S.D.;
Methodology, T.H.N.; 
Project administration, S.D.; 
Resources, S.D., D.G., C.S. and J-M.L.C.;
Supervision, S.D., J-M.L.C.; 
Validation, T.H.N., S.D.; 
Visualization, T.H.N.; 
Writing--original draft preparation, T.H.N.; 
Writing--review \& editing, T.H.N., S.D., D.G., C.S. and J-M.L.C.
}

%Conceptualization, Thanh Huy Nguyen; Funding acquisition, Sylvie Daniel, Didier Guériot, Christophe Sintès and Jean-Marc Le Caillec; Investigation, Thanh Huy Nguyen and Sylvie Daniel; Methodology, Thanh Huy Nguyen; Project administration, Sylvie Daniel; Resources, Sylvie Daniel, Didier Guériot, Christophe Sintès and Jean-Marc Le Caillec; Supervision, Sylvie Daniel and Jean-Marc Le Caillec; Validation, Thanh Huy Nguyen and Sylvie Daniel; Visualization, Thanh Huy Nguyen; Writing – original draft, Thanh Huy Nguyen; Writing – review & editing, Thanh Huy Nguyen, Sylvie Daniel, Didier Guériot and Jean-Marc Le Caillec.

%%%%%%%%%%%%%%%%%%%%%%%%%%%%%%%%%%%%%%%%%%
%\funding{Please add: ``This research received no external funding'' or ``This research was funded by NAME OF FUNDER grant number XXX.'' and  and ``The APC was funded by XXX''. Check carefully that the details given are accurate and use the standard spelling of funding agency names at \url{https://search.crossref.org/funding}, any errors may affect your future funding.}
\funding{This research was funded in part by the  Ministère de la Sécurité publique, Gouvernement du Qu\'{e}bec, Canada, \newtexttt{project ORACLE-2}, \newtextttt{in part by the Natural Sciences and Engineering Research Council of Canada (NSERC) grant number RGPIN-2018-04046,  and in part by the Brittany region, France.}}

%%%%%%%%%%%%%%%%%%%%%%%%%%%%%%%%%%%%%%%%%%
%\acknowledgments{In this section you can acknowledge any support given which is not covered by the author contribution or funding sections. This may include administrative and technical support, or donations in kind (e.g., materials used for experiments).}
\acknowledgments{The authors would  like to thank the Centre G\'{e}oStat (Universit\'{e} Laval), as well as the Communaut\'{e} M\'{e}tropolitaine de Qu\'{e}bec (QC, Canada) for providing the Quebec City datasets used in this work. They also would like to thank the City of Quebec for providing the \textit{Empreintes des bâtiments} dataset used as ground truth building footprints in this work.
A special thank goes to Dr. Eric Janssens-Coron from the Centre de Recherche en Données et Intelligence Géospatiales (Universit\'{e} Laval) for his help on the management of the Quebec City datasets.
They also would like to thank the Microsoft Bing Maps team for developing and releasing the open Canada building footprints.
The Vaihingen dataset was provided by the German Society for Photogrammetry, Remote Sensing and Geoinformation (DGPF) \cite{cramer2010dgpf}.
Lastly, they would like to thank Dr. Dirk-Jan Kroon from University of Twente for his contributions on the development of the conventional snake models.
}

%%%%%%%%%%%%%%%%%%%%%%%%%%%%%%%%%%%%%%%%%%
\conflictsofinterest{The authors declare no conflict of interest.} %The funders had no role in the design of the study; in the collection, analyses, or interpretation of data; in the writing of the manuscript, or in the decision to publish the results.}
%\conflictsofinterest{Declare conflicts of interest or state ``The authors declare no conflict of interest.'' Authors must identify and declare any personal circumstances or interest that may be perceived as inappropriately influencing the representation or interpretation of reported research results. Any role of the funders in the design of the study; in the collection, analyses or interpretation of data; in the writing of the manuscript, or in the decision to publish the results must be declared in this section. If there is no role, please state ``The funders had no role in the design of the study; in the collection, analyses, or interpretation of data; in the writing of the manuscript, or in the decision to publish the results''.} 

%%%%%%%%%%%%%%%%%%%%%%%%%%%%%%%%%%%%%%%%%%%
%% optional
\abbreviations{The following abbreviations are used in this manuscript:\\

\noindent 
\begin{tabular}{@{}ll}
%	ACM & Active Contour Model\\
	CNN & Convolutional Neural Network\\
	DSM & Digital Surface Model\\
	DTM & Digital Terrain Model\\
	FCN & Fully Convolutional Neural Network\\
	FISTA & Fast Iterative Shrinkage-Thresholding Algorithm\\
	GVF & Gradient Vector Flow\\
	ISTA & Iterative Shrinkage-Thresholding Algorithm\\
	LiDAR & Light Detection And Ranging\\
	NDVI & Normalized Difference Vegetation Index \\
	RMSE & Root Mean Square Error \\
	SSDG & Sum of squared directional gradients\\
	SR & Super-resolution \\
	SRSM & Super-resolution-based Snake Model \\
%MDPI & Multidisciplinary Digital Publishing Institute\\
%DOAJ & Directory of open access journals\\
%TLA & Three letter acronym\\
%LD & linear dichroism
\end{tabular}}

%%%%%%%%%%%%%%%%%%%%%%%%%%%%%%%%%%%%%%%%%%%
%%% optional
%\appendixtitles{no} %Leave argument "no" if all appendix headings stay EMPTY (then no dot is printed after "Appendix A"). If the appendix sections contain a heading then change the argument to "yes".
\appendix
%%\section{}
%%\unskip
%%\subsection{}
%%The appendix is an optional section that can contain details and data supplemental to the main text. For example, explanations of experimental details that would disrupt the flow of the main text, but nonetheless remain crucial to understanding and reproducing the research shown; figures of replicates for experiments of which representative data is shown in the main text can be added here if brief, or as Supplementary data. Mathematical proofs of results not central to the paper can be added as an appendix.
%%
%%\section{}
%%All appendix sections must be cited in the main text. In the appendixes, Figures, Tables, etc. should be labeled starting with `A', e.g., Figure A1, Figure A2, etc. 
%
\section{External Image-based Energy Term of Snake Model}
%\subsection{Image-based energy terms}
\label{app:img_ext}
The line functional is defined based on the intensity of the image $ I(x,y) $, with a filter for smoothing or noise reduction, such as Gaussian filter:

\begin{equation}
E_{line}=G_\sigma(x,y) \ast I(x,y)
\end{equation}

The  edge functional is based on the image gradient, which attracts for the snake to move towards edges with high gradient value.

\begin{equation}
E_{edge}=-\left|\nabla \left[G_\sigma(x,y) \ast I(x,y)\right]\right|^2
\end{equation}
where $ G_\sigma(x,y) $ is a two-dimensional Gaussian function with a standard deviation $ \sigma $ and $ \ast $ denotes the 2-D convolution operator.

Curvature of level lines in a slightly smoothed image can be used to detect corners and line segment terminations in an image. Using this method, let $ C(x,y) =G_\sigma \ast I(x,y)$ be the smoothed image. With an angle $ \theta=\tan^{-1}({C_y}/{C_x}) $, the unit vectors which are along and perpendicular to the gradient direction are:

%\begin{subequations}
%\begin{array}{l}
\begin{equation}
\mathbf{n}=(\cos\theta, \sin\theta),~
%\end{equation}
%\begin{equation}
\mathbf{n}_{\bot}=(- \sin\theta, \cos\theta)
\end{equation}
%\end{array}
%\end{subequations}
The termination functional of energy is  defined as:

\begin{equation}
E_{term}=\dfrac{\partial\theta}{\partial\mathbf{n}_{\bot}}=\dfrac{\partial^2 C/ \partial\mathbf{n}^2_{\bot}}{\partial C/ \partial\mathbf{n}}=\dfrac{C_{yy}C^2_x-2C_{xy}C_xC_y+C_{xx}C^2_y}{(C_x^2+C_y^2)^{3/2}}
\end{equation}

\section{Super-solution quality metrics}\label{app:psnr_ssim}
Given the super-resolved image $ I $ and the reference image $ R $, the Structural Similarity (SSIM) quality assessment index is based on the computation of three terms, namely the luminance term, the contrast term and the structural term. 

\begin{equation}\label{eq:ssim}
\mathrm{SSIM}(I,R)=\left[l(I,R)\right]^\gamma \cdot \left[c(I,R)\right]^\delta \cdot \left[s(I,R)\right]^\epsilon
\end{equation}
where 
\begin{equation}\label{eq:ssim3}%\nonumber
%\begin{array}{rl}
l(X,Y) = \dfrac{2\mu_X\mu_Y+C_1}{\mu_X^2+\mu_Y^2+C_1},~
c(X,Y) = \dfrac{2\sigma_X\sigma_Y+C_2}{\sigma_X^2+\sigma_Y^2+C_2},~
s(X,Y) = \dfrac{\sigma_{XY}+C_3}{\sigma_X\sigma_Y+C_3}
%\end{array}
\end{equation}
with $ \mu_X $, $\mu_Y$, $\sigma_X$, $\sigma_Y $, and $ \sigma_{XY} $ respectively are the local means, standard deviations, and cross-covariance for images $ X $ and $ Y $.  The parameters for SSIM index are set as follows, $\gamma=\delta=\epsilon=1 $; and $ C_1=(0.01\times L)^2, C_2=(0.03\times L)^2, C_3=C_2/2 $, where $ L=2^{\textnormal{\#bits per pixels}}-1 $ denotes the dynamic  range value of the images. With these parameters, the SSIM index \eqref{eq:ssim} is simplified into,

\begin{equation}\label{eq:ssim2}
\mathrm{SSIM}(I,R)=\dfrac{2\mu_I\mu_R+C_1}{\mu_I^2+\mu_R^2+C_1}\cdot\dfrac{2\sigma_{I,R}+C_2}{\sigma_I^2+\sigma_R^2+C_2}
\end{equation}

Another metric for evaluating a method of super-resolution of image is Peak Signal-to-Noise Ratio (PSNR) in decibels, which is defined by Equation \eqref{eq:psnr}.

\begin{equation}\label{eq:psnr}
\mathrm{PSNR}(I,R)=10\times \log_{10} \left(\dfrac{\mathrm{peak\_val}^2}{\mathrm{MSE}(I,R)}\right)
\end{equation}
where $ \mathrm{peak\_val} $ is the maximum possible value of the images, and $ \mathrm{MSE} $ is the mean square error between  $ I $ and $ R $.

%%%%%%%%%%%%%%%%%%%%%%%%%%%%%%%%%%%%%%%%%%
\reftitle{References}


%\externalbibliography{yes}
%\bibliography{your_external_BibTeX_file}

%=====================================
% References, variant B: internal bibliography
%=====================================

\bibliography{IEEEabrv,reference_abbr} 
		
%\begin{thebibliography}{999}
%% Reference 1
%\bibitem[Author1(year)]{ref-journal}
%Author1, T. The title of the cited article. {\em Journal Abbreviation} {\bf 2008}, {\em 10}, 142--149.
%% Reference 2
%\bibitem[Author2(year)]{ref-book}
%Author2, L. The title of the cited contribution. In {\em The Book Title}; Editor1, F., Editor2, A., Eds.; Publishing House: City, Country, 2007; pp. 32--58.
%\end{thebibliography}

% The following MDPI journals use author-date citation: Arts, Econometrics, Economies, Genealogy, Humanities, IJFS, JRFM, Laws, Religions, Risks, Social Sciences. For those journals, please follow the formatting guidelines on http://www.mdpi.com/authors/references
% To cite two works by the same author: \citeauthor{ref-journal-1a} (\citeyear{ref-journal-1a}, \citeyear{ref-journal-1b}). This produces: Whittaker (1967, 1975)
% To cite two works by the same author with specific pages: \citeauthor{ref-journal-3a} (\citeyear{ref-journal-3a}, p. 328; \citeyear{ref-journal-3b}, p.475). This produces: Wong (1999, p. 328; 2000, p. 475)

%%%%%%%%%%%%%%%%%%%%%%%%%%%%%%%%%%%%%%%%%%
%% optional
%\sampleavailability{Samples of the compounds ...... are available from the authors.}

%% for journal Sci
%\reviewreports{\\
%Reviewer 1 comments and authors’ response\\
%Reviewer 2 comments and authors’ response\\
%Reviewer 3 comments and authors’ response
%}

%%%%%%%%%%%%%%%%%%%%%%%%%%%%%%%%%%%%%%%%%%
\end{document}

%% file: registration_and_SR_new.tex
\subsubsection{Generation of $ z $-image by the super-resolution of LiDAR data}\label{ssec:sr}
%\subsection{Generation of $ z $-image by the super-resolution of LiDAR data}\label{ssec:sr}
The accuracy of a building extraction method using LiDAR data is usually compromised by the sparsity problem \cite{chen2012building}. 
Therefore, we propose a process dedicated to the projection and propagation of LiDAR data {onto the image space} in order to augment its spatial resolution.
Such process is called super-resolution (SR), and illustrated by the flowchart in Figure \ref{fig:flowchart_SR}.
It consists in generating a $ z $-image that contains the altitude values derived from the LiDAR 3-D point cloud.
Such image has the same size and resolution as the optical image. 
%Pixels of the super-resolved image (called the $ z $-image) contain the values derived from LiDAR 3-D points, i.e. altitude values. % or laser return intensity values. 
%The super-resolved image of LiDAR-derived altitude values is called the $ z $-image. %, whereas the image of intensity values is called $ i $-image.
%\paragraph{Mathematical notation}
The inputs of the SR process are the LiDAR point cloud, %, metadata of the optical image such as size, and lastly 
a set of camera pose parameters, the {frame of reference} and the size of the optical image. 
%%We denote the optical image by $ u \in \mathbb{R}^{n_x \times n_y \times 3} $ where $ n_x $ and $ n_y $  are respectively the number of rows and columns of the image. The georeferencing frame of the optical image $ u $ is denoted by $ r_u $.
%%The LiDAR 3-D point cloud is represented by $ \psi \in \mathbb{R}^{m \times 3} $, with $ m $ is the number of LiDAR points. Each point contains three spatial coordinates $ (x,y,z) $. Also, we use $ \psi^z \in \mathbb{R}^{m}$ to denote the column of altitude values, and $ \phi  $ to denote the resulting $ z $-image.
%%whereas $ \psi^i \in \{0, 1,..., 255\}^{m}$ stands for the intensity value of LiDAR points. For the sake of simplicity, we use the same notation $ \phi  $ to denote the result of the SRs, i.e. the $ z $-image and $ i $-image. 
%%During the SR process, $ \phi $ is vectorized into a column vector of $ n={n_x \times n_y} $ elements.
%%\paragraph{Transferring of LiDAR values}
%\modif{The principle of the SR is mathematically summarized by Equation \eqref{eq:sr}. }% describes the principle of the proposed SR. % (Eq. \eqref{eq:sr}).
%
%
%\begin{equation}\label{eq:sr}
%{\phi} = f_{\mathrm{SR}}(\psi^z, \theta, r_u, n_x, n_y)
%\end{equation}
%where $ \phi  $ denotes the resulting $ z $-image. 
The LiDAR 3-D point cloud is denoted by $ \psi \in \mathbb{R}^{m \times 3} $ where $ m $ is the number of points. 
Each point has three spatial coordinates $ (x, y, z) $.
We also use $ \psi^z \in \mathbb{R}^{m}$ for the column of altitude values. 
The $ z $-image is denoted by $ \phi \in \mathbb{R}^{n_x \times n_y}$, where $ n_x $ and $ n_y $  are, respectively, the number of rows and columns.
During the SR process, $ \phi $ is vectorized into a column vector of $ n={n_x \times n_y} $ elements.
The set of camera pose parameters $ \theta $ results from the registration. %\cite{nguyen2019coarsetofine}.
It is used to define the projection of 3-D points onto the image space.
%If the optical image and the LiDAR data are already registered or they exhibit relatively small misalignment, the projection is orthographic (i.e. along the $ z $-axis). % \cite[Sec. 6.3]{hartley2003multiple}.
%\modif{Lastly, $ r_u $ is the \modif{frame of reference} of the optical image $ u \in \mathbb{R}^{n_x \times n_y \times 3} $, where $ n_x $ and $ n_y $  are respectively the number of rows and columns of the image. }

\begin{figure}[h]
	\centering
	\begin{tikzpicture}[every text node part/.style={align=center},every node/.style={scale=1}]%,font=\sffamily]
	\node[trapezium,draw,trapezium stretches=false,trapezium left angle=110, trapezium right angle=70,minimum height=0.35cm,text width=3cm] (Input1) at (-0.5,0) {\footnotesize LiDAR point cloud $ \psi $};
	\node[trapezium,draw,trapezium stretches=false,trapezium left angle=110, trapezium right angle=70,minimum height=0.3cm,text width=2cm] (P) at (3.1,0.85) {\footnotesize Camera pose $ {\theta} $};
	\node[draw,fill=almond,text=black,minimum height=0.35cm,text width=2.25cm] (transf) at (3.1,0) {\footnotesize  3-D projection};
	\node[trapezium,draw,trapezium stretches=false,trapezium left angle=110, trapezium right angle=70,minimum height=0.35cm,text width=1.75cm] (phi om) at (6.1,0) {\footnotesize Sparse $ z $-image {\small $ \phi_{\Omega^*} $}};
	\node[draw,fill=almond,text=black,minimum height=0.35cm,text width=2.5cm] (propa) at (9.2,0) {\footnotesize Value propagation};
	\node[trapezium,draw,trapezium stretches=false,trapezium left angle=110, trapezium right angle=70,minimum height=0.35cm,text width=1.5cm] (phi) at (12,0) {\footnotesize $ z $-image {\small $ \phi $}};
	\draw[->,draw=black] (Input1) -- (transf);
	\draw[->,draw=black] (transf) -- (phi om);
	\draw[->,draw=black] (phi om) to (propa);
	\draw[->,draw=black]  (propa) -- (phi);
	\draw[->,draw=black] (P)  -- (transf);
	%	\draw[->,draw=black] (P) --node[near start,left,text width=0.5cm]{\small $ H_{\Omega^*} $}  (2.5,0);
	\end{tikzpicture}
	\caption{Overview of the super-resolution process, generating a high-resolution LiDAR-based $ z $-image.}% ($ z $- or $ i $-image).}
	\label{fig:flowchart_SR}
\end{figure}

\paragraph{\textit{(a) Projection of LiDAR 3-D points}}
%First, LiDAR 3-D points are projected onto the frame of the optical image 
The first step of the SR process consists in projecting the LiDAR 3-D points onto the $ z $-image space using the camera pose parameters $ \theta $.
As the LiDAR point cloud is subsampled compared to the optical image, such a projection leads to a sparsity effect on the $ z $-image $ \phi $.
Here, we use $ \Omega^* $ and $ \Omega $ to denote, respectively, the subset of the pixel indices in the $ z $-image $ \phi $, having or not a projected altitude value. 
In other words, $ \phi_{\Omega^*} $ denotes the sparse $ z $-image or the sub-vector containing the pixels of projected altitude value, whereas $ \phi_\Omega $ denotes the sub-vector containing the null pixels. 
The dimension of $ \phi_{\Omega^*} $ and $ \phi_\Omega $, respectively, are $ m\times 1 $ and $ (n-m)\times 1 $.
As such, $ \phi=\phi_{\Omega \cup \Omega^*}  $ is the vector containing all pixels, i.e. the whole $ z $-image. 
%At the first iteration of the fine registration, $ \theta $ is given by $ \theta_{global} $ obtained from the coarse registration. 
The projection is mathematically presented as follows,

\begin{equation}\label{eq:val_transfer}
\phi_{\Omega^*}=\mathcal{P}_\theta  (\psi^z) %\textnormal{ or }  \phi_{\Omega^*}=H_{\Omega^*}\psi^i
\end{equation}
%where  $ \Omega^* $ and $ \Omega $ denote, respectively, the subsets containing the indices of pixels from $ \phi $, having or not an associated altitude value projected from $ \psi $. 
%Thus, $ \phi_{\Omega^*} $ and $ \phi_\Omega $, respectively, denote the sub-vector containing the pixels with and without a projected altitude value; whereas $ \phi $ denotes the vector containing all pixels. 
where $ \mathcal{P}_\theta $ is the 3-D projection associated with the camera pose parameters $ \theta $.
\newtextt{The $ x $- and $ y $-coordinates of the LiDAR 3-D points are used to \newtexttt{locate the pixels in the $ z $-image associated with such points}.}
%The matrix $ H_{\Omega^*} $ associated to the camera pose parameters $ \theta $, is an index matrix allowing to select only the pixels with projected values.
%It is computed based on the projection related to $ \theta $ of the LiDAR 3-D point cloud onto the frame of reference of the 2-D optical image.
Next, the projected values indexed by $ \Omega^* $ will be propagated to their neighboring pixels (which are indexed by $ \Omega $).

\paragraph{\textit{(b) Propagation of the projected values}}
%The propagation of projected values is carried out through the minimization of a cost function $ \mathcal{F}(\phi) $, defined by Equation \eqref{eq:val_propa}. It is composed of the sum of squared directional gradients (SSDGs), and an $ l_1 $-norm term to promote the sparsity of $ \phi $, subjecting to the projected values (from the point cloud) described by Equation \eqref{eq:val_transfer}. 
%
%\begin{equation}\label{eq:val_propa}
%\begin{array}{c}
%\widehat{\phi}=  \underset{\phi}{\arg\min}~\left\lbrace\underbrace{\overbrace{\left\| \nabla_x\phi\right\|^2_{2} + \left\| \nabla_y\phi\right\|^2_{2}}^{\vspace{0.5cm}f_{\mathrm{SSDG}}(\phi)}+\lambda\left\|\phi\right\|_{1}}_{\vspace{0.5cm} \mathcal{F}(\phi)}\right\rbrace, %\\[20pt]
%\textnormal{subject to }  \phi_{\Omega^*}=\mathcal{P}_\theta(\psi^z) %\textnormal{ or }  \phi_{\Omega^*}=H_{\Omega^*}\psi^i 
%\end{array}
%\end{equation}
%where $ \left\|\cdot\right\|_{p} $ stands for the $ l_p $-norm, $ \nabla_x $ and $ \nabla_y $, respectively, represent the directional gradient operators along the $ x $-axis and $ y $-axis. 

Our SR approach is inspired by the work of Castonera \textit{et al.} \cite{castorena2018motion} on the fusion of terrestrial LiDAR data with optical imagery.
It involves reconstructing a sparse depth map by minimizing the sum of its squared directional gradients (SSDGs).
This approach relies on hypothetical characteristics of a depth map, which involve the magnitude and occurrence of depth discontinuities inside the depth map to be minimized. 
%The advantages of this cost function is its convexity and ease to compute. 
%This method also showed good results in propagating depth values across homogeneous regions. 
In an airborne nadir view context, their method shows good performance in propagating elevation values across homogeneous regions. 
However, in elevation-discontinued transitioning regions, e.g. near the edges of a building, the propagated elevation values would be gradually flattened as a result of the minimized SSDGs.
In other words, such hypothetical characteristics are not suitable in this context, where the off-terrain objects like trees and buildings always exhibit strong elevation discontinuities. 
Such discontinuities should be preserved during the  value propagation process.
Thus, an $ l_1 $-norm term is added in our minimization approach. % in order to preserve such elevation discontinuities. 
This preservation allows the resulting $ z $-image to exhibit elevation changes as tight as possible compared to the scene reality.
%Since the snake model operates on such image, it provides the snake model with the building boundaries as precise as possible.
%Therefore, the $ l_1 $-norm term is additionally proposed in our approach. %, in order to promote sparsity of the $ z $-image, i.e. preserving elevation discontinuities. 

The propagation of the projected values is carried out through the minimization of a cost function $ \mathcal{F}(\phi) $, defined by Equation \eqref{eq:val_propa}. It is composed of the SSDGs and an $ l_1 $-norm term of the $ z $-image $ \phi $, subjecting to the values previously projected from the point cloud (i.e. described by Equation \eqref{eq:val_transfer}). 

\begin{equation}\label{eq:val_propa}
\begin{array}{c}
\widehat{\phi}=  \underset{\phi}{\arg\min}~\left\lbrace\underbrace{\overbrace{\left\| \nabla_x\phi\right\|^2_{2} + \left\| \nabla_y\phi\right\|^2_{2}}^{\vspace{0.5cm}f_{\mathrm{SSDG}}(\phi)}+\lambda\left\|\phi\right\|_{1}}_{\vspace{0.5cm} \mathcal{F}(\phi)}\right\rbrace, %\\[20pt]
\textnormal{subject to }  \phi_{\Omega^*}=\mathcal{P}_\theta (\psi^z) %\textnormal{ or }  \phi_{\Omega^*}=H_{\Omega^*}\psi^i 
\end{array}
\end{equation}
where $ \left\|\cdot\right\|_{p} $ stands for the $ l_p $-norm, $ \nabla_x $ and $ \nabla_y $, respectively, represent the directional gradient operators along the $ x $-axis and $ y $-axis. The parameter $ \lambda >0 $ controls the amount of the $ l_1 $-regularization.

\paragraph{\textit{(c) Propagation implementation}}
The minimization of the cost function described in Equation \eqref{eq:val_propa} is carried out using the Fast Iterative Shrinkage-Thresholding algorithm (FISTA) \cite{beck2009fast}. 
Its computational efficiency is adequate for solving large-scale problems, with a convergence rate of \newtextt{$ O(1/k^2) $, where $ k $ is the iteration counter. }
FISTA is significantly faster than standard gradient-based methods such as Iterative Shrinkage-Thresholding algorithms (ISTA). 
Full details on the implementation the proposed SR process can be found in \cite{nguyen2019coarsetofine}.

\begin{figure}[h]
	\centering
	\begin{subfigure}{0.46\linewidth}
		\centering
		\includegraphics[trim=1cm 0cm 1cm 0cm,clip,width=\linewidth]{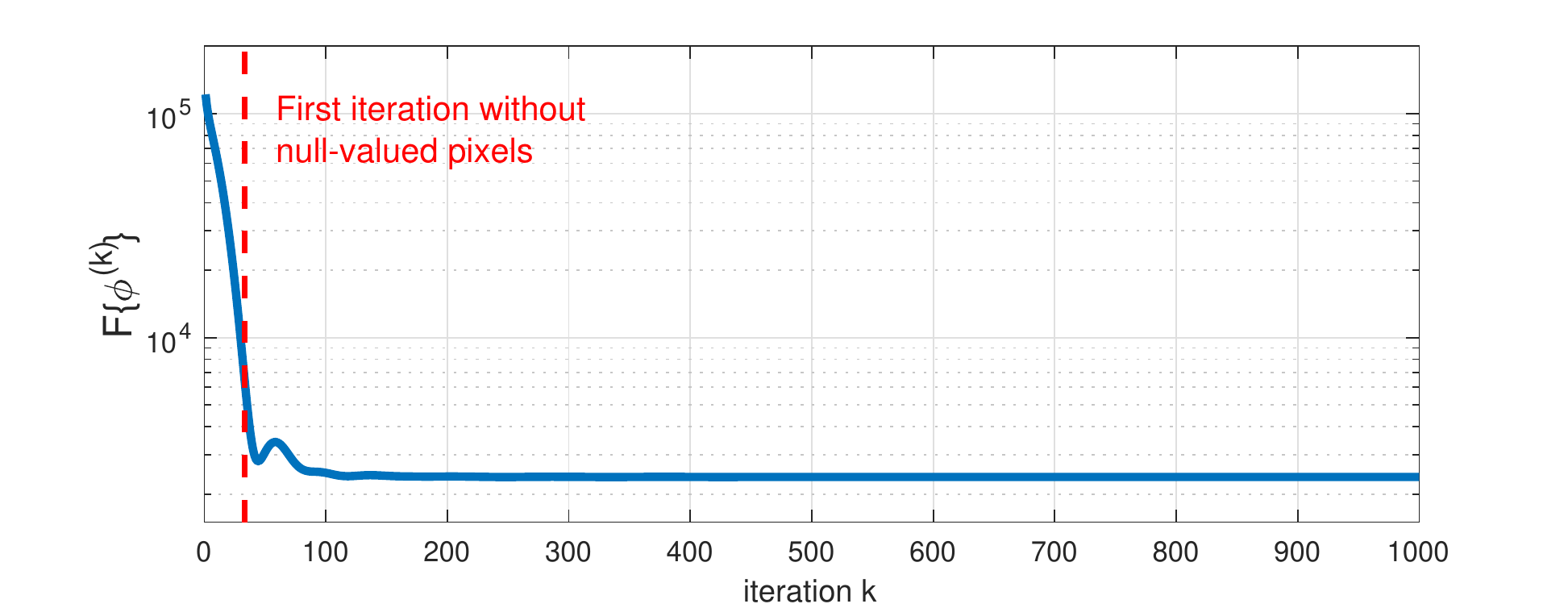}
		\caption{Difference $\left\|\phi^{(k+1)}-\phi^{(k)}\right\|_{2} $}\label{subfig:error}
	\end{subfigure}
	\hspace{0.1cm}
	\begin{subfigure}{0.46\linewidth}
		\centering
		\includegraphics[trim=1cm 0cm 1cm 0.25cm,clip,width=\linewidth]{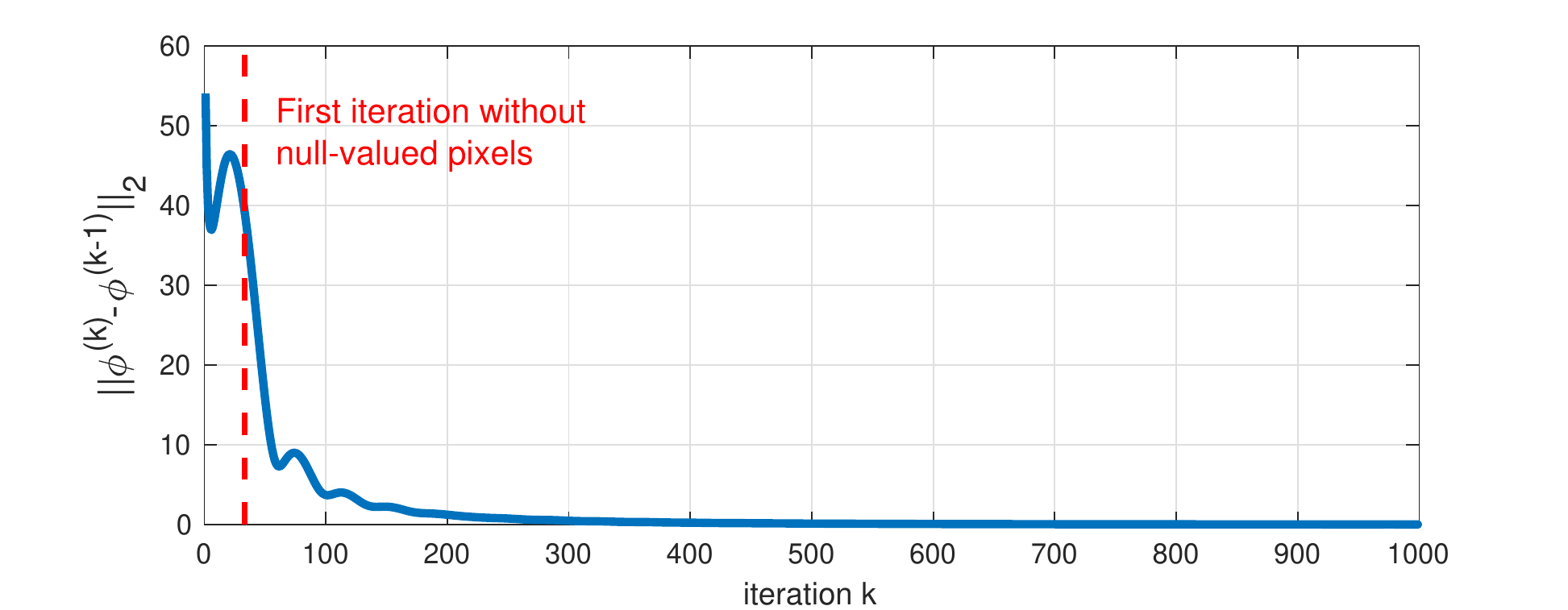}
		\caption{Cost function value $ \mathcal{F}(\phi^{(k)}) $ (in logarithmic scale)}\label{subfig:function value}
	\end{subfigure}
	\vspace{0.15cm}
	\caption{(\textbf{a}) Difference $\left\|\phi^{(k+1)}-\phi^{(k)}\right\|_{2} $ and (\textbf{b}) cost function value $ \mathcal{F}(\phi^{(k)}) $ displayed as a function of iterations, from the SR process of generating $ z $-image $ \phi $. The vertical red-dashed lines represent the first iteration where every pixels of the estimate $ z $-image is filled.}
	\label{fig:errors}
\end{figure}

\begin{figure}[h]
	\centering
	\begin{subfigure}{0.29\linewidth}
		\centering
		\includegraphics[trim=3.5cm 1.5cm  2cm 1cm,clip,width=\linewidth]{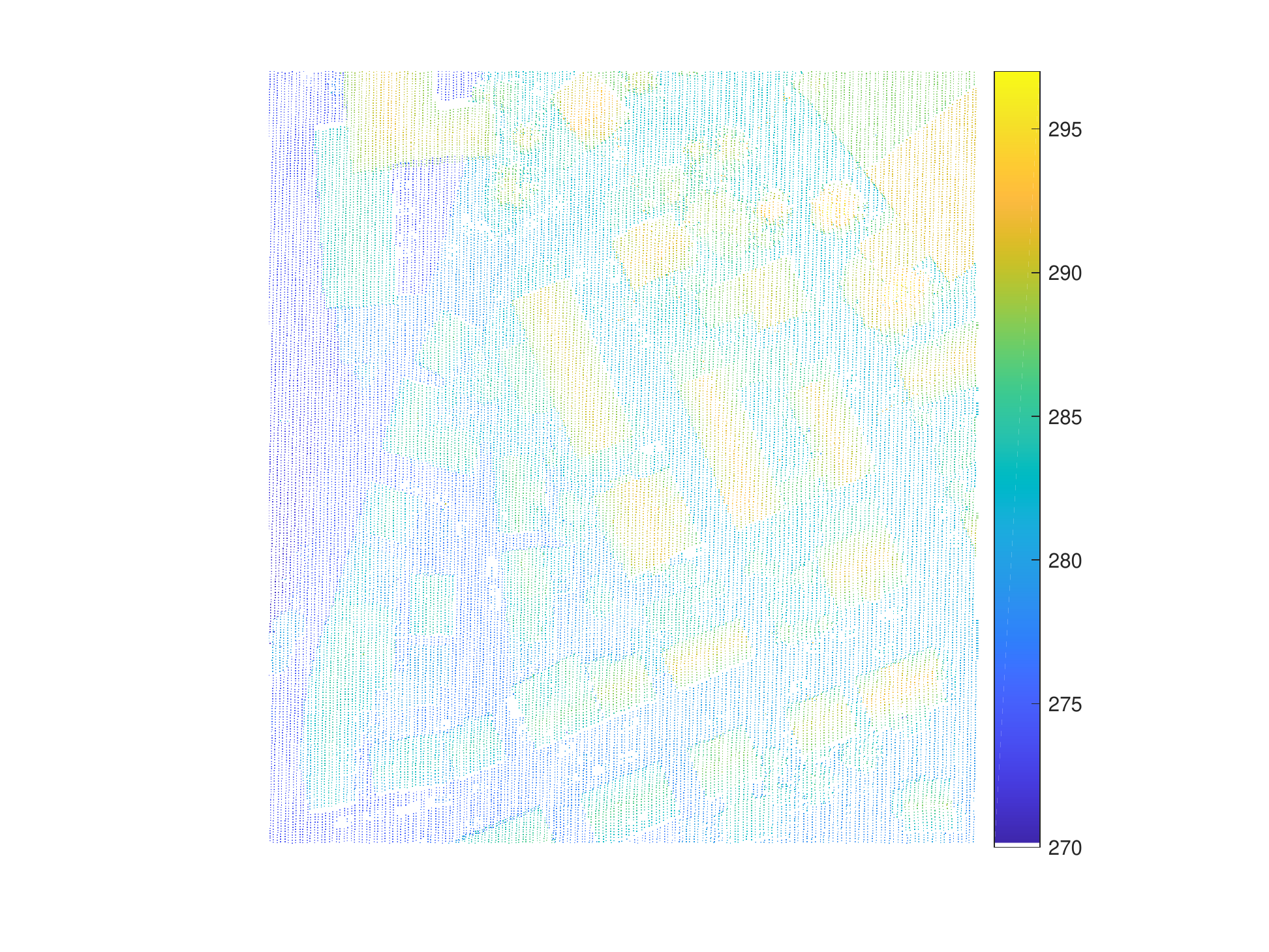}
		\caption{Sparse $ z $-image $ \phi_{\Omega^*} $}\label{sfig:sparse-z-image}
	\end{subfigure}
	\hspace{0.1cm}
	\begin{subfigure}{0.29\linewidth}
		\centering
		\includegraphics[trim=3.5cm 1.5cm  2cm 1cm,clip,width=\linewidth]{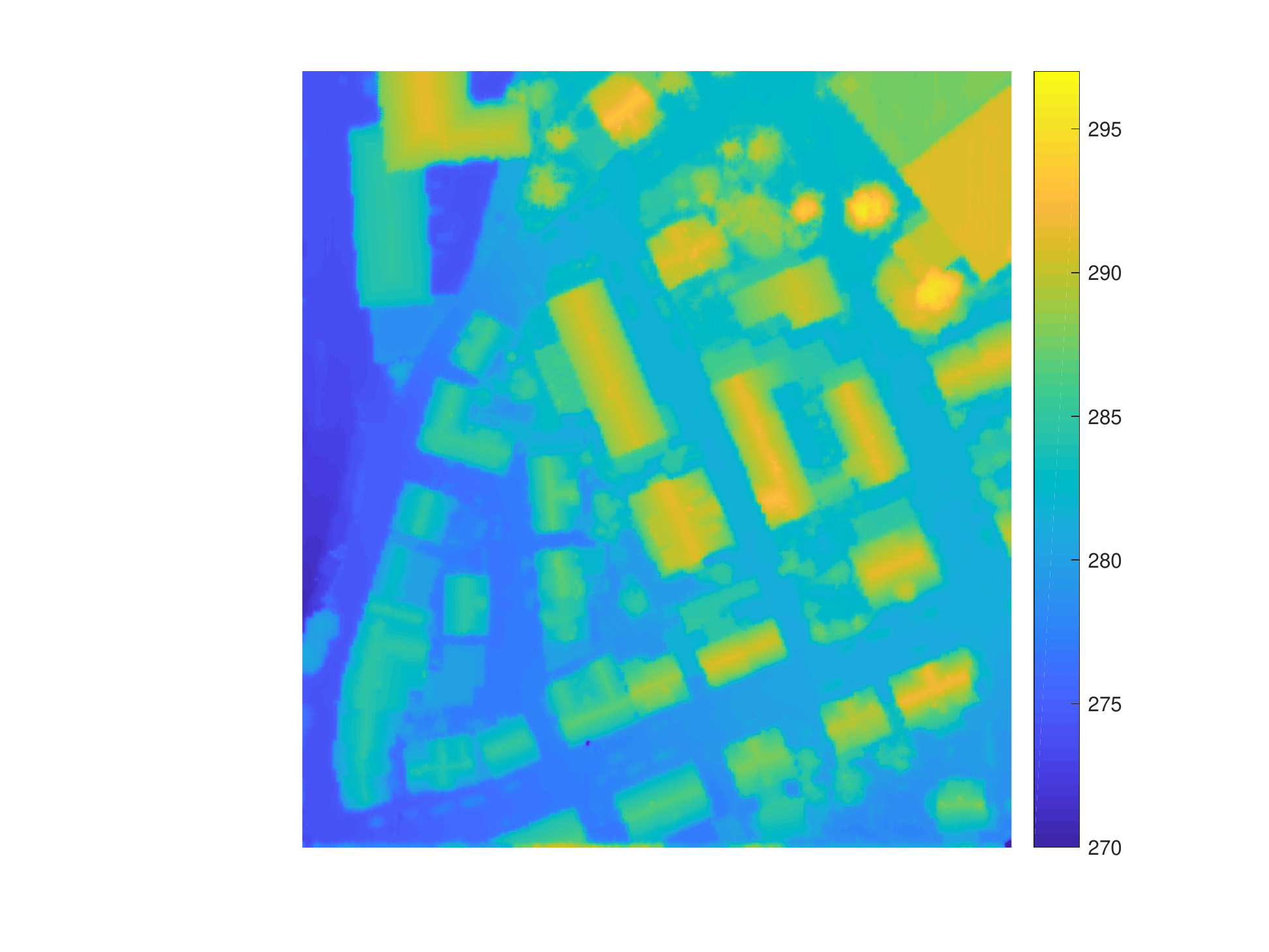}
		\caption{Dense $ z $-image $ \phi $}\label{sfig:z-image}
	\end{subfigure}
	\begin{subfigure}{0.33\linewidth}
		\centering
		\includegraphics[trim=0.75cm 0.5cm 1.25cm 0.5cm,clip,width=\linewidth]{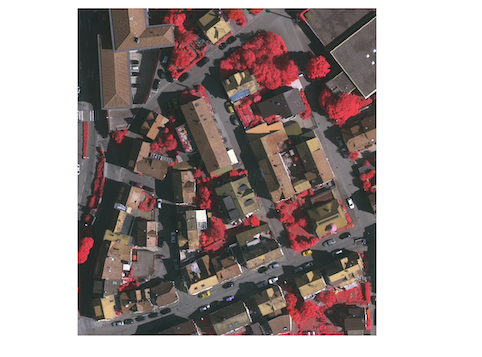}
		\caption{Reference optical image}\label{sfig:opt-image}
	\end{subfigure}
	\caption{Examples of super-resolution outcome. (\textbf{a}) The sparse $ z $-image $ \phi_{\Omega^*} $ from the projection; (\textbf{b}) The dense $ z $-image $ \phi $ from the whole SR process; (\textbf{c}) The reference optical image of the same scene for visual comparison.}
	\label{fig:sparse_dense}
\end{figure}

The convergence rate of the SR is illustrated in Figure \ref{fig:errors}. 
Figure \ref{subfig:error} depicts the differences between the estimated $ z $-images at consecutive iterations, i.e. $ \left\|\phi^{(k+1)}-\phi^{(k)}\right\|_2 $. 
The cost values $ \mathcal{F}(\phi^{(k)}) $ through iterations are shown in Figure \ref{subfig:function value}. 
One can observe that the $ z $-image has nearly converged into a stable solution after approximately four hundred iterations.
Figure \ref{fig:sparse_dense} shows the outcomes of the projection and the propagation of altitude values from the LiDAR data onto the optical image space. 
The value projection outcome is depicted by the sparse $ z $-image $ \phi_{\Omega^*} $ (Figure \ref{sfig:sparse-z-image}), whereas the value propagation outcome is shown by the dense $ z $-image $ \phi $ (Figure \ref{sfig:z-image}). 
The pixel color of the sparse and dense $ z $-images represents the surface elevation in meters.  
Figure \ref{sfig:opt-image} shows the reference optical image on the same urban scene, in order to assess visually the quality of the super-resolved $ z $-image. 
It can be noted that the elevation of buildings and other objects (e.g. trees, cars, etc.) %, \modif{as well as the relief of the scene} 
are well presented on the dense $ z $-image (Figure \ref{sfig:z-image}), and correspond to the information in the optical image. 
The proposed SR process is shown to achieve the purpose of augmenting the spatial resolution of the LiDAR point cloud.
An assessment of the SR performance will be presented in sub-section \ref{ssec:sr_eval}.

%% file: chap4_result.tex
\section{Experimental Results}\label{sec:results}
In this Section, multiple performance evaluations are carried out.
First, we introduce the building extraction accuracy metrics, as well as the study areas and the datasets used in this work.
Then, the performance of the SR process is evaluated.
Next, a visual assessment between the snake models is also carried out.
Lastly, the proposed SRSM is evaluated on various urban and residential scenes.

\subsection{Building extraction accuracy metrics}\label{ssec:metric}
Multiple accuracy assessments, thematically and geometrically, are proposed to evaluate the performance of a building extraction method based on the ground truth boundaries.
\subsubsection{Thematic accuracy metrics}
Based on the evaluation methodology described by Rutzinger \textit{et al.} \cite{rutzinger2009comparison}, three metrics, namely \textit{Quality} ($ Q $), %\textit{$ F1 $-score} ($ F1 $), 
\textit{Completeness} ($ Cp $) and \textit{Correctness} ($ Cr $), are measured \newtexttt{per-object and per-area. %according to \newtexttt{two levels, namely per-object and per-area. 
Particularly, the per-object evaluation involves either all objects regardless of their area, or only the objects with an area larger than 50 m$ ^2 $.}
The three metrics are computed based on the count of true positive (TP), false positive (FP), false negative (FN) elements between the extracted and the reference building boundaries from the ground truth. 
These elements (TP, FP, FN) are defined differently \newtexttt{if the evaluation is carried out per-object or per-area}.
%according to the object-based or area-based assessments.
%\begin{itemize}[leftmargin=*,labelsep=5.8mm]
%	\item Area-based: relying on counting TP/FP/FN pixels on the image.
%	\item Object-based: by counting the number of buildings.
%	A building is counted as a TP if at least 50\% of its area coincide with its ground truth.
%	\item Object-based (50 m$ ^2 $): show the object-based performance for the most relevant objects (i.e. buildings that cover an area larger than 50 square meters).
%\end{itemize}

For the per-object evaluation, an extracted building is counted as a TP if at least 50\% of its area coincide with its ground truth. 
On the other hand, a FP is an extracted building \newtexttt{without a corresponding building in the ground truth, or if the coincided area with the ground truth is less than 50\%. 
Whereas a FN means the proposed approach fails to extract a building existing in the ground truth.}
The corresponding $ Cp, Cr, Q $ metrics are then computed using Equation \eqref{eq:iou_comp_corr}.
%The \textit{Completeness} and \textit{Correctness}  are respectively measured by the \textit{Recall} and \textit{Precision} criteria.

\begin{equation}\label{eq:iou_comp_corr}
Cp = \dfrac{TP}{TP+FN},~
Cr = \dfrac{TP}{TP+FP},~
Q = \dfrac{TP}{TP+FP+FN}%,~
%F1 = \dfrac{2TP}{2TP+FP+FN}
\end{equation}
%	\modif{More...}
%	Full descriptions of these metrics can be found in  \cite{rutzinger2009comparison}.

\newtextt{For the per-area evaluation, such metrics are computed using the count of pixels on the image. 
%TP, FP, and FN denote, respectively, the number of building pixels correctly identified, the number of non-building pixels identified as building pixels, and the number of building pixels not identified. 
The area-based \textit{Quality} $ Q $ is equal to the Intersection over Union (IoU) metric, which measures the ratio between the intersection area over the union area of the extracted building boundary $ E $ and the corresponding ground-truth $ R $ (Equation \eqref{cri:area}). %\textit{$F1$-score} is also a metric widely used in assessing the building extraction.
It reflects the overall accuracy of the building extraction method according to the ground truth.
The \textit{Completeness} $ Cp $ measures the fraction of relevant identified building pixels over the total number of actual building pixels,
whereas the \textit{Correctness} $ Cr $ computes the fraction of relevant identified building pixels among all identified pixels.
%	Two related criteria, namely \textit{Completeness} ($ Cp $) and \textit{Correctness} ($ Cr $), are also used by other works, e.g. \cite{fazan2013rectilinear,gilani2016automatic}. 
}

\newtextt{
%\begin{subequations}
\begin{equation}\label{cri:area}
Cp %= \dfrac{TP}{TP+FN}
= \dfrac{\mathcal{\#}(E\cap R)}{\mathcal{\#}(R)},~%\times 100\%
%	\end{equation}
%	\begin{equation}\label{cri:corr}
Cr %= \dfrac{TP}{TP+FP} 
= \dfrac{\mathcal{\#}(E\cap R)}{\mathcal{\#}(E)},~%\times 100\%
%	\end{equation}
%	\begin{equation}\label{cri:iou}
Q %= \dfrac{TP}{TP+FP+FN} 
=\dfrac{\mathcal{\#}(E\cap R)}{\mathcal{\#}(E\cup R)}%\times 100\%
\end{equation}
%	\begin{equation}\label{cri:f1-score}
%	F1 = 2\times\dfrac{Cp\times Cr}{Cp+Cr} = \dfrac{2TP}{2TP+FP+FN} =\dfrac{2\times\mathcal{\#}(E\cap R)}{\mathcal{\#}(E) +\mathcal{\#}(R)}%\times 100\%
%	\end{equation}
%\end{subequations}
}

\noindent where $ \mathcal{\#}(\cdot) $ denotes the number of pixels inside the given region.
All three metrics $ Cp, Cr $ and $ Q $ reach their best value at 100\% and worst at 0\%.

\subsubsection{Geometrical accuracy metrics}
The geometrical accuracy of the method can also be evaluated by measuring the root-mean-square error (RMSE) of distances from extracted building outlines to the reference outlines, without considering points with distance greater than 3 meters. \newtexttt{Such threshold is defined by the assessment methodology \cite{rottensteiner2014results}}. %More details of the assessment based on the RMSE metric are described by \cite{rutzinger2009comparison}.
A smaller distance indicates a better geometrical accuracy.
% Rottensteiner\footnote{\url{http://www2.isprs.org/tl_files/isprs/wg34/docs/EvaluationObjectDetection.pdf}}.

%In addition, in the case of Quebec cities datasets, we also measure the dominant angle rotation error (DARE) between the extracted building outlines and the ground truth,
%\begin{equation}\label{eq:dare}
%	\mathrm{DARE}=|\theta_{E}-\theta_{R}|
%\end{equation}
%where $ \theta_{E} $ and $ \theta_{R} $ represent the dominant angle of the extracted building and the ground truth building. 
%\modif{The dominant angle of a building polygon is determined according to its longest line segment. }

%Another point that can be relevant for the next research article is another additional metric to evaluate the performance of building extraction. We propose using the metric called PoLIS \cite{avbelj2014metric}, proposed by Avbeji \textit{el al.} This metric is designed specifically for the building extraction problem, thus very interesting to be taken into consideration.

\subsection{Study areas and involved datasets}
\unskip
\subsubsection{Vaihingen dataset}
The proposed building extraction method is tested using the ISPRS benchmark dataset on Vaihingen, Germany \cite{cramer2010dgpf}.
The test aims to demonstrate its effectiveness on complex environments, and to compare it with other methods. 
The ISPRS Vaihingen benchmark dataset involves three test areas consisting of buildings with diversified characteristics. %, as well as their ground truth boundaries. 
In these test areas, the ground truth boundaries consisting of roof outline polygons were generated based on manual stereo plotting, with an associated planimetric accuracy of approximately 10 cm \cite{rottensteiner2014results}.
The columns two and three of the Table \ref{tab:datasets_ISPRS} describe the involved LiDAR and optical imagery datasets on these areas.
\newtextt{Concerning the LiDAR data, we only use the data from one strip for each area. % (i.e. only LiDAR strip 9 for the Area 1, strip 5 for the Area 2, and strip 3 for the Area 3).
%The employed airborne LiDAR dataset is acquired on  August 21, 2008. Its average point density (PD) is of 4 points/m$ ^2 $.
%For each Area, we only use the LiDAR data from one strip, i.e. only LiDAR strip 9 for the Area 1, and only strip 5 for the Area 2, and lastly  only strip 3 for the Area 3.
%%Since the misalignment between the  orthophoto and the airborne LiDAR data was already small, a registration between the datasets is not carried out.
%The orthoimage given as a NIR-R-G\footnote{Near-infrared - Red - Green} image of 9-cm ground sampling distance (GSD) is generated from aerial images that are taken between July 24 and August 06, 2008. 
The orthoimage was generated based on the  DSM derived from the LiDAR data.
As a result, the misalignment between them is relatively small (i.e. less than 30 cm).}

\begin{table}[H]
	\caption{Description of the ISPRS Vaihingen benchmark dataset, and the Quebec City dataset.}
	\label{tab:datasets_ISPRS}
	\centering
%	\tablesize{\footnotesize} %% You can specify the fontsize here, e.g., \tablesize{\footnotesize}. If commented out \small will be used.
	\begin{tabular}{cccccc}
		\toprule
		& \multicolumn{2}{c}{\textbf{Vaihingen}} && \multicolumn{2}{c}{\textbf{Quebec City}}  \\
		\cline{2-3}\cline{5-6}\\[-7pt]
		\textbf{Specifications} & \textbf{Optical image} & \textbf{LiDAR} && \textbf{Optical image} & \textbf{LiDAR} \\
		\midrule
		Spectral resolution & NIR, R, G & 1064 nm && R, G, B & 1064 nm \\\midrule
		Spatial resolution & {9 cm} & \newtexttt{50 cm} && {15 cm} & \newtexttt{35.4 cm} \\
		(point density) & - & (4 pts/m$^2$) && - & (8 pts/m$^2$)  \\\midrule
%		Nominal relative spatial  & \multicolumn{2}{c}{$ \times $ 5.56} && \multicolumn{2}{c}{$ \times $ 2.36} \\
%		resolution ratio & & && &\\\midrule
%		(point spacing) & - & (50 cm) && - & (35.4 cm) \\\midrule
		Acquisition time & July-August 2008  & August 21, 2008 && June 2016  & May 2017 \\\midrule
		%		(season) & (summer) & (winter) \\\midrule
		\multirow{2}{*}{Geometry/Properties} & Orthorectified & Mostly single-return && Orthorectified & Multi-return (4) \\
		& Georeferenced & Unclassified && Georeferenced & Classified \\\midrule
		\multirow{2}{*}{Relative misalignment} & \multicolumn{2}{c}{\multirow{2}{5cm}{\centering Less than 30 cm}} && \multicolumn{2}{c}{1.05 m (before registration),}  \\
		& & && \multicolumn{2}{c}{0.35 m (after registration \cite{nguyen2019coarsetofine})} \\
		\bottomrule
	\end{tabular}
\end{table}

\subsubsection{Quebec City dataset}
\begin{figure}[h]
	\tiny
	\centering\includegraphics[trim=0 22cm 10.5cm 0,clip,width=10cm]{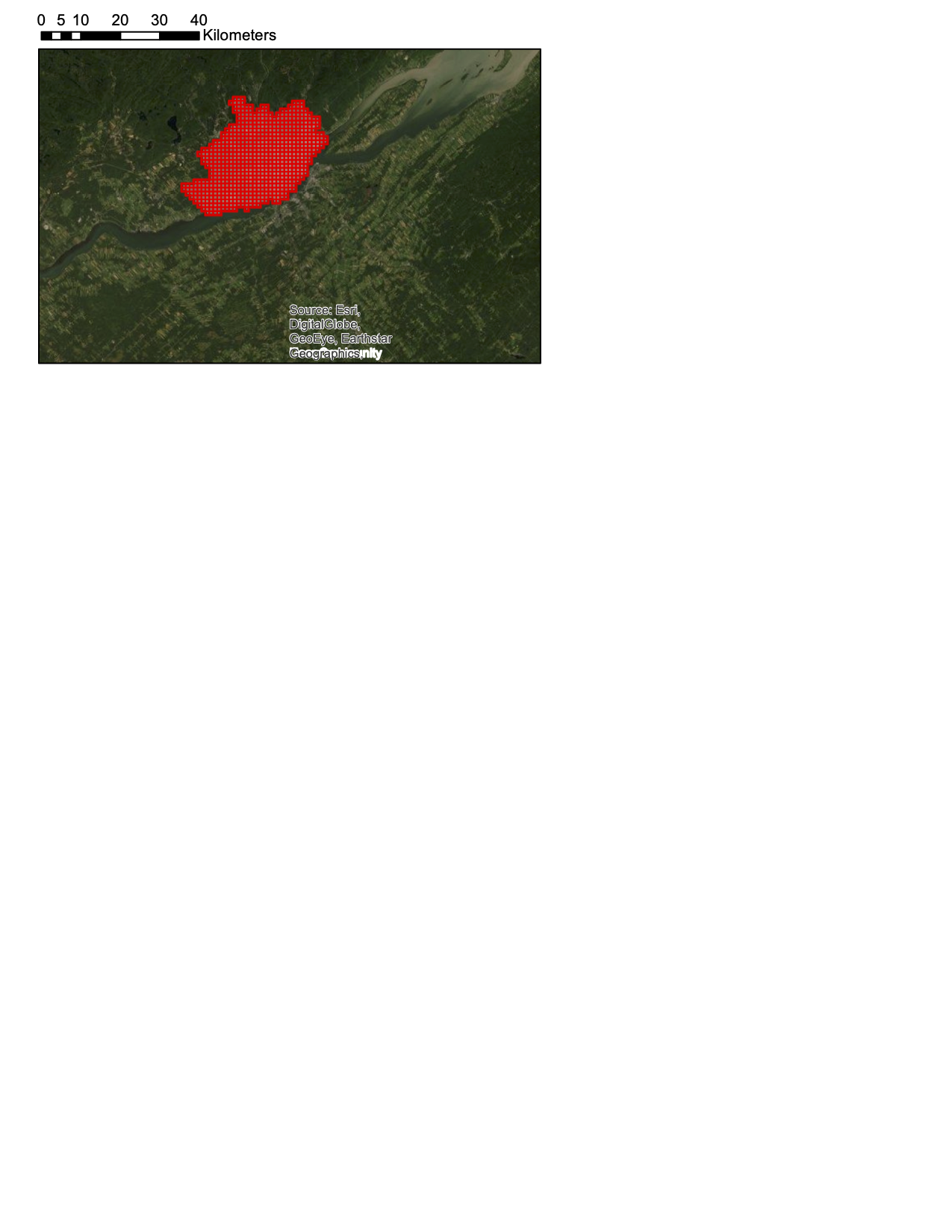}

%	\hspace{0.01cm}
%	\begin{subfigure}[b]{0.4\linewidth}
%		\centering\includegraphics[trim=0 0.25cm 0 0,clip,height=3.5cm]{fig/.png}\caption{Improved balloon force}\label{sfig:}
%	\end{subfigure}
	\caption{Quebec City dataset coverage visualized with ERSI ArcGIS Online World Imagery basemap (Source: Esri, DigitalGlobe, GeoEye, Earthstar Geographics, CNES Airbus DS, USDA, USGS, AeroGRID, IGN, and the GIS User Community).}
	\label{fig:qc_lidar_coverage}
\end{figure}

%The main goal of this study is to provide an accurate large-scale building extraction method, for the Quebec province, Canada. 
%In order to validate the generality of the proposed method, 
Beside the assessments on the Vaihingen dataset representing a European urban context, we additionally conduct a performance assessment in another geographic context, namely North America.
%In this regard, the method is performed on multiple cities in the \modif{Quebec province, namely Quebec City, Montreal and Repentigny.
%	The LiDAR and optical imagery datasets acquired on these cities are described in Table \ref{tab:datasets}.}
In this regard, the method is carried out on the urban areas of Quebec City, QC, Canada. 
\newtextt{They cover a total area of 656 square kilometers.
The whole area is divided into tiles of 1 km $ \times $ 1 km, as shown in Figure \ref{fig:qc_lidar_coverage}, for the sake of processing time and memory constraint.
The involved LiDAR and optical imagery datasets are described in the column four and five of Table \ref{tab:datasets_ISPRS}.
The ground truth boundaries of buildings in the whole test area are provided} \newtexttt{and updated monthly by the %\modif{Open  Database of Buildings (ODB) from Statistics Canada\footnote{\url{https://www.statcan.gc.ca/eng/lode/databases/odb}}}.
City of Quebec, named \textit{Empreintes des bâtiments}\footnote{\url{https://www.donneesquebec.ca/recherche/fr/dataset/empreintes-des-batiments}}.
%In order to reduce the impact of temporal difference between this ground truth dataset and the used datasets, we use 
The ground truth dataset used in this work was downloaded on March 4$ ^{th} $, 2019.}

%\begin{table}[H]
%	\caption{\modif{Description of Quebec City, Montreal and Repentigny datasets.}}
%	\label{tab:datasets}
%	\centering
%%	\tablesize{\footnotesize} %% You can specify the fontsize here, e.g., \tablesize{\footnotesize}. If commented out \small will be used.
%	\begin{tabular}{ccccccc}
%		\toprule
%		& \multicolumn{2}{c}{\textbf{Quebec City}} & \multicolumn{2}{c}{\textbf{Montreal}} & \multicolumn{2}{c}{\textbf{Repentigny}} \\\midrule
%		\textbf{Data type} & \textbf{Optical image} & \textbf{LiDAR} & \textbf{Optical image} & \textbf{LiDAR} & \textbf{Optical image} & \textbf{LiDAR}\\
%		\midrule
%		Spectral resolution & R, G, B & 1064 nm & R, G, B & 1064 nm & & \\\midrule
%		Spatial resolution & \multirow{2}{*}{15 cm} & \multirow{2}{*}{8 pts/m$^2$}  & \multirow{2}{*}{8 cm} & \multirow{2}{*}{8 pts/m$^2$} & &\\
%		(GSD or PD) & &  & & & &\\\midrule
%		Acquisition time & June 2016  & May 2017 & April 2018 & Nov. 2015 & & \\\midrule
%		%		(season) & (summer) & (winter) \\\midrule
%		Geometry/ & Orthorectified & Multi-return (4) & Orthorectified & Multi-return (7) & & \\
%		Properties & Georeferenced & Classified & Georeferenced & Classified & & \\\midrule
%		Relative & \multicolumn{2}{c}{1.05 m (before registration)} & \multicolumn{2}{c}{-} & \multicolumn{2}{c}{-} \\
%		misalignment & \multicolumn{2}{c}{0.35 m (after registration)}  & & & & \\
%		\bottomrule
%	\end{tabular}
%\end{table}

In March 2019, Microsoft has released an open Canada building footprint dataset (consisting of twelve million buildings) collaborating with Statistics Canada\footnote{\url{https://github.com/microsoft/CanadianBuildingFootprints}}. 
The dataset covers all the Quebec City territory with more than two hundred thousand building footprints. 
\newtextt{This work is carried out using a deep neural network, of which the foundation is the ResNet-34 \cite{he2016deep}.
The training set consists of three million labeled Bing images.	
%This method has been reported to yield an area-based \textit{Quality} of 76\% on a test dataset of forty-five thousand buildings.
In this paper, we conduct an individual performance assessment using the mentioned ground truth building boundaries in Quebec City, in order to compare with the SRSM results.}
%\modif{(We anticipate the results from Microsoft Canada Building Footprint are erroneous in terms of building orientations...)}

\newtextttt{It should be noted that this assessment does not only allow evaluating the performance of the proposed SRSM on such a large dataset, but it also serves as an example demonstrating the scale of the study---i.e. the Quebec province, in which other cities and large areas should not cause any adaptability problem on such an unsupervised method.}

\input{SR_eval.tex}
\subsection{Comparison between snake models}\label{ssec:compare_snakes}
\begin{figure}[h]
	\centering
	\begin{subfigure}[b]{0.3\linewidth}
		\centering
		\includegraphics[trim=1cm 0 1cm 0,clip,height=5.75cm]{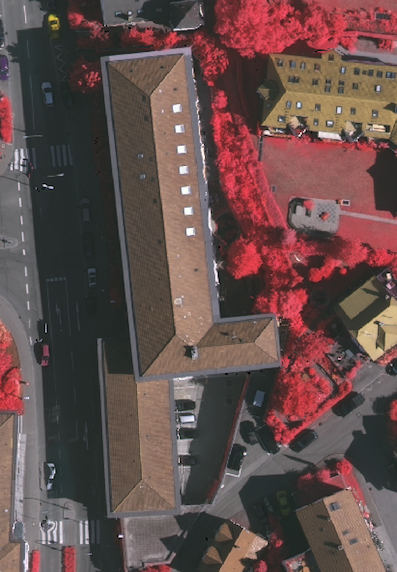}
		%		\caption{Building and ground truths.}\label{sfig:building_and_gt}
		\caption{}\label{sfig:building}
	\end{subfigure}
	\hspace{0.01cm}
	\begin{subfigure}[b]{0.3\linewidth}
		\centering
		\includegraphics[trim=0cm 0cm 0cm 0.5cm,clip,height=5.75cm]{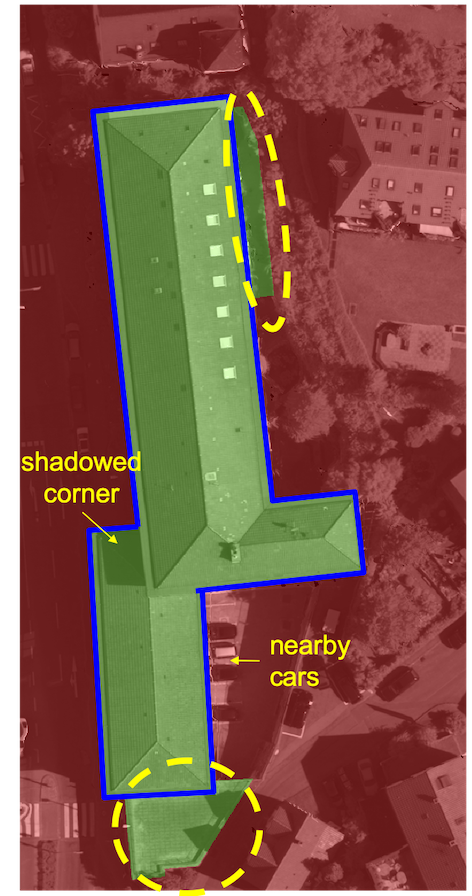}%compare_all_snakes_with_highlights}
%		\caption{Building and ground truths.}\label{sfig:building_and_gt}
		\caption{}\label{sfig:building_and_gt}
	\end{subfigure}
	\hspace{0.01cm}
	\begin{subfigure}[b]{0.34\linewidth}
		\centering
		\includegraphics[trim=0cm 0.6cm 0cm 0cm,clip,height=6.5cm]{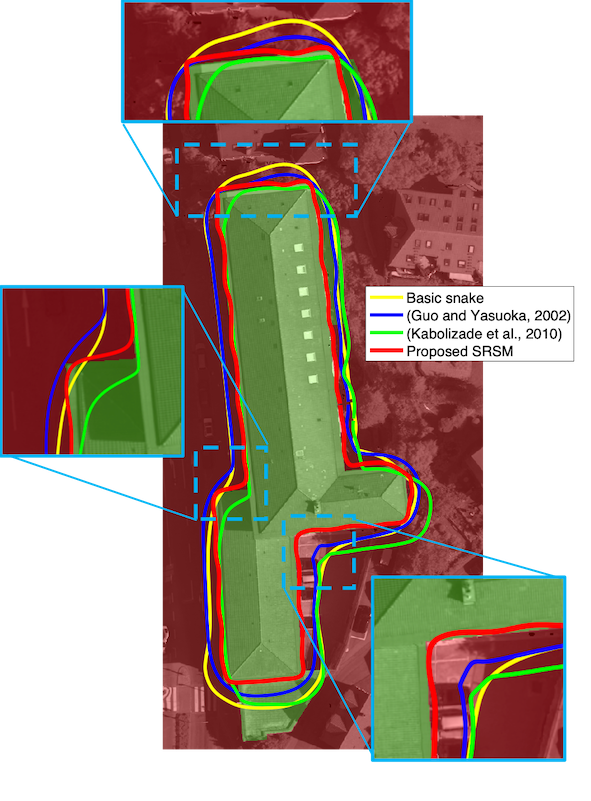}%compare_all_snakes_with_highlights}
%		\caption{Comparison among snake models.}\label{sfig:compare_all_snakes}
		\caption{}\label{sfig:compare_all_snakes}
	\end{subfigure}
%	\hspace{0.01cm}
%	\begin{subfigure}[b]{0.32\linewidth}
%		\centering
%		\includegraphics[trim=0cm 0cm 0cm 0cm,clip,height=6.5cm]{fig/11b_full_vert_new2}%w_poly_with_highlights}
%%		\caption{Before and after polygonization.}\label{sfig:snake_w_poly}
%		\caption{}\label{sfig:snake_w_poly}
%	\end{subfigure}
	
	\caption{Visual assessment of the proposed snake model on a gable-roof and complex shaped building. (\textbf{a}) Reference optical image of the considered building (\textbf{b}) The building with the ground truth region (in transparent green), and a modified ground truth boundary (in blue); % after removing two building parts in magenta-dashed circles; 
	(\textbf{c}) Visual comparison of performance among the snake models.}%; (\textbf{c}) Building boundary resulted by the SRSM with and without the polygonization.}
	\label{fig:snake_compare}
\end{figure}

We also perform an assessment on the performance of the proposed SRSM, and compare it with other existing snake models previously mentioned in sub-section \ref{ssec:review_snake_model}.
They are carried out on the gable-roof building previously discussed in sub-section \ref{sssec:img-term} and displayed in Figure \ref{fig:z_snake}.
First, the ground truth building region is overlaid by a transparent green area, while the surrounding ground is displayed in transparent red color, as  in Figure \ref{sfig:building_and_gt}.
These overlaying colors allow to assess the building extraction more straightforwardly.
Figure \ref{sfig:compare_all_snakes} presents the result of snake models on the exemplified building.
\newtextt{Four snake models are compared}, namely basic snake with GVF, snake model of Guo and Yasuoka \cite{guo2002snake}, snake model of Kabolizade \textit{et al.} \cite{kabolizade2010improved}, and the proposed SRSM. %snake model proposed in the present research work. 
\newtextt{They all are unsupervised snake models, which are substantially different from each other. 
	In addition, they are not constrained to one particular range of building sizes. 
	These also are the reasons why the snake model-based methods \cite{ahmadi2010automatic,fazan2013rectilinear,marcos2018learning} are not involved in this comparison.}
%In terms of snake parametrization, %we keep  $ \alpha $ and $ \beta $ as scalar value for every buildings.
The snake parameters are set as follows, $ \alpha = \beta = 0.2 $, balloon force magnitude $ \kappa = 0.1 $, and image-based energy term weights $ w_{line}=0.04, w_{edge}=2, w_{term}=0.01 $.

%\modif{It is worth-noting that three snake model-based methods are not concerned in this comparison, for different reasons.
%Firstly, since we are focusing on an unsupervised approach for snake models without a priori information of building gray-level, the method of Ahmadi \textit{et al.} \cite{ahmadi2010automatic} is not concerned.
%Secondly, the method proposed by Fazan and Dal Poz \cite{fazan2013rectilinear} does not involve a different energy function formulation than the basic snake model. 
%Lastly, as aforementioned, the CNN-based approach of Marcos \textit{et al.} \cite{marcos2018learning} involves only buildings of similar size.
%Hence, it requires a data preparation and a re-training of the CNN with samples constituted of buildings with similar size to the considered building.
%Therefore, this method is also not suitable for this comparison.}

Here, all concerned snake models are initialized by the same LiDAR-based building boundary.
These initial points are already an improvement compared to the ones proposed in the respective snake model.
However, one can remark in Figure \ref{sfig:compare_all_snakes} that the other snake models (i.e. the basic snake, the snake model of Guo and Yasuoka, and the snake model of Kabolizade \textit{et al.}) have problems approaching the true edges of the building.
On the other hand, the proposed SRSM, under the influence from salient features of the $ z $-image, converges very well towards the edges and the corners.
As we previously stated in \ref{sssec:img-term}, this exemplified building involves many challenging regions.
One of them is the building corner under shadow (circled in green in Figure \ref{sfig:3.4_4}), which is now shown in the left sub-figure in Figure \ref{sfig:compare_all_snakes}. Another difficult region is found with many nearby cars (circled in red in Figure \ref{sfig:3.4_4}). It is now zoomed in the bottom-right sub-figure in Figure \ref{sfig:compare_all_snakes}.
It is shown that, on these two corner regions, all three of the previous snake models yield poor results. 
In contrast, the proposed SRSM approaches very well the ground truth boundary.
%This stems from the fact that they operate on the optical image, which usually exhibits undesirable salient features. 
This visual assessment shows that the $ z $-image-based snake model yields much more accurate building boundary 
compared to other existing snake models. %, while the next best one is the our previous snake model \cite{nguyen2019unsupervised2}.

\begin{table}[h]
	\centering
	\caption{Quantitative results of snake models on the considered building.
		The best result for each metric (i.e. the smallest value for RMSE, and the greatest for \textit{Quality} $ Q $) is highlighted.}\label{tab:compare_all_snakes_iou}
	\tablesize{\footnotesize} 
	\begin{tabular}[b]{cccccc}
		\toprule
		& \multicolumn{2}{c}{\textbf{Benchmark ground truth}} &&  \multicolumn{2}{c}{\textbf{Modified ground truth}} \\
		\cline{2-3}\cline{5-6}\\[-7pt]
		\textbf{Model} & \textbf{Q} & \textbf{RMSE (m)} && \textbf{Q} & \textbf{RMSE (m)} \\ \midrule
		Basic snake model & 76.92 \% & 2.05 && 74.36 \% & 2.21 \\ \midrule
		%		GVF snake & 84.35 \% & 91.51 \% &  2.31 &&  92.48 \% &  96.09 \% & \\ \midrule
		Guo and Yasuoka \cite{guo2002snake} & 77.38 \% & 1.90  && 78.15 \% & 1.92 \\\midrule
		Kabolizade \textit{et al.} \cite{kabolizade2010improved}  & 79.66 \% &  2.08 &&  76.01 \% & 2.36 \\\midrule
%		Nguyen \textit{et al.} \cite{nguyen2019unsupervised2}  &  \underline{86.13 \%} &  1.94  &&  93.04 \% & 1.89 \\\midrule
		SRSM & \textbf{86.25 \%}  & \textbf{1.80} && \textbf{95.57 \%} & \textbf{1.75} \\%\midrule
%		SRSM (after polygonization)  & 85.71 \% & \textbf{0.73} &&  \underline{94.30 \%} & \textbf{0.66} \\
		\bottomrule
	\end{tabular}
\end{table}

%\begin{table}[h]
%	\centering
%	\caption{Quantitative results of snake models on the considered building.
%		The best result for each metric (i.e. the smallest value for RMSE, and the greatest for $ Q $) is highlighted, whereas the second best is underlined.}\label{tab:compare_all_snakes_iou}
%	\tablesize{\footnotesize} 
%	\begin{tabular}[b]{cccccccc}
%		\toprule
%		& \multicolumn{3}{c}{\textbf{Benchmark ground truth}} &&  \multicolumn{3}{c}{\textbf{Modified ground truth}} \\
%		\cline{2-4}\cline{6-8}\\[-7pt]
%		\textbf{Model} & \textbf{Q} & \textbf{F1} & \textbf{RMSE (m)} && \textbf{Q} & \textbf{F1} & \textbf{RMSE (m)} \\ \midrule
%		Basic snake model & 76.92 \% & 86.95 \% & 2.05 && 74.36 \% & 85.29 \% & 2.21 \\ \midrule
%		%		GVF snake & 84.35 \% & 91.51 \% &  2.31 &&  92.48 \% &  96.09 \% & \\ \midrule
%		Snake model with balloon force & 83.13 \% & 90.79 \% & 2.11  && 88.42 \% & 93.85 \% & 2.10 \\\midrule
%		Kabolizade \textit{et al.} \cite{kabolizade2010improved}  & 79.66 \% & 88.68 \% &  2.08 &&  76.01 \% &  86.37 \% & 2.36 \\\midrule
%		Nguyen \textit{et al.} \cite{nguyen2019unsupervised2}  &  \underline{86.13 \%} & \underline{92.55 \%}  &  1.94  &&  93.04 \% &  96.39 \% & 1.89 \\\midrule
%		Proposed method (before registration)  & \textbf{86.25 \%}  & \textbf{92.62 \%}  & 1.80 && \textbf{95.57 \%}  &  \textbf{97.73 \%}  & 1.75 \\\midrule
%		Proposed method (after registration)  & 85.71 \% & 92.30 \% & \textbf{0.73} &&  \underline{94.30 \%} &  \underline{97.06 \%} & \textbf{0.66} \\
%		\bottomrule
%	\end{tabular}
%\end{table}

Table \ref{tab:compare_all_snakes_iou} summarizes the quantitative results of the compared snake models, based on the area-based \textit{Quality} and RMSE metrics. 
%These metrics are defined in sub-section \ref{ssec:metric}.
First of all, one can remark that, in general, the resulting \textit{Quality} and RSME from all snake models are relatively high (more than 70\%). 
This stems from the benefit of using the LiDAR-based building boundary as initial points. % for the snake models.
Then, the quantitative results also show the relevance of the proposed snake models, compared to other snake models.
However, the margin of gain between them according to these values---i.e. a maximum \textit{Quality}  gain of 9.33\% between the basic and the SRSM---is not as high as expected, given the clear advantage drawn from the visual assessment from Figure \ref{sfig:compare_all_snakes}.

%\modif{Then, we also illustrate the building extraction result before and after the polygonization. 
%Figure \ref{sfig:snake_w_poly} shows the proposed snake model before and after the polygonization step. 
%As one can see, the boundary regularization algorithm \cite{gribov2017searching} provides a very accurate polygonized building boundary, especially at the building corners as shown in the sub-figures (Figure \ref{sfig:snake_w_poly}).
%Quantitatively, the polygonized boundary results in a \textit{Quality} of 85.71\%, and a RMSE of 0.73 meters. 
%Even though the resulting \textit{Quality} is shown to decrease slightly after the polygonization, the geometric accuracy given by the RMSE is considerably improved.
%Indeed, the RMSE is reduced from 1.80 to 0.73 meters, in other words, a reduction of 59.44\%.}

One can also remark that the snake models were not able to extract two particular parts of the building \newtextttt{(highlighted by yellow-dashed circles in Figure \ref{sfig:building_and_gt})} because they do not exhibit significant elevation change \newtexttt{or color change from the surrounding ground (cf. Figure \ref{fig:z_snake}).  
On one hand, the inability of the SRSM stems from the absence of elevation changes. On the other hand, the other snake models are unable to extract these parts because of the absence of color changes.}
%These elevation indifferences can be seen previously in Figure \ref{sfig:3.4_0}.
%As a result, these parts are not detected by the preliminary extraction from the LiDAR point cloud and the $ z $-image-based snake model.
We are convinced that these undetected parts can be the reason of the low margin between the snake models mentioned above. 
%Indeed, for the SRSM that operates on the $ z $-image, such parts are invisible because of no elevation change, while the other snake models operate on the optical image, 
Therefore, we also conduct another evaluation of all four snake models with a modified version of the ground truth building boundary, in which the two undetected parts are removed.
\newtexttt{Such modified ground truth boundary aims to provide the unbiased reference for the snake models.}
In Figure \ref{sfig:building_and_gt}, this modified ground truth boundary is depicted in blue outlines.
The columns 4 and 5 of Table \ref{tab:compare_all_snakes_iou} reveals the involved comparison based on this modified ground truth.
As expected, the new margin between the proposed snake model and the others is now much larger, i.e. a margin of 21.21\% of \textit{Quality} between the basic snake model and the proposed SRSM.
It is coherent with the inference drawn from the visual assessment (Figure \ref{sfig:compare_all_snakes}).
%Also, the same trade-off behavior can be noted after having applied the polygonization, where the resulting \textit{Quality} slightly decreases, but the RMSE is reduced significantly.
%Figure \ref{sfig:snake_w_poly} shows the proposed SRSM before and after the polygonization step. 
%The resulting polygon depicts accurately the true building boundary, especially at the building corners as shown in the sub-figures.
This comparison has shown that the proposed SRSM yields better accuracy than the other snake models.
In the next two sub-sections, the overall performance of the SRSM on different datasets will be assessed.
%the experimental results  will be presented.

\subsection{Performance on ISPRS Vaihingen dataset}
\begin{figure}[h]
	\centering
	%	\subfloat[Area 1]{\frame{}\label{subfig:area1}}
	%	\subfloat[Area 2]{\frame{\includegraphics[trim=8.5cm 5cm 8.5cm 5cm,clip,height=5cm]{fig/isprs_area2.png}}\label{subfig:area2}}
	%	\subfloat[Area 3]{\frame{\includegraphics[trim=8.5cm 5cm 8.5cm 5cm,clip,height=5cm]{fig/isprs_area3.png}}\label{subfig:area3}}
%	\begin{subfigure}{0.31\textwidth}
%		\centering
%		\includegraphics[trim=11cm 6cm 9.5cm 5cm,clip,height=5cm]{fig/isprs_area1.png}
%		\caption{Area 1 (with 37 buildings)}
%		\label{subfig:area1}
%	\end{subfigure}
%	\begin{subfigure}{0.31\textwidth}
%		\centering
%		\includegraphics[trim=8.5cm 5cm 8.5cm 5cm,clip,height=5cm]{fig/isprs_area2.png}
%		\caption{Area 2 (with 14 buildings)}
%		\label{subfig:area2}
%	\end{subfigure}
%	\begin{subfigure}{0.31\textwidth}
%		\centering
%		\includegraphics[trim=8.5cm 5cm 8.5cm 5cm,clip,height=5cm]{fig/isprs_area3.png}
%		\caption{Area 3 (with 56 buildings) }
%		\label{subfig:area3}
%	\end{subfigure}
%	
%	\vspace{0.25cm}
	
	\begin{subfigure}{0.31\textwidth}
		\centering
		\includegraphics[trim=0 0 0 0,clip,height=4.5cm]{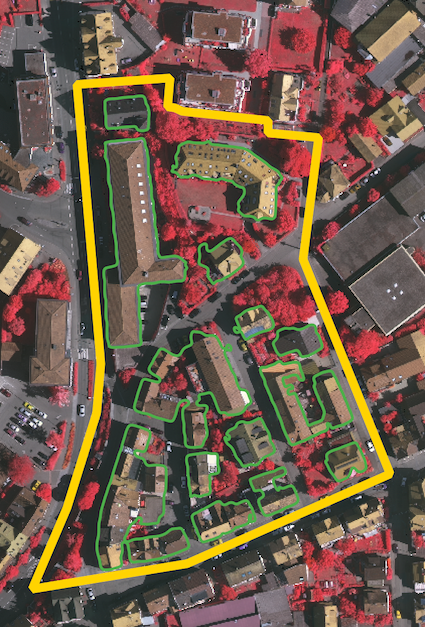}
		\caption{Area 1 (with 37 buildings)}
		\label{subfig:area1}
	\end{subfigure}
	\begin{subfigure}{0.31\textwidth}
		\centering
		\includegraphics[trim=0 0 0 0,clip,height=4.5cm]{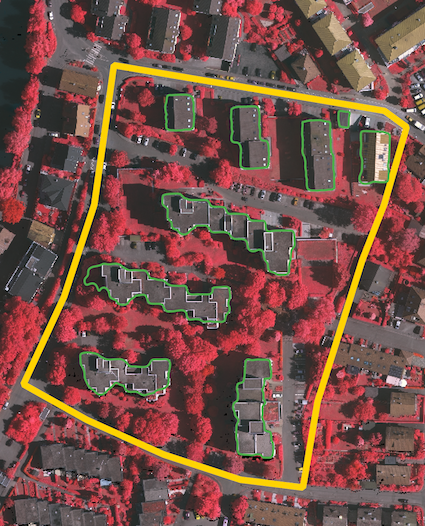}
		\caption{Area 2 (with 14 buildings)}
		\label{subfig:area2}
	\end{subfigure}
	\begin{subfigure}{0.31\textwidth}
		\centering
		\includegraphics[trim=0 0 0 0,clip,height=4.5cm]{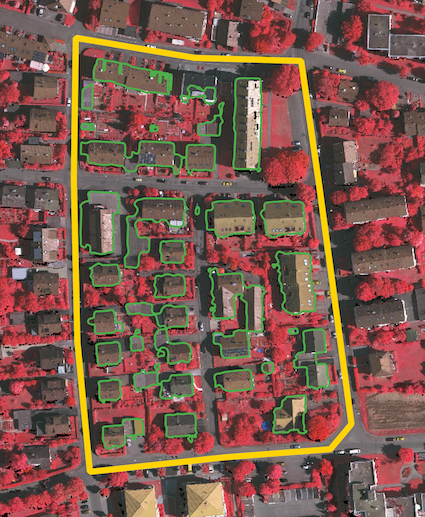}
		\caption{Area 3 (with 56 buildings) }
		\label{subfig:area3}
	\end{subfigure}
	
	\vspace{0.25cm}
	
	\begin{subfigure}{0.31\textwidth}
		\centering
		\includegraphics[trim=10cm 4cm 13cm 1cm,clip,height=4cm]{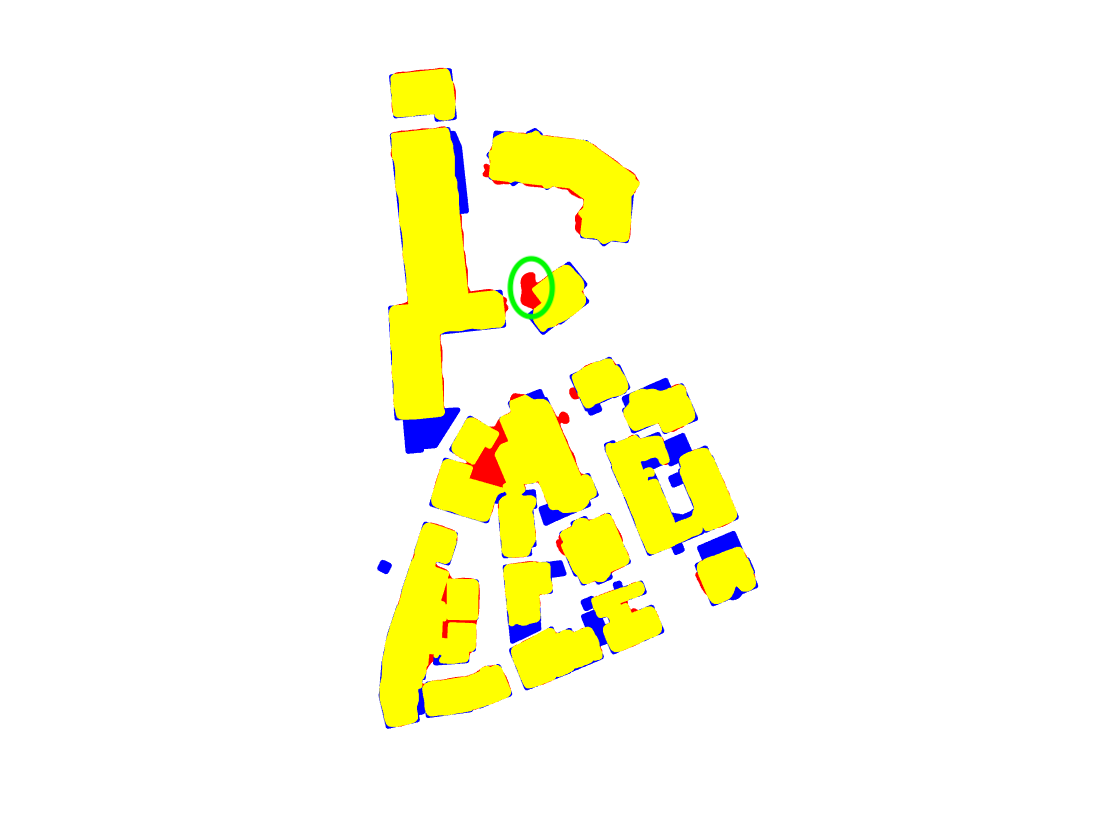}
		%		\caption{Result on Area 1 (before polygonization)}
		\caption{}
		\label{subfig:area1_new3}
	\end{subfigure}\hspace{0.1cm}
	\begin{subfigure}{0.31\textwidth}
		\centering
		\includegraphics[trim=8cm 3cm 7cm 0cm,clip,height=4cm]{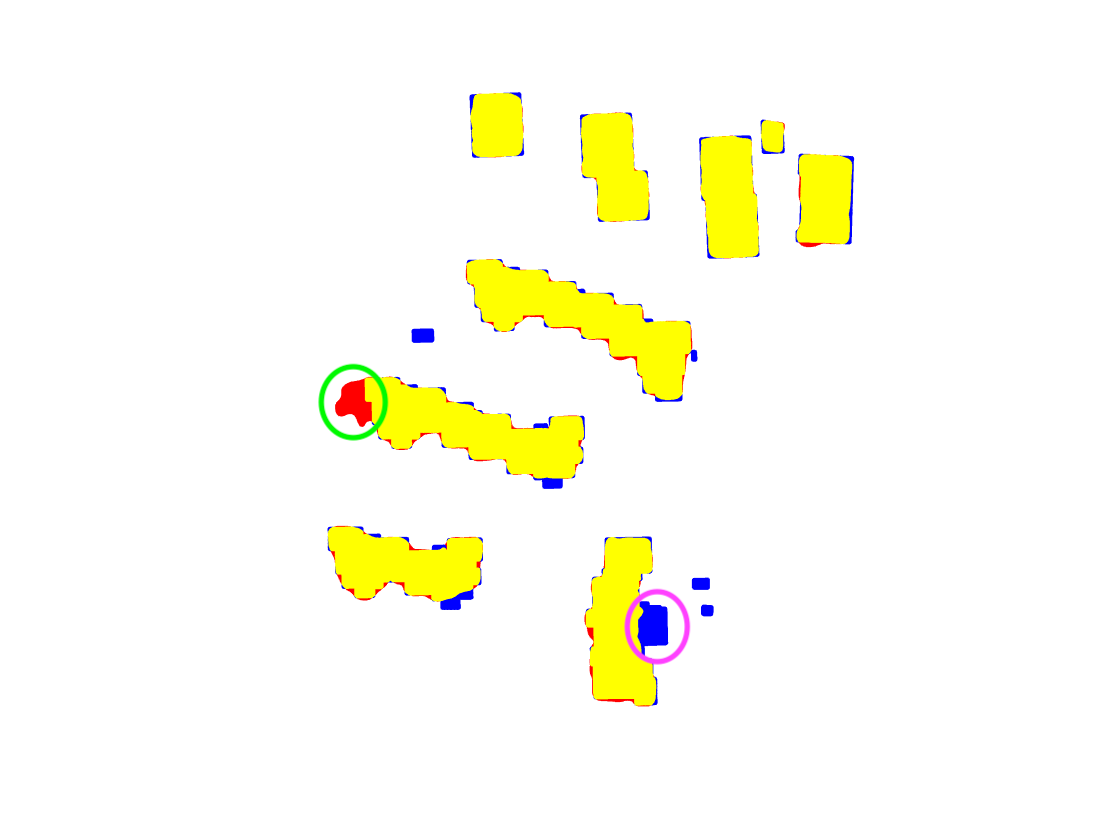}
		%		\caption{Result on Area 2 (before polygonization)}
		\caption{}
		\label{subfig:area2_new3}
	\end{subfigure}\hspace{0.1cm}
	\begin{subfigure}{0.31\textwidth}
		\centering
		\includegraphics[trim=6cm 4cm 7cm 1cm,clip,height=4cm]{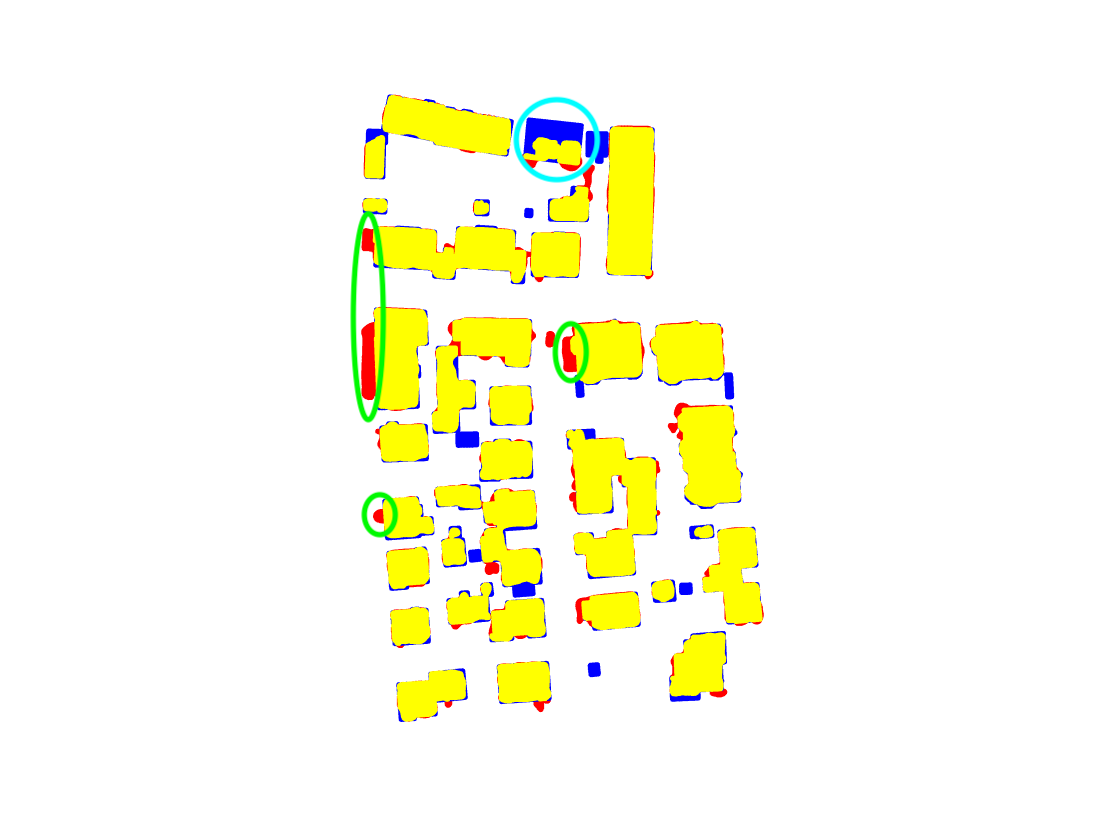}
		%		\caption{Result on Area 3 (before polygonization)}
		\caption{}
		\label{subfig:area3_new3}
	\end{subfigure}
	\caption{Area-based assessment on test areas in ISPRS Vaihingen benchmark dataset. (\textbf{a}-\textbf{c}) Areas 1-3, with the SRSM results in green outlines; (\textbf{d}-\textbf{f}) SRSM results on areas 1-3 with respect to their ground truth. Yellow, red and blue pixels, respectively, represent the TP, FP and FN pixels.}
	\label{fig:isprs_areas}
\end{figure}

\begin{figure}[h]
	\centering
	\begin{subfigure}{0.3\linewidth}
		\centering\includegraphics[trim=3cm 1.25cm 3cm 0.75cm,clip,height=3cm]{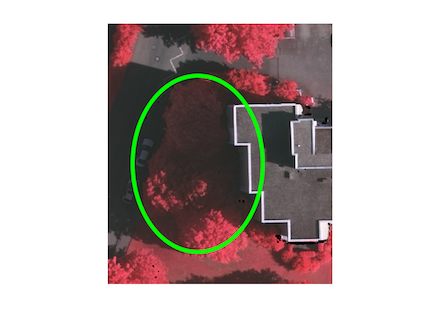}\caption{}\label{sfig:shadowed_vege}
	\end{subfigure}
	\begin{subfigure}{0.3\linewidth}
		\centering\includegraphics[trim=3cm 1.25cm 3cm 0.75cm,clip,height=3cm]{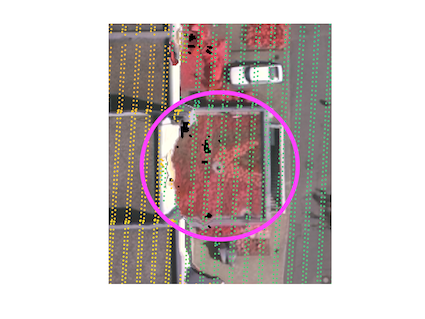}\caption{}\label{sfig:basement_vege}
	\end{subfigure}
	\begin{subfigure}{0.3\linewidth}
		\centering\includegraphics[trim=3cm 1.25cm 3cm 0.75cm,clip,height=3cm]{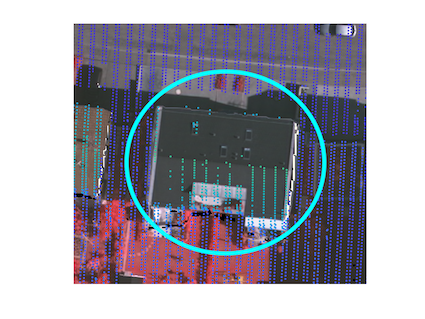}\caption{}\label{sfig:incomplete_building}
	\end{subfigure}
	\caption{Examples of the problems unresolved by the SRSM. (\textbf{a}) Shadowed vegetation next to a building; (\textbf{b}) The roof of a basement covered by vegetation which is of similar elevation with its surrounding area; (\textbf{c}) The building in area 3 with very few LiDAR returns.}
\end{figure}

The three test areas of the ISPRS Vaihingen dataset are shown by Figure \ref{fig:isprs_areas}. %\ref{subfig:area1}, \ref{subfig:area2} and \ref{subfig:area3}.
Area 1 (Figure \ref{subfig:area1}) is situated in the center of the city and characterized by dense construction consisting of historic buildings with rather complex shapes. %, along with roads and  trees. 
Area 2 (Figure \ref{subfig:area2})  is composed of high-rise residential buildings surrounded by trees. 
Lastly, area 3 (Figure \ref{subfig:area3}) is residential with detached houses and many surrounding trees. 
%The proposed method is performed on all three test areas.
%Extracted boundaries are then evaluated with the ground truth building boundaries, provided in the benchmark dataset.
%The performance of the building extraction method is evaluated without the polygonization step.
The results of SRSM also are depicted in Figure \ref{subfig:area1}, \ref{subfig:area2} and \ref{subfig:area3} in green.
Then Figure \ref{subfig:area1_new3}, \ref{subfig:area2_new3} and \ref{subfig:area3_new3} illustrate the area-based accuracy assessment, denoting TP (in yellow), FP (in red), and FN (in blue) pixels.
Overall, the proposed method yields a very high accuracy, reflecting by very high number of TPs on all three areas. 
However, a number of unresolved problems can be remarked in Figure \ref{fig:isprs_areas}.
Firstly, many FP pixels can still be noted in all three areas.
They relate to the problem of shadowed tree regions near buildings. Such tree regions are circled in green in Figures \ref{subfig:area1_new3}, \ref{subfig:area2_new3} and \ref{subfig:area3_new3}.
An example of this problem is from area 2, which is shown by Figure \ref{sfig:shadowed_vege}.
Secondly, several small buildings from all three areas have not been detected. 

Table \ref{tab:ISPRS_accuracy_pixel} summarizes the area-based accuracy assessment result on all three test areas.
In averaging on three areas, the proposed SRSM achieves a \textit{Quality} of 86.57\%, a \textit{Completeness} of 91.63\% and a \textit{Correctness} of 93.99\%. 
%%The geometrical accuracy assessment is carried out based on the snake model followed by the polygonization.
The last column of Table \ref{tab:ISPRS_accuracy_pixel} shows the resulting geometrical accuracy of the SRSM. % followed by the polygonization \cite{gribov2017searching}.
It can be noted that the RMSE on the area 3 is much lower than the other areas. 
It is due to the fact that area 3 are composed mostly rectangular buildings and less complex than the other areas. % where the building shape is more complex.
%Such accuracy measurements will be compared with other existing building extraction methods.

\begin{table}[H]
	\centering
	\caption{Area-based accuracy of the SRSM on the ISPRS Vaihingen benchmark dataset, and geometrical accuracy of the SRSM after the polygonization.}
	\label{tab:ISPRS_accuracy_pixel}
	\begin{tabular}{ccccc}
		\toprule
		\textbf{Area} & $ \mathbf{Cp} $ & $ \mathbf{Cr} $ & $ \mathbf{Q} $ %& $ \mathbf{F1} $ 
		& \textbf{RMSE}\\\midrule
		1 & 90.42 \% & 94.20 \% & 85.65 \% % & 92.27 \% 
		& 1.24 \\%\midrule
		2 & 93.47 \% & 94.75 \% & 88.87 \% %& 94.11 \% 
		& 1.11\\%\midrule
		3 & 91.00 \% & 93.02 \% & 85.18 \% %& 92.00 \% 
		& 0.92\\\midrule
		Average & 91.63 \% & 93.99 \% & 86.57 \% %& 92.79 \% 
		& 1.09\\
		\bottomrule
	\end{tabular}
\end{table}
%		\textbf{RMSE} & 1.24 & 1.11 & 0.92  & 1.09 \\

Table \ref{tab:ISPRS_accuracy} presents the resulting object-based accuracy of the SRSM.
The columns two to four show the accuracy metrics on all objects, whereas the columns five to seven provide the metrics when considering only objects with an area larger than  50 m$ ^2 $. 
The differences between these two results reflect the aforementioned problem of undetected small buildings.
%It is highlighted by the differences between the resulting accuracy on all objects and on only objects of an area larger than  50 m$ ^2 $. 

\begin{table}[H]
	\centering
	\caption{Object-based accuracy of the SRSM on the ISPRS Vaihingen benchmark dataset for all buildings (columns 2 to 4), and for buildings with an area larger than 50 square meters (columns 5 to 7).}
	\label{tab:ISPRS_accuracy}
	\begin{tabular}{ccccccc}
		\toprule
		%		& \multicolumn{3}{c}{\textbf{Object-based metrics}} & \multicolumn{3}{c}{\textbf{Object-based metrics (>50m$ ^2 $)}}\\
		\textbf{Area} & $ \mathbf{Cp} $ & $ \mathbf{Cr} $ & $ \mathbf{Q} $ %& $ \mathbf{F1} $  
		& $ \mathbf{Cp_{50}} $ & $ \mathbf{Cr_{50}} $ & $ \mathbf{Q_{50}} $ %& $ \mathbf{F1_{50}} $ 
		\\\midrule
		1 & 83.78 \% & 100 \% & 83.78 \% %& 91.17 \% 
		& 100 \% & 100 \% & 100 \% %& 100 \% 
		\\%\midrule
		2 & 78.57 \% & 100 \% & 78.57 \% %& 88.00 \% 
		& 100 \% & 100 \% & 100 \% %& 100 \%
		\\%\midrule
		3 & 83.93 \% & 97.92 \% & 82.46 \% %& 90.39 \% 
		& 97.30 \% & 100 \% & 97.30 \% %& 98.63 \%
		\\ \midrule
		Average & 82.09 \% & 99.31 \% & 81.60 \% %& 89.85 \% 
		& 99.10 \% & 100 \% & 99.10 \% %& 99.54 \% 
		\\
		\bottomrule
	\end{tabular}
\end{table}

In addition, several buildings have only been partially extracted, due to the fact that the non-extracted parts have very similar elevation to their surrounding area.
Particularly, one of these building parts (circled in magenta in Figure \ref{subfig:area2_new3}), in reality, is the roof of a basement covered by vegetation in the area 2, as shown in Figure \ref{sfig:basement_vege}.
This part has not been extracted by any existing building extraction methods submitted to this benchmark \cite{rottensteiner2014results}.
In the area 3, there is also one building (circled in cyan in Figure \ref{subfig:area3_new3}) that is very poorly extracted.
This problem is caused by the incompleteness of LiDAR data on this building, as shown in Figure \ref{sfig:incomplete_building}. 

%% file: SR_eval.tex
\subsection{Performance evaluation of the super-resolution}\label{ssec:sr_eval}
\begin{figure}[h]
	\centering
	\begin{subfigure}{0.35\linewidth}
		\centering
		\includegraphics[trim=1cm 5cm 0cm 3cm,clip,width=\linewidth]{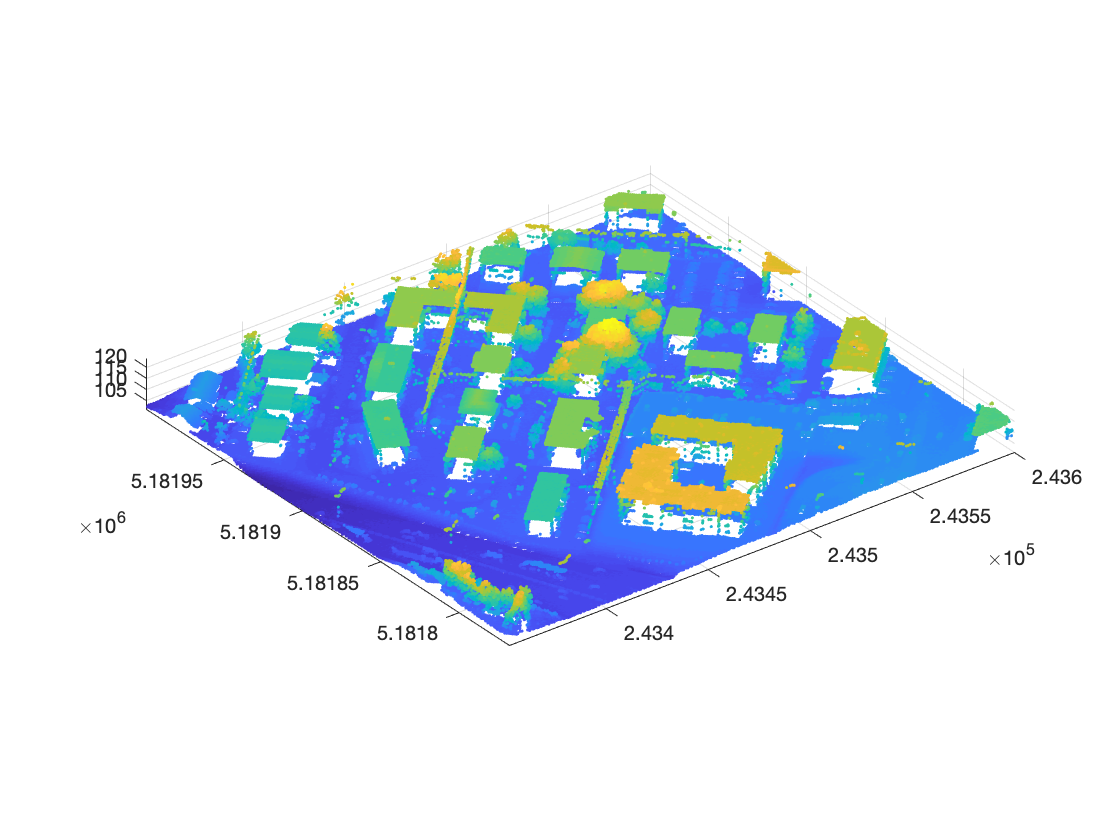}\caption{Original 3-D point cloud}\label{sfig:SR_full_pcl}
	\end{subfigure}
	\begin{subfigure}{0.35\linewidth}
		\centering
		\includegraphics[trim=1cm 5cm 0cm 3cm,clip,width=\linewidth]{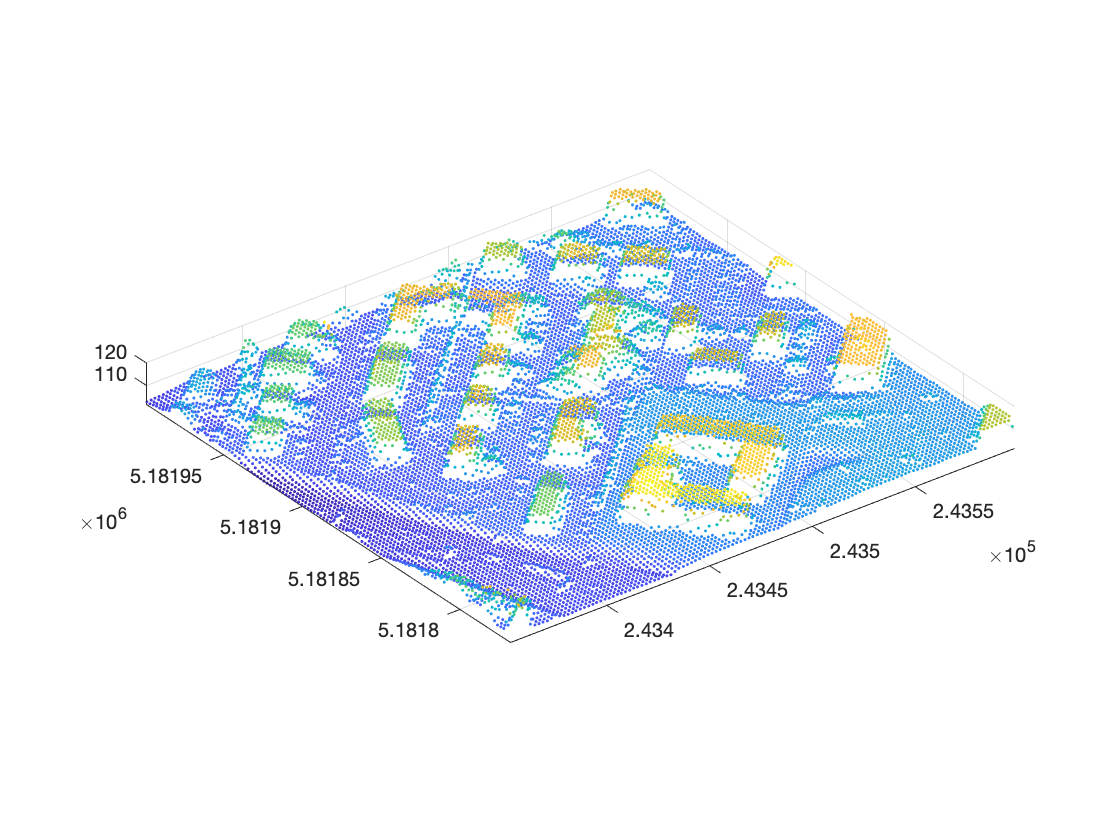}\caption{Subsampled 3-D point cloud}\label{sfig:SR_sub_pcl}
	\end{subfigure}

	\vspace{0.15cm}
	\begin{subfigure}{0.2\linewidth}
		\centering
		\includegraphics[trim=6cm 1.5cm  6cm 1cm,clip,width=\linewidth]{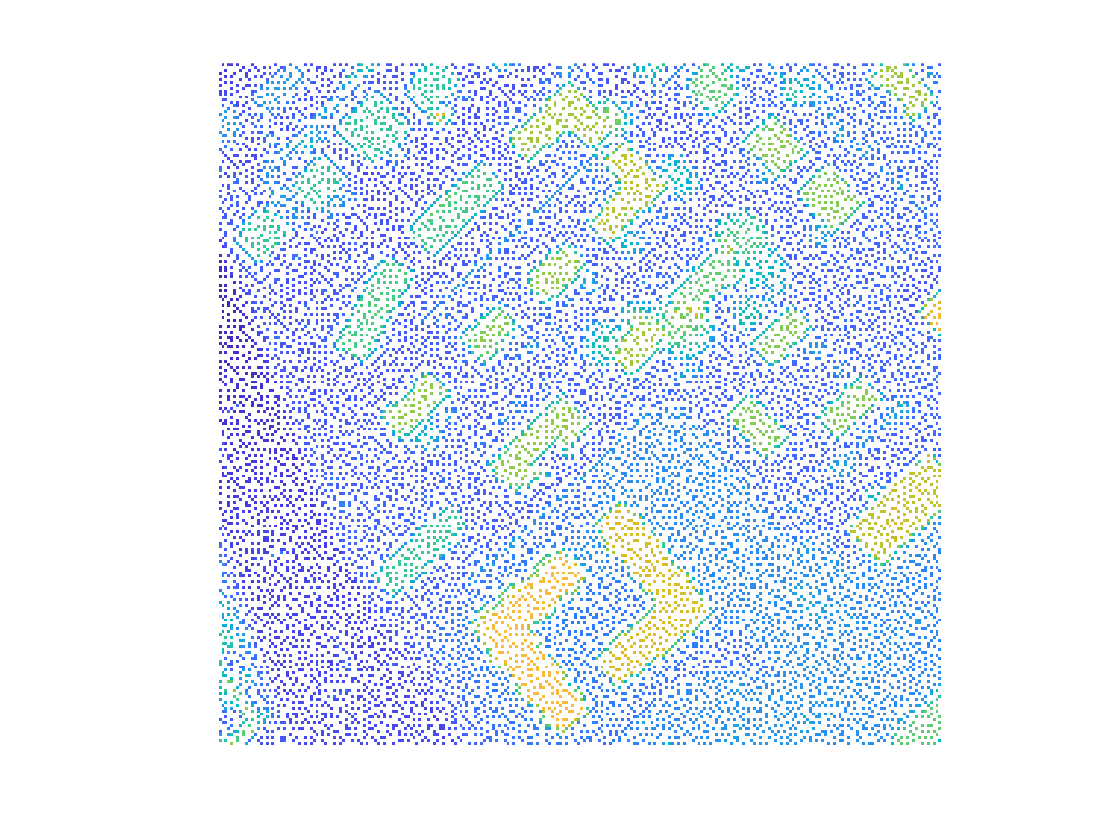}\caption{Sparse DSM}\label{sfig:SR_sub_DSM}
	\end{subfigure}
	\begin{subfigure}{0.2\linewidth}
		\centering
		\includegraphics[trim=6cm 1.5cm  6cm 1cm,clip,width=\linewidth]{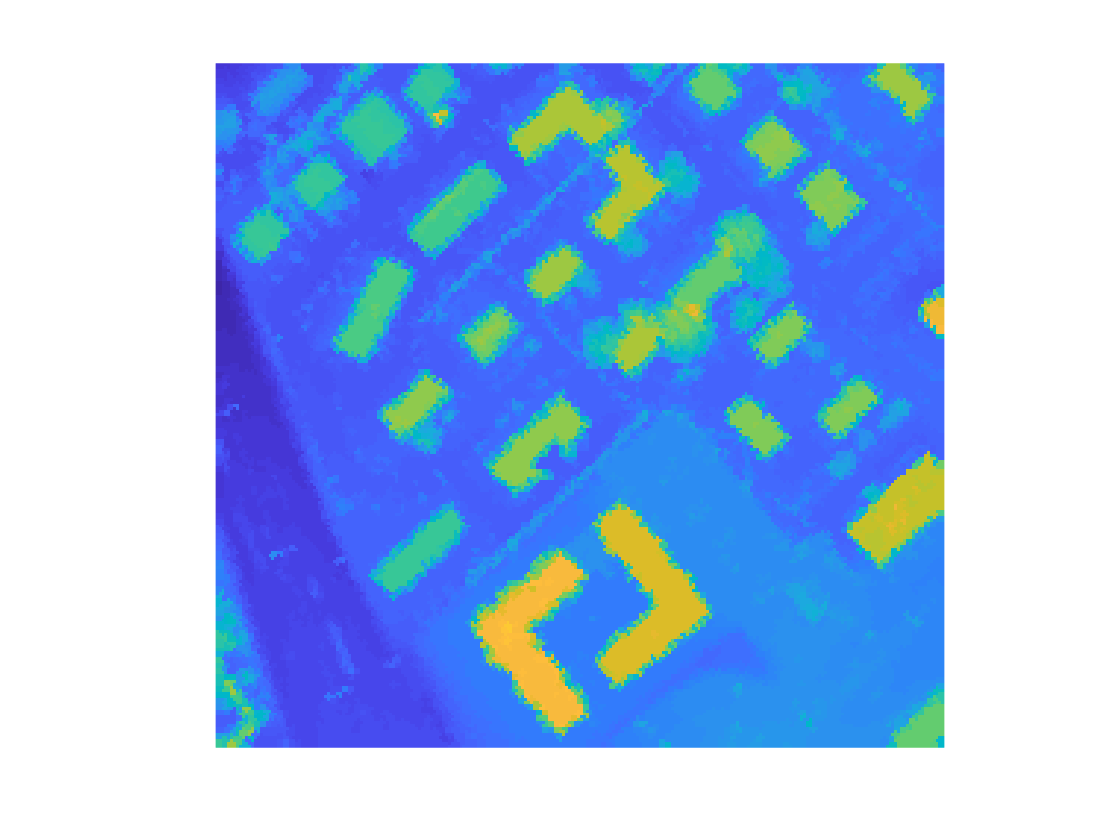}\caption{NN}\label{sfig:SR_NN}
	\end{subfigure}
%	\hspace{0.01cm}
%	\begin{subfigure}{0.15\linewidth}
%	\centering
%	\includegraphics[trim=4cm 1.5cm  4cm 1cm,clip,width=\linewidth]{fig/SR_linear}\caption{Bilinear}
%	\end{subfigure}	
%	\hspace{0.01cm}
%	\begin{subfigure}{0.15\linewidth}
%	\centering
%	\includegraphics[trim=5cm 1.5cm  5cm 1cm,clip,width=\linewidth]{fig/SR_natural}\caption{Natural}
%	\end{subfigure}	
	\hspace{0.01cm}
	\begin{subfigure}{0.2\linewidth}
		\centering
		\includegraphics[trim=6cm 1.5cm  6cm 1cm,clip,width=\linewidth]{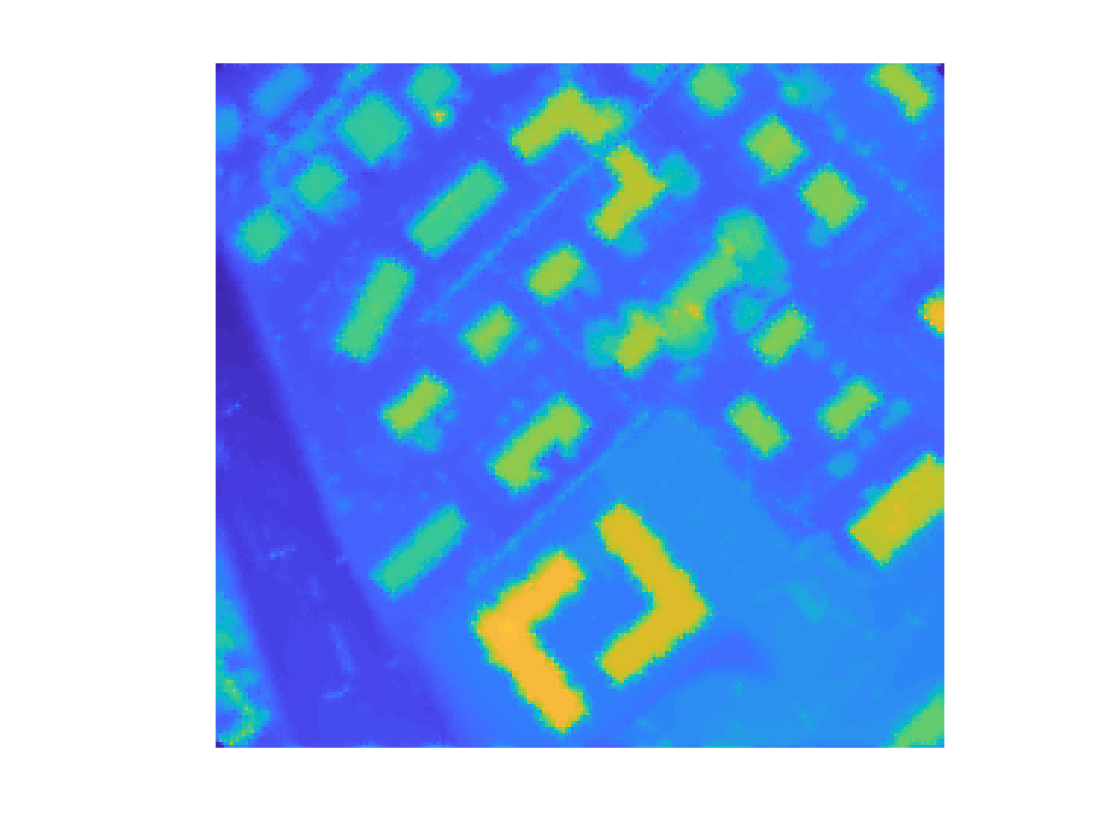}\caption{Proposed SR}\label{sfig:SR_SR}
	\end{subfigure}
	\hspace{0.01cm}
	\begin{subfigure}{0.2\linewidth}
		\centering
		\includegraphics[trim=6cm 1.5cm  6cm 1cm,clip,width=\linewidth]{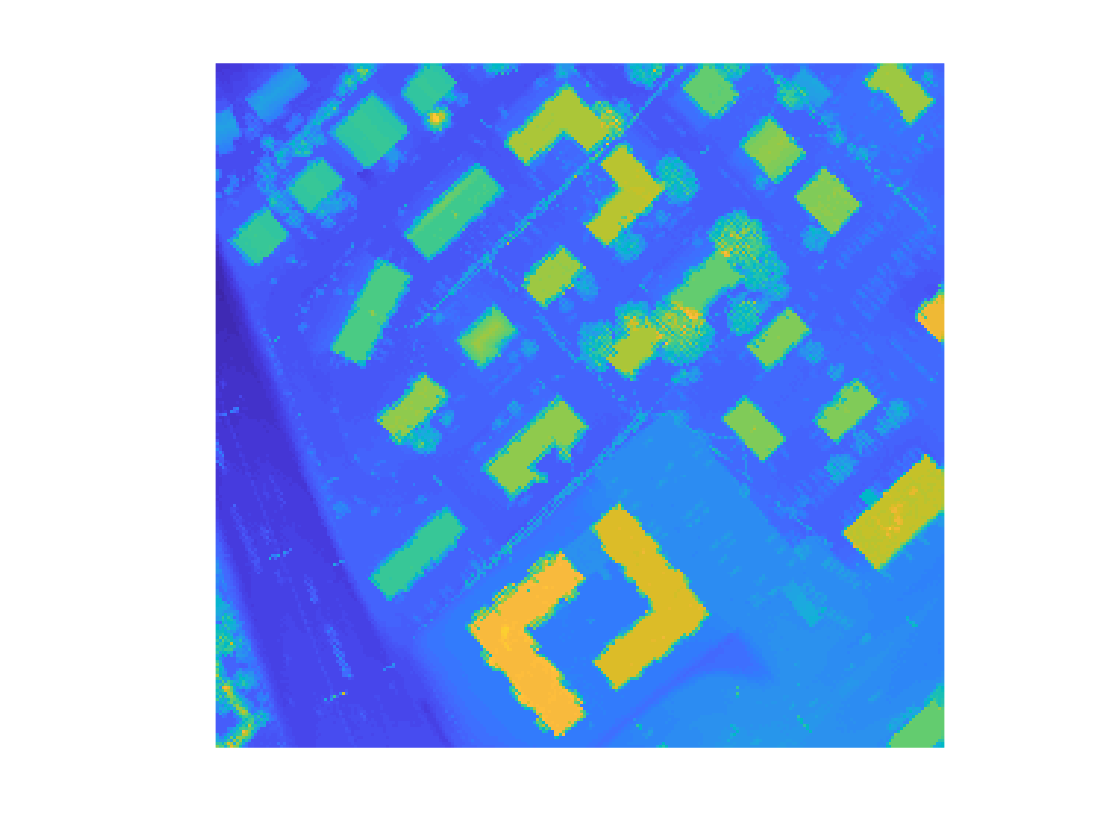}\caption{Ground truth}\label{sfig:SR_GT}
	\end{subfigure}
%
%	\vspace{0.15cm}
%	\begin{subfigure}{0.15\linewidth}
%		\centering
%		\includegraphics[trim=6cm 1.5cm  6cm 1cm,clip,width=\linewidth]{fig/SR_NN_small}\caption{NN}
%	\end{subfigure}
%	\hspace{0.01cm}
%	\begin{subfigure}{0.15\linewidth}
%		\centering
%		\includegraphics[trim=6cm 1.5cm  6cm 1cm,clip,width=\linewidth]{fig/SR_SR_small}\caption{Proposed SR}
%	\end{subfigure}
%	\hspace{0.01cm}
%	\begin{subfigure}{0.15\linewidth}
%		\centering
%		\includegraphics[trim=6cm 1.5cm  6cm 1cm,clip,width=\linewidth]{fig/SR_GT_small}\caption{Ground truth}
%	\end{subfigure}

	\caption{Illustration of the conducted assessment. (\textbf{a}) The original LiDAR 3-D point cloud; (\textbf{b}) the subsampled 3-D point cloud (by a factor of 2); (\textbf{c}) sparse DSM generated from (b); (\textbf{d}) result of the NN interpolation; (\textbf{e}) result of the proposed SR; (\textbf{f}) the ground truth DSM generated from (a).}
	\label{fig:SR_eval}
\end{figure}

Beside the visual assessment provided in \ref{ssec:sr}, the performance of the proposed SR process is also quantitatively evaluated.
We compare it with other conventional 2-D interpolation methods, namely nearest neighbor (NN), bilinear and natural interpolation \cite{sibson1981brief}. 
This evaluation and comparison are depicted by Figure \ref{fig:SR_eval}. 
The four methods are examined on a real LiDAR point cloud with an average density of 3.8 points/m$ ^2 $ (Figure \ref{sfig:SR_full_pcl}). 
Such point cloud is then subsampled by a chosen factor, namely 2, 4 and 8, yielding a subsampled point cloud which serves as an input for these SR/interpolation methods.
\newtexttt{These experimented factors are chosen based on the proportion between the respective spatial resolution of the datasets (cf. Table \ref{tab:datasets_ISPRS}).}
For example, Figure \ref{sfig:SR_sub_pcl} depicts the 3-D point cloud subsampled by a factor of 2. 
Based on the sparse DSM generated from this subsampled point cloud (Figure \ref{sfig:SR_sub_DSM}), each interpolation method generates a DSM having the spatial resolution equal to that of the subsampled LiDAR point cloud times the upscaling factor---in other words, equivalent to the spatial resolution of the original point cloud.
The resulting interpolated DSM provided by each method (e.g. Figure \ref{sfig:SR_NN} or \ref{sfig:SR_SR}) is compared with the DSM generated from the full-resolution LiDAR point cloud, which is considered as the ground truth for the assessment (Figure \ref{sfig:SR_GT}).
%The assessment is carried out for three upscaling factors, .

%\begin{table}[H]
%	\caption{Performance evaluation of the super-resolution process.}
%	\label{tab:SR_eval}
%	\centering
%	\tablesize{\footnotesize} %% You can specify the fontsize here, e.g., \tablesize{\footnotesize}. If commented out \small will be used.
%	\begin{tabular}{cccccccccccc}
%		\toprule
%		& \multicolumn{3}{c}{$ \times 2 $} && \multicolumn{3}{c}{$ \times 4 $} && \multicolumn{3}{c}{$ \times 8 $} \\
%		\cline{2-4}\cline{6-8}\cline{10-12}\\[-5pt]
%		Method & MSE & SSIM & PSNR && MSE & SSIM & PSNR && MSE & SSIM & PSNR \\
%		\midrule
%		Nearest & 2.20 & 0.72 & -3.42 && 4.66 & 0.50 & -6.68 && 9.81 & 0.26 & -9.92\\
%		Bilinear & 0.82 & 0.85 & 0.84 && 2.40 & 0.63 & -3.80 && 6.19 & 0.35 & -7.92 \\
%		Bicubic & 0.80 & 0.85 & 0.97 && 2.38 & 0.63 & -3.76 && 6.32 & 0.37 & -8.01\\
%%		\midrule
%		Proposed SR & 0.96 & 0.82 & 0.17 && 3.21 & 0.54 & -5.06 && 8.07 & 0.25 & -9.07 \\
%		\bottomrule
%	\end{tabular}
%\end{table}

In order to evaluate the quality of these interpolation and SR methods---\newtextt{i.e. the closeness between the interpolated image and the ground truth image}---we measure the following metrics: root-mean-square error (RMSE),  structural similarity (SSIM) \cite{wang2004image}, and the peak signal-to-noise ratio (PSNR). 
SSIM and PSNR are two widely used objective metrics for evaluating image super-resolution quality \cite{hore2010image}.
%\newtextt{They allows to assess the reliability of the }
Their mathematical explanations can be found in Appendix \ref{app:psnr_ssim}. 
Table \ref{tab:SR_eval} summarizes the quality measurements of each interpolation method for all three upscaling factors, i.e. $ \times 2, \times 4 $ and $ \times 8 $.

\begin{table}[h]
	\caption{Performance evaluation of the SR process. The best result for each upscaling factor and each metric (i.e. the smallest value for RMSE, and the greatest for SSIM and PSNR) is highlighted, whereas the second best is underlined.}
	\label{tab:SR_eval}
	\centering
	\tablesize{\footnotesize} %% You can specify the fontsize here, e.g., \tablesize{\footnotesize}. If commented out \small will be used.
	\begin{tabular}{cccccccccccc}
		\toprule
		& \multicolumn{3}{c}{$ \mathbf{\times 2} $} && \multicolumn{3}{c}{$ \mathbf{\times 4} $} && \multicolumn{3}{c}{$ \mathbf{\times 8} $} \\
		\cline{2-4}\cline{6-8}\cline{10-12}\\[-5pt]
		\textbf{Method} & \textbf{RMSE} & \textbf{SSIM} & \textbf{PSNR (dB)} && \textbf{RMSE} & \textbf{SSIM} & \textbf{PSNR (dB)} && \textbf{RMSE} & \textbf{SSIM} & \textbf{PSNR (dB)} \\
		\midrule
		NN & 2.18 & 0.40 & -6.76 && 2.47 & 0.30 & -7.85 && \underline{3.08} & 0.18 & \underline{-9.76} \\
		Bilinear & 2.08 & 0.37 & -6.36 && 2.41 & \underline{0.34} & -7.65 && 4.39 & \underline{0.24} & -12.86 \\
		Natural & \underline{2.00} & \underline{0.40} & \underline{-6.03} && \underline{2.34} & \textbf{0.36} & \underline{-7.40} && 4.33 & \textbf{0.25} & -12.74 \\
		Proposed SR & \textbf{1.96} & \textbf{0.40} & \textbf{-5.83} && \textbf{2.04} & {0.33} & \textbf{-6.21} && \textbf{2.80} & 0.19 & \textbf{-8.94} \\
		\bottomrule
	\end{tabular}
\end{table}

Overall, compared to the other methods, the proposed SR process yields better results, i.e. smaller RMSE, higher SSIM and PSNR. 
\newtextt{However, it yields a disadvantageous SSIM compared to the natural interpolation and the bilinear interpolation, in the $ \times 4 $ and $ \times 8 $ upscaling.
%In order to understand such difference, it is worth mentioning that the SSIM metric is composed of three terms, namely luminance, contrast and structural term. 
%The three terms are empirically weighted equally. 
%Table \ref{tab:compare_SSIM} presents the three terms obtained in the $ \times 4 $ upscaling.
%As one can see, the SR yielded a lower contrast, but a higher structural term.}
%\newtextt{\begin{table}[h]
%	\caption{Composition of SSIM in the $ \times 4 $ upscaling.}
%	\label{tab:compare_SSIM}
%	\centering
%	\tablesize{\footnotesize} %% You can specify the fontsize here, e.g., \tablesize{\footnotesize}. If commented out \small will be used.
%	\begin{tabular}{ccccc}
%		\toprule
%%		& \multicolumn{3}{c}{\textbf{SSIM composition terms}} \\
%%		\cline{2-4}\\[-5pt]
%		& \textbf{Luminance} & \textbf{Contrast} & \textbf{Structural} & \textbf{Yielded SSIM} \\
%		\midrule
%		Bilinear & 0.9999 & 0.7521 & 0.4012 & 0.34 \\
%		Natural & 0.9999 & 0.7458 & 0.4347 & 0.36\\
%		Proposed SR & 0.9999 & 0.7461 & 0.4085 & 0.33 \\
%		\bottomrule
%	\end{tabular}
%\end{table}
%
Considering the RMSE, one can remark that the improvement in the case of  $ \times 2 $ upscaling between the proposed SR and the others is only marginal (i.e. 1.96 compared to 2.00-2.18).
In contrast, in the $ \times 4 $ and $ \times 8 $ upscaling, this margin of RMSE improvement becomes more significant.
Similar remarks can be made when considering the PSNR.
\newtexttt{These improved quality measures show that the proposed SR is more reliable compared to the conventional interpolation methods. 
This quantitative assessment and the visual assessment (previously presented in \ref{ssec:sr}) have demonstrated the relevance of the proposed SR method.}
It is deemed to fit the purpose to be used in the proposed SRSM.}